\documentclass[11pt]{article}
\usepackage{booktabs}
\usepackage{longtable}
\usepackage{amsmath}
\usepackage{booktabs}
\usepackage{pifont} 
\newcommand{\cmark}{\textcolor{green!60!black}{\ding{51}}}
\newcommand{\xmark}{\textcolor{red}{\ding{55}}}
\usepackage[most]{tcolorbox}
\usepackage[table]{xcolor} 

\usepackage[final]{acl}

\usepackage{times}
\usepackage{latexsym}
\usepackage[utf8]{inputenc}
\usepackage[T2A]{fontenc}
\usepackage[russian,english]{babel}
\usepackage{multirow}
\usepackage[most]{tcolorbox}
\usepackage{xcolor} 
\usepackage[most]{tcolorbox} 
\tcbuselibrary{skins, breakable}
\usepackage{amssymb}
\usepackage{makecell}
\usepackage{tabularx}

\usepackage[T1]{fontenc}

\usepackage[utf8]{inputenc}

\usepackage{microtype}

\usepackage{inconsolata}

\usepackage{graphicx}
\usepackage{url}
\usepackage{multirow}

\usepackage{fontspec}
\newfontfamily\kalpurush{kalpurush.ttf}

\usepackage{inconsolata}
\usepackage{import}

\newcommand{\logoaff}[1]{\raisebox{-0.2em}{\includegraphics[height=1em]{#1}}}
\newcommand{\iconlink}[1]{\raisebox{-0.25em}{\includegraphics[height=1.1em]{#1}}}

\newcommand{\UIU}{\logoaff{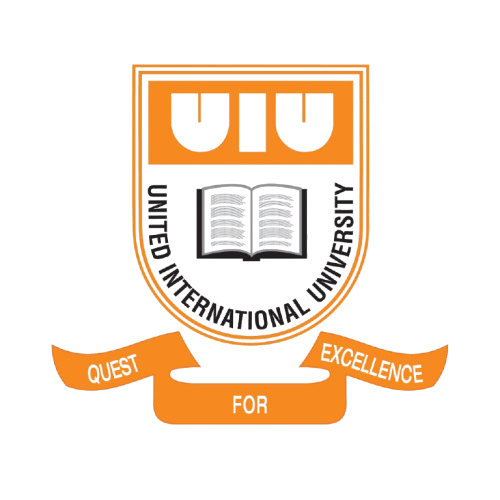}}
\newcommand{\BRACU}{\logoaff{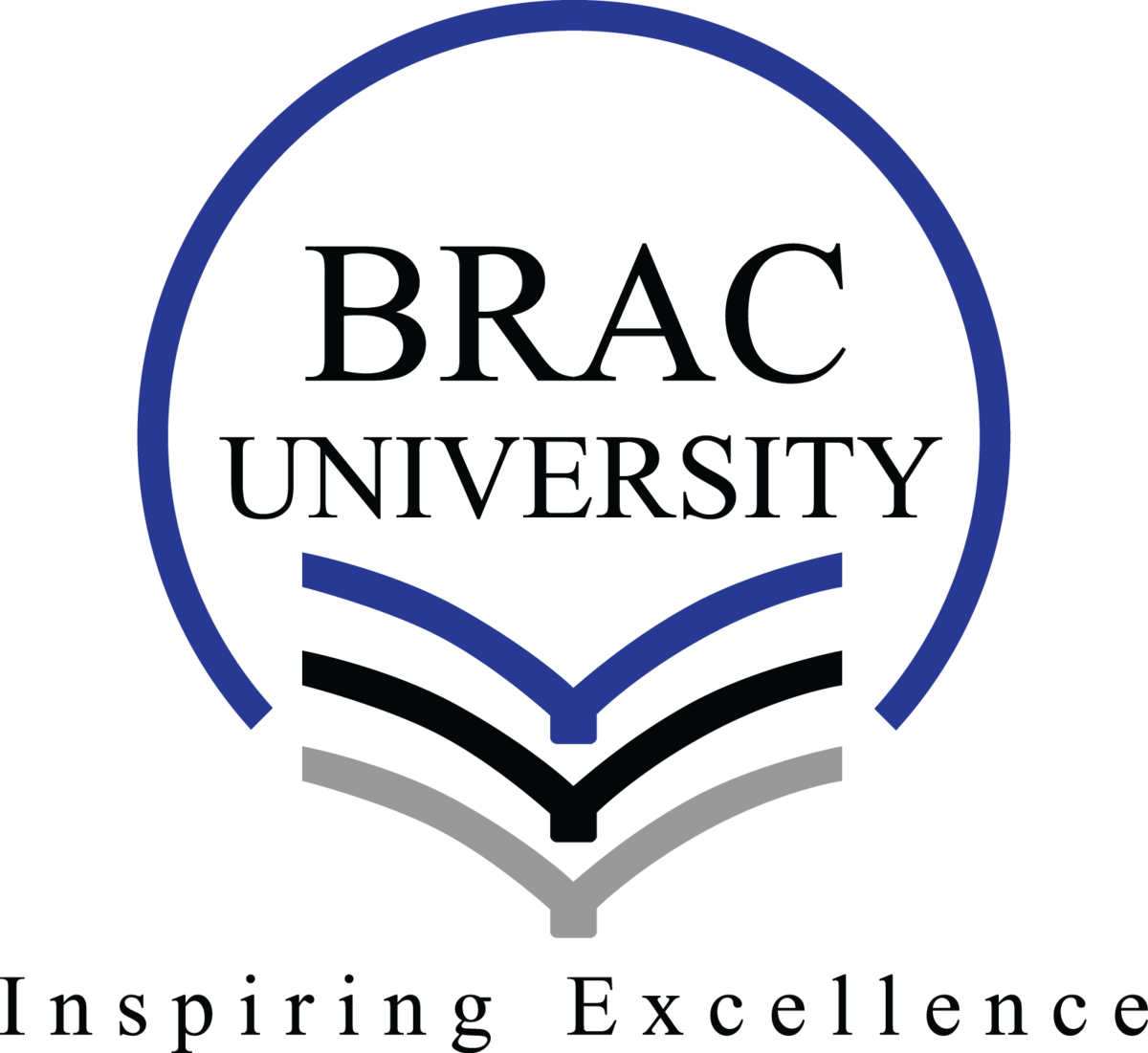}}
\newcommand{\UMBC}{\logoaff{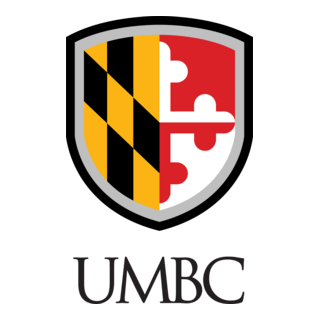}}

\newcommand{\titleicon}[1]{\raisebox{-0.25em}{\includegraphics[height=1.5em]{#1}}}

\makeatletter
\def\@fnsymbol#1{\ensuremath{\ifcase#1\or \spadesuit\or \clubsuit\or \diamondsuit\else\@ctrerr\fi}}
\makeatother

\title{\texorpdfstring{\titleicon{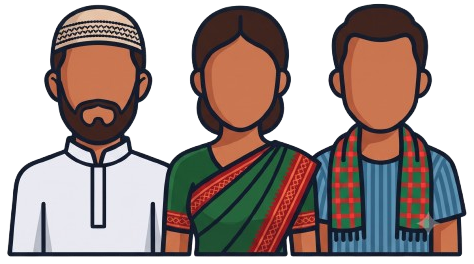}\,}{}%
\textit{Many Dialects, Many Languages, One Cultural Lens}: \\ Evaluating Multilingual VLMs for Bengali Culture Understanding \\ Across Historically Linked Languages and Regional Dialects}

\author{
 \textbf{Nurul Labib Sayeedi\,\UIU\textsuperscript{$\spadesuit$}},
 \textbf{Md. Faiyaz Abdullah Sayeedi\,\UIU\BRACU\textsuperscript{$\spadesuit$}},
 \textbf{Shubhashis Roy Dipta\,\UMBC\textsuperscript{$\clubsuit$}},\\[0.3em]
 \textbf{Mahbub E Sobhani\,\UIU \BRACU},
 \textbf{Rubaya Tabassum\,\UIU},
 \textbf{Ariful Ekraj Hridoy\,\UIU},\\[0.3em]
 \textbf{Mehraj Mahmood\,\UIU},
 \textbf{Md. Tarek Hasan\,\UIU},
 \textbf{Swakkhar Shatabda\,\BRACU}
\\[0.5em]
 \UIU\ United International University, Bangladesh \quad
 \BRACU\ BRAC University, Bangladesh
\\[0.2em]
 \UMBC\ University of Maryland, Baltimore County, USA
\\[0.5em]
{
 \href{https://labib1610.github.io/BanglaVerse/}{%
   \iconlink{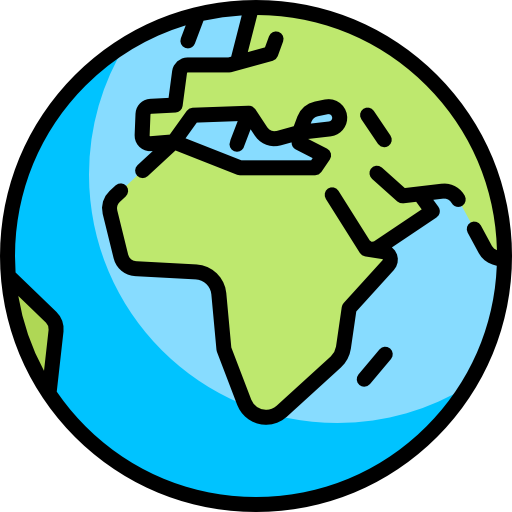}\enspace Project Page} \quad
 \href{https://github.com/faiyazabdullah/BanglaVerse}{%
   \iconlink{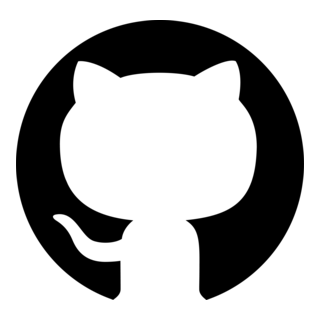}\enspace GitHub/BanglaVerse} \quad
 \href{https://huggingface.co/datasets/FaiyazAbdullah114708/BanglaVerse}{%
   \iconlink{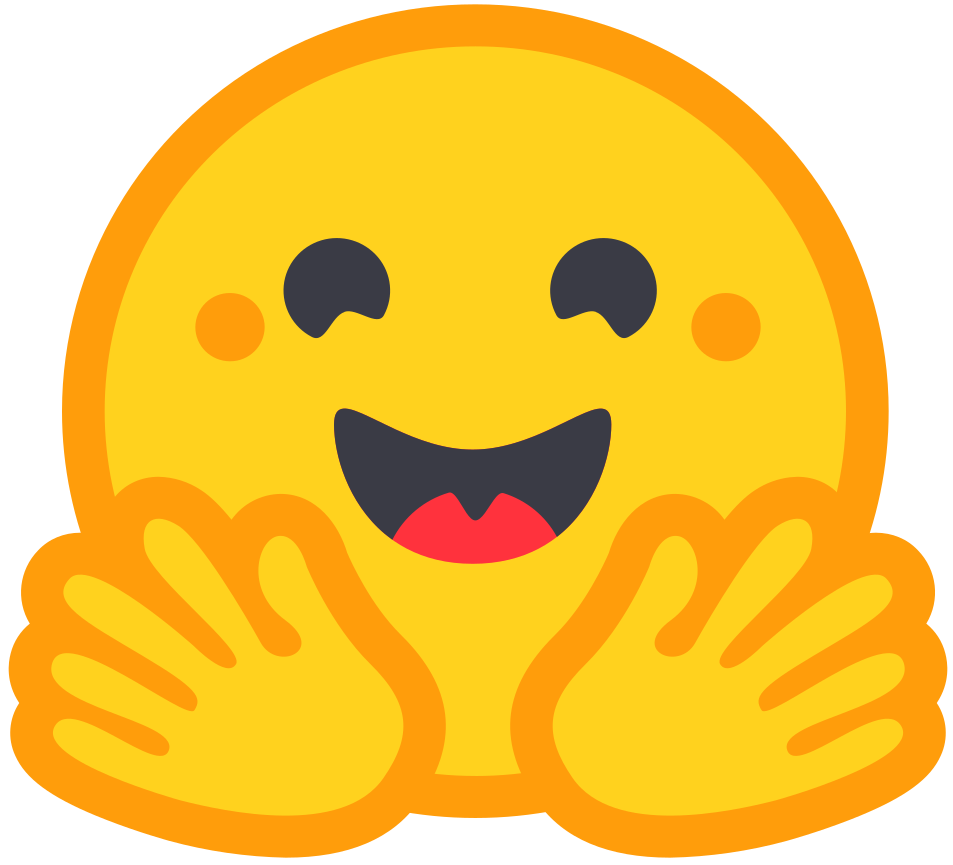}\enspace HuggingFace/Datasets/BanglaVerse}
}
}

\begin{document}
\maketitle

\renewcommand{\thefootnote}{\fnsymbol{footnote}}
\setcounter{footnote}{1}\footnotetext{Equal contribution.}
\setcounter{footnote}{2}\footnotetext{Corresponding author: \href{mailto:sroydip1@umbc.edu}{sroydip1@umbc.edu}}
\renewcommand{\thefootnote}{\arabic{footnote}}
\setcounter{footnote}{0}

\begin{abstract}  

Bangla culture is richly expressed through region, dialect, history, food, politics, media, and everyday visual life, yet it remains underrepresented in multimodal evaluation. To address this gap, we introduce \textsc{BanglaVerse}, a culturally grounded benchmark for evaluating multilingual vision–language models (VLMs) on Bengali culture across historically linked languages and regional dialects. Built from 1,152 manually curated images across nine domains, the benchmark supports visual question answering and captioning, and is expanded into four languages and five Bangla dialects, yielding $\sim$32.2K artifacts. Our experiments show that evaluating only standard Bangla overestimates true model capability: performance drops under dialectal variation, especially for caption generation, while historically linked languages such as Hindi and Urdu retain some cultural meaning but remain weaker for structured reasoning. Across domains, the main bottleneck is missing cultural knowledge rather than visual grounding alone, with knowledge-intensive categories. These findings position \textsc{BanglaVerse} as a more realistic test bed for measuring culturally grounded multimodal understanding under linguistic variation.








\end{abstract}

\section{Introduction}  
Multilingual vision--language models (MVLMs) are increasingly deployed in culturally and linguistically diverse settings, yet their evaluation remains overwhelmingly centered on standard language varieties \citep{nayak-etal-2024-benchmarking}. For Bangla, existing multimodal resources are still scarce and limited in scope \citep{10.1162/COLI.a.14}. Most available datasets focus narrowly on visual question answering and often depend heavily on synthetic annotation pipelines \citep{DBLP:journals/corr/abs-2410-14991}, resulting in benchmarks that insufficiently capture culturally grounded and linguistically authentic usage. Recent work on culturally specific benchmarks in other regions has further shown that multimodal evaluation must go beyond surface-level translation and account for cultural context, local knowledge, and linguistic diversity \citep{faraz2025indicvisionbench, liu2025culturevlm}. However, there remains no comprehensive dialect-aware benchmark for evaluating MVLMs on Bangla culture.

A central limitation of existing evaluation practices is that they treat standard Bangla as a sufficient proxy for Bengali cultural understanding. In reality, culture is not expressed only through standardized language, but also through regional dialects, historically linked languages, and locally situated forms of description, reference, and reasoning \citep{adilazuarda-etal-2024-towards}. As a result, evaluating only standard Bangla underestimates how fragile multilingual VLMs can be when the same cultural content is expressed through dialectal and cross-lingual variation. A model that appears competent in standard Bangla may still fail to preserve cultural meaning, visual grounding, or linguistic consistency when tested on regional varieties.

\begin{table*}[t]
\centering
\small
\resizebox{0.95\textwidth}{!}{%
\begin{tabular}{l|c|c|c|c|c|c|c|c}
\toprule
\rowcolor{gray!15}
\textbf{Dataset} & \textbf{VQA} & \textbf{CAP.} & \textbf{\#ART.} & \textbf{CA} & \textbf{CMA} & \textbf{Multi} & \textbf{DIAL} & \textbf{H. Eval} \\
\midrule
BengaliVQA \citep{10.1145/3530190.3534837} & \cmark & \xmark & 5,000  & \xmark & \xmark & \xmark & \xmark & \xmark \\
Bengali CLEVR \citep{hasan-etal-2023-visual} & \cmark & \xmark & 12,291 & \xmark & \xmark & \xmark & \xmark & \xmark \\
Bangla VQA v2.0 \citep{rafi2022deep} & \cmark & \xmark & 13,046 & \xmark & \xmark & \xmark & \xmark & \xmark \\
CVQA \citep{DBLP:conf/nips/RomeroLWGMPOVBJ24} & \cmark & \xmark & 286 & \cmark & \cmark & \cmark & \xmark & \xmark \\
ChitroJera \citep{DBLP:journals/corr/abs-2410-14991} & \cmark & \xmark & 15,292 & \cmark & \cmark & \xmark & \xmark & \cmark \\
BanglaProtha \citep{fahim2026banglaprotha} & \cmark & \xmark & 8,034 & \cmark & \cmark & \xmark & \xmark & \cmark \\
\midrule
\rowcolor{blue!15}
\textbf{\textsc{BanglaVerse} (Ours)} & \cmark & \cmark & \textbf{32,247} & \cmark & \cmark & \cmark & \cmark & \cmark \\
\bottomrule
\end{tabular}
}
\caption{Comparison with related Bangla and culturally grounded multimodal benchmarks. VQA = Visual Question Answering, CAP. = Captioning, \#ART. = Total Number of Artifacts, CA = Cultural Awareness, CMA = Categorical Metadata Availability, Multi = Multilingual, DIAL = Dialectal Coverage, and H. Eval = Human Evaluation.}
\label{tab:dataset_comparison}
\end{table*}


To address this gap, we present \textsc{\textsc{BanglaVerse}}, a culturally grounded multilingual and multidialectal benchmark for evaluating MVLMs on Bengali culture across standard Bangla, historically linked languages, and regional dialects. \textsc{BanglaVerse} is built from a manually curated, image-centric core dataset spanning nine culturally rich domains, and is extended across four languages: Bangla, English, Hindi, and Urdu; as well as five Bangla dialects: Barishal, Chittagong, Noakhali, Rangpur, and Sylhet. The benchmark supports two image-grounded tasks: Visual Question Answering (VQA) and Image Captioning (CAP). Our main contributions are as follows:
\begin{itemize}
    \item We introduce \textbf{\textsc{BanglaVerse}}, a culturally grounded multilingual and multidialectal benchmark for Bengali culture understanding, built on 1,152 manually curated images with high-quality annotations verified through a collaborative cross-checking process.
    \item We expand the benchmark to four languages and five Bangla dialects, yielding $\sim$32.2K total artifacts across captioning and visual question answering tasks, and enabling systematic evaluation across historically linked languages and regional varieties.
    \item We conduct a comprehensive empirical study of state-of-the-art multilingual VLMs under multiple prompting strategies, showing that evaluation only on standard Bangla can obscure substantial weaknesses that emerge under dialectal and cross-lingual variation.
\end{itemize}


\section{Background and Related Works}

\subsection{Background}
Bangla has evolved through a long history of cultural, political, and linguistic exchange across South Asia, making Bengali cultural expression deeply connected to neighboring languages and regional speech varieties \citep{shahen2019globalization}. Motivated by this history, we extend our benchmark beyond standard Bangla to English, Hindi, and Urdu, which remain important due to their historical, cultural, and communicative links with Bangla, as well as their relevance in education, media, and digital communication \citep{sakib2023comparing}. At the same time, evaluating Bengali culture only through standard Bangla is insufficient, since much of Bangladesh’s cultural life is expressed through regional dialects and locally grounded forms of speech \citep{hasan2014standard, karmaker2019dialectical, hasan-roy-dipta-2025-banglatalk, hasan2026banglaipa}. We therefore include five major Bangla dialects, Barishal, Chittagong, Noakhali, Rangpur, and Sylhet, which together capture a broad range of socially and culturally salient variation across Bangladesh. 


\begin{figure*}[t] 
    \centering
    \includegraphics[width=\textwidth]{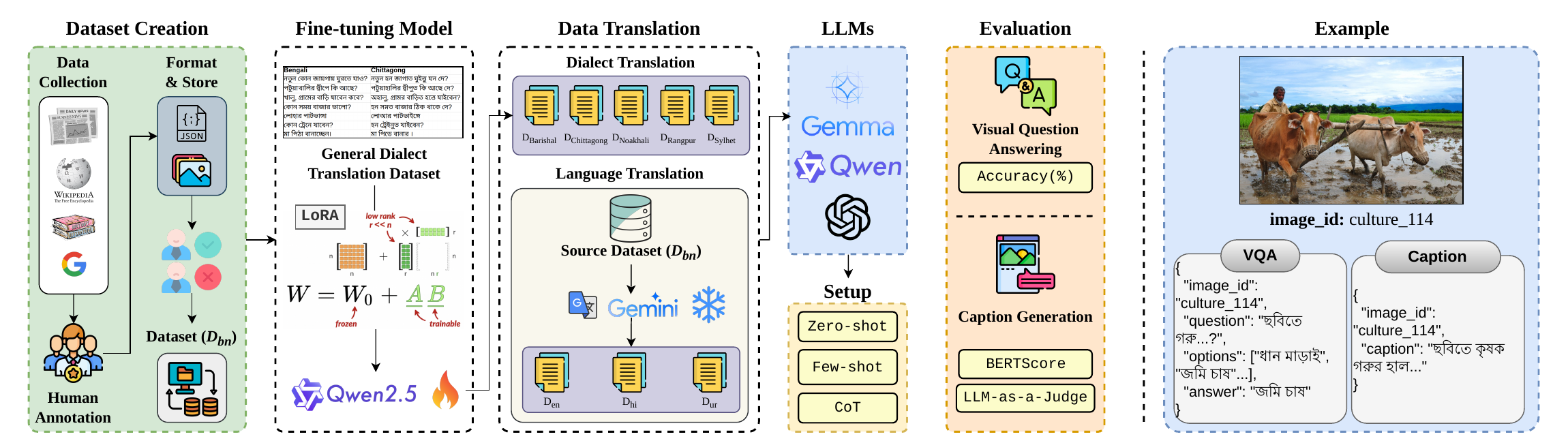} 
    \caption{Overview of the \textsc{BanglaVerse} dataset and experimental setup. The figure shows the two task types, example annotations for each task, artifacts generation and evaluation pipeline with multiple metrics.}
    \label{fig:method}
\end{figure*}

\subsection{Related Works}

A growing body of work has emphasized that multimodal evaluation should reflect cultural context rather than relying solely on generic or Western-centric imagery. In Bangla, several multimodal datasets have been introduced, but most remain limited either in cultural scope or annotation authenticity. BengaliVQA \citep{10.1145/3530190.3534837}, Bengali CLEVR \citep{hasan-etal-2023-visual}, and Bangla VQA v2.0 \citep{rafi2022deep} primarily target visual question answering, but the first two are translated from English benchmarks while the third remains relatively narrow in cultural coverage. CVQA \citep{DBLP:conf/nips/RomeroLWGMPOVBJ24} and ChitroJera \citep{DBLP:journals/corr/abs-2410-14991} move closer to culturally aware evaluation, yet CVQA is small in scale and ChitroJera relies heavily on GPT-4 Turbo-generated annotations, which may limit authenticity and grounding in real cultural usage.


A recent and closely related effort is BanglaProtha \citep{fahim2026banglaprotha}, which studies Bengali cultural understanding as a long-tail multimodal evaluation problem. It introduces a Bengali cultural VQA benchmark organized around multiple cultural aspects, with native Bengali questions and semantically similar answer options, and evaluates different classes of VLMs under prompting and fine-tuning settings. While it’s a vital step forward for representing Bengali culture, it strictly focuses on VQA and misses the region's rich dialectal and multilingual variations. Beyond Bangla, there is growing interest in culturally grounded multimodal benchmarks for other languages and regions. \citet{maji2025drishtikon} introduced DRISHTIKON, a multilingual benchmark centered on Indian cultural contexts, while \citet{alwajih-etal-2025-palm} proposed PALM, a large-scale, community-driven dataset spanning all 22 Arab countries with instruction--response pairs in both Modern Standard Arabic and dialectal variants. These efforts highlight the importance of cultural inclusivity, contextual grounding, and broader linguistic coverage in multimodal and language model evaluation.

Despite this progress, an important limitation remains across existing Bangla benchmarks: evaluation is almost always conducted only in standard Bangla. In practice, however, culture is expressed not only through standardized language but also through regional dialects, local lexical choices, and historically linked languages \citep{keleg2025llm}. As a result, existing resources cannot reveal whether multilingual VLMs genuinely understand Bengali culture or merely perform well on standardized forms of it. A comparison of existing Bangla and culturally grounded multimodal benchmarks is provided in Table~\ref{tab:dataset_comparison}.


\section{Dataset Creation}

In this section, we present the overall pipeline for the \textsc{BanglaVerse} dataset, shown in Figure~\ref{fig:method}. 

\subsection{Data Collection}
We collected images and associated textual context from a combination of online and offline sources, including news articles, Wikipedia, Banglapedia, historical books, and general knowledge books. A comprehensive list of these sources are provided in Appendix~\ref{app:data-sources}. Initially, we gathered a preliminary pool of 2,019 candidate instances. We then rigorously curated the dataset following specific taxonomy-driven guidelines (Appendix \ref{app:curation_guidelines}) to ensure the data accurately reflected the target cultural concepts. This curation was conducted by the five authors, who were all native Bengali speakers. After the curation, the final dataset contained 1,152 unique images. We organized the final curated materials into nine major domains: \textit{Culture, Food, History, Media \& Movies, National Achievements, Nature, Personalities, Politics, and Sports}. These specific domains were carefully selected because they collectively encapsulate the multifaceted nature of Bengali identity. Rather than defining culture strictly through traditional customs or attire, we adopt a holistic framework where historical milestones, regional geography, traditions, public figures, and everyday entertainment all contribute to the societal ethos. Each domain contains an \texttt{images} directory and an \texttt{annotations} directory.

\begin{table*}[t]
\centering
\small
\setlength{\tabcolsep}{8pt}
\renewcommand{\arraystretch}{1.15}
\begin{tabular}{l l r l l r l l r}
\toprule
\rowcolor{gray!15}
\textbf{Category} & \textbf{Type} & \textbf{Count} &
\textbf{Category} & \textbf{Type} & \textbf{Count} &
\textbf{Category} & \textbf{Type} & \textbf{Count} \\
\midrule
\multirow{4}{*}{\makecell[c]{\includegraphics[height=2em]{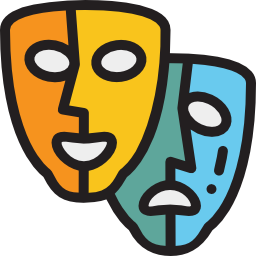}\\\textbf{Culture}}}
& Images & 114 & \multirow{4}{2cm}{\makecell[c]{\includegraphics[height=2em]{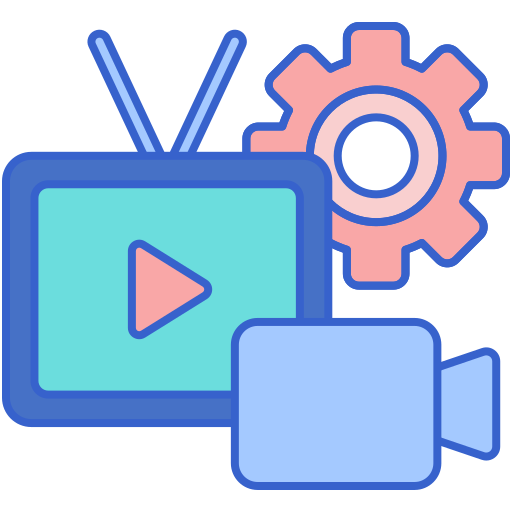}\\\textbf{Media \&}\\\textbf{Movies}}}
& Images & 150 & \multirow{4}{*}{\makecell[c]{\includegraphics[height=2em]{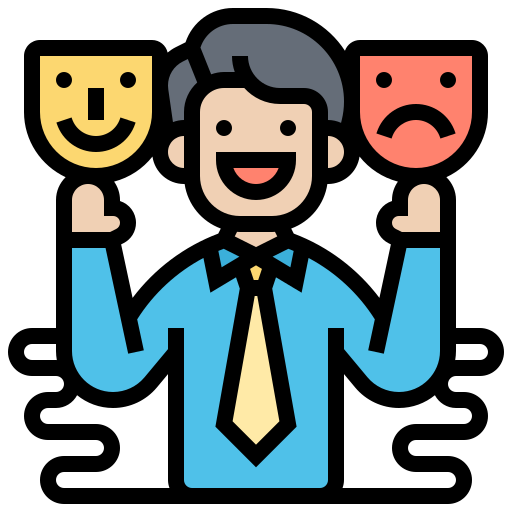}\\\textbf{Personalities}}}
& Images & 150 \\
& Captions & 1,026 && Captions & 1,350 && Captions & 1,350 \\
& VQA & 2,052 && VQA & 2,592 && VQA & 2,700 \\
& \textbf{Total} & \textbf{3,192} && \textbf{Total} & \textbf{4,092} && \textbf{Total} & \textbf{4,200} \\
\midrule
\multirow{4}{1cm}{\makecell[c]{\includegraphics[height=2em]{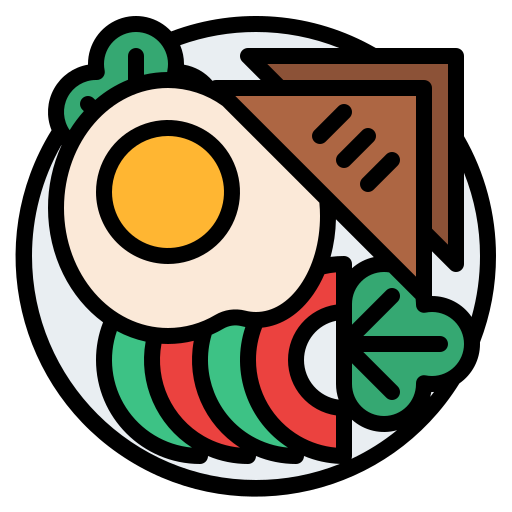}\\\textbf{Food}}}
& Images & 150 & \multirow{4}{*}{\makecell[c]{\includegraphics[height=2em]{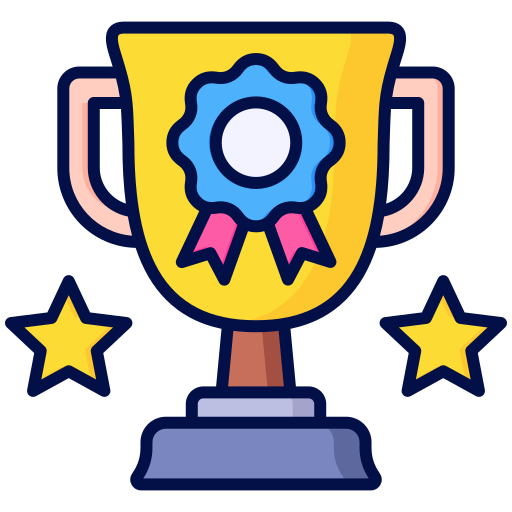}\\\textbf{National}\\\textbf{Achievements}}}
& Images & 75 & \multirow{4}{1.7cm}{\makecell[c]{\includegraphics[height=2em]{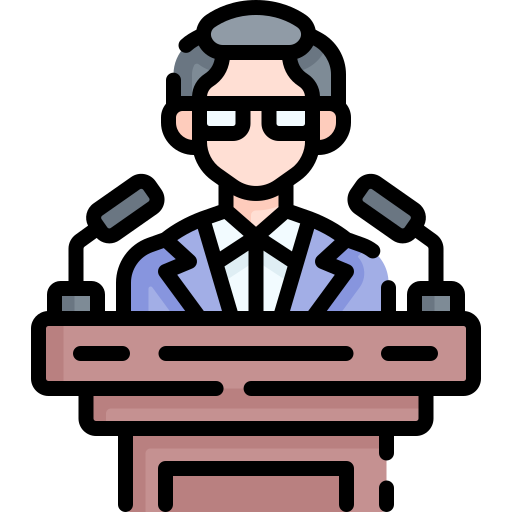}\\\textbf{Politics}}}
& Images & 150 \\
& Captions & 1,350 && Captions & 675 && Captions & 1,350 \\
& VQA & 2,709 && VQA & 1,332 && VQA & 2,754 \\
& \textbf{Total} & \textbf{4,209} && \textbf{Total} & \textbf{2,082} && \textbf{Total} & \textbf{4,254} \\
\midrule
\multirow{4}{*}{\makecell[c]{\includegraphics[height=2em]{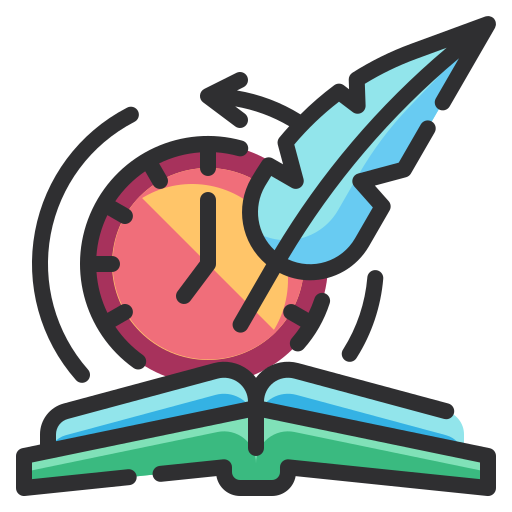}\\\textbf{History}}}
& Images & 150 & \multirow{4}{2cm}{\makecell[c]{\includegraphics[height=2em]{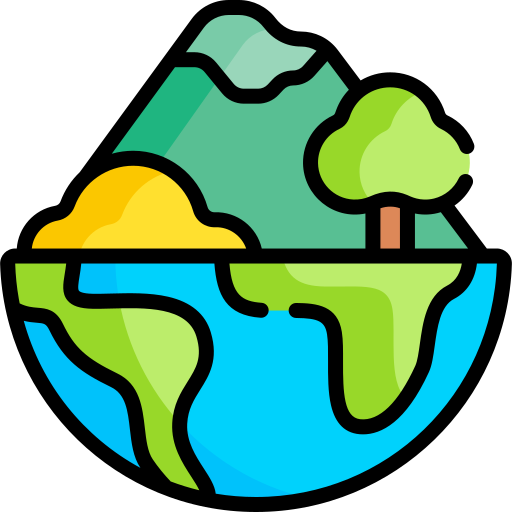}\\\textbf{Nature}}}
& Images & 150 & \multirow{4}{1.7cm}{\makecell[c]{\includegraphics[height=2em]{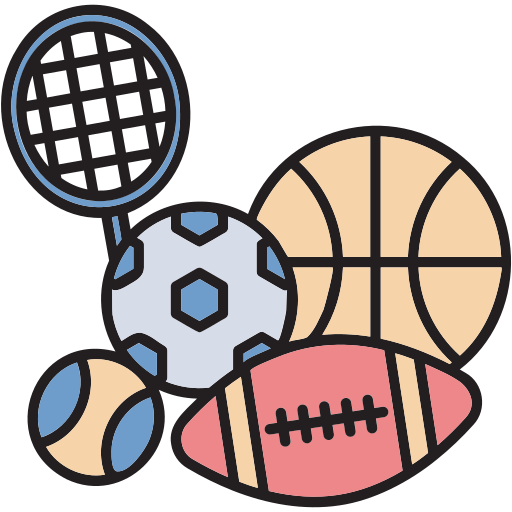}\\\textbf{Sports}}}
& Images & 63 \\
& Captions & 1,350 && Captions & 1,350 && Captions & 567 \\
& VQA & 2,772 && VQA & 2,691 && VQA & 1,125 \\
& \textbf{Total} & \textbf{4,272} && \textbf{Total} & \textbf{4,191} && \textbf{Total} & \textbf{1,755} \\
\midrule
\rowcolor{blue!10}
\multicolumn{8}{l}{\textbf{Unique Images}} & \textbf{1,152} \\
\rowcolor{blue!5}
\multicolumn{8}{l}{\textbf{Captions (9 linguistic varieties)}} & \textbf{10,368} \\
\rowcolor{blue!10}
\multicolumn{8}{l}{\textbf{VQA Items (9 linguistic varieties)}} & \textbf{20,727} \\
\midrule
\rowcolor{blue!20}
\multicolumn{8}{l}{\textbf{Total Artifacts}} & \textbf{32,247} \\
\bottomrule
\end{tabular}
\caption{Overall statistics of \textsc{BanglaVerse} across 4 languages and 5 Bangla dialects. Caption and VQA counts are expanded across the 9 linguistic varieties; each domain total is the sum of its images, captions, and VQA items.}
\label{tab:dataset-stats}
\end{table*}

\subsection{Annotation Setup}

The dataset was annotated by the same five authors through a collaborative process. Four primary annotators are senior undergraduate students majoring in Data Science who possess a deep academic knowledge of LLMs and their relationship with cultural understanding. A fifth author, serving as the adjudicator, is a domain expert with prior research experience in LLM cultural understandings, who was responsible for resolving any disagreements among the annotators. Each annotator was initially assigned two domains, with one annotator covering three domains to account for all nine domains. The annotated instances were then reassigned to other annotators on a domain-wise basis. This process helped minimize annotator-specific bias and improve consistency. We ensured that each domain was annotated by at least two primary annotators, achieving an overall inter-annotator agreement of $\kappa = 0.87$. For a detailed domain-wise breakdown, see Appendix~\ref{app:inter_annotator_details}. Disagreements most commonly involved the degree of cultural specificity to include, the interpretation of culturally nuanced visual elements, and the formulation of semantically similar but clearly incorrect answers. Such cases were discussed among the annotators and resolved by the adjudicating author. 

\subsection{Artifacts Generation}

We generated the final benchmark artifacts through a two-stage pipeline built on top of the manually curated Bangla source dataset. Each source instance consists of an image paired with two image-grounded task annotations: a caption and a VQA item. The source Bangla annotations served as the canonical reference point for all subsequent multilingual and dialectal artifact generation.

In the first stage, we performed multilingual expansion by translating the Bangla annotations into three historically linked languages: English, Hindi, and Urdu. For each target language, we used \texttt{Gemini-2.5-Flash} to translate both captions and VQA instances. During this process, we aimed to preserve not only the literal semantic content of the source annotation but also its cultural specificity, pragmatic intent, and alignment with the visual content. In particular, we took care to ensure that culturally salient references, named entities, and locally meaningful expressions were retained as faithfully as possible rather than being replaced with more generic alternatives. 


In the second stage, we generated multidialectal variants of the benchmark to capture regional linguistic diversity within Bangla itself. We considered five major Bangla dialects: Barishal, Chittagong, Noakhali, Rangpur, and Sylhet. Rather than directly prompting a general-purpose model to produce dialectal outputs, we adopted a dedicated dialect generation strategy. We first fine-tuned the \texttt{Qwen2.5-3B-Instruct} model on a native standard-Bangla-to-dialect translation dataset derived from the BanglaDial corpus \citep{mahi2025bangladial}, so that it could better capture lexical and morphological features specific to each dialect. Because the original corpus does not feature a predefined evaluation set, we randomly partitioned the dataset into an 80\% training and 20\% held-out test split. To validate this, we conducted an intrinsic evaluation on this newly created test split, achieving a strong average BLEU of 30.16 and ChrF of 54.57 across all dialects, and demonstrating exceptional semantic preservation with an average BS-L3Cube F1 of 0.787 and a Gemini Embedding Similarity of 0.973. See Appendix \ref{app:fine-tuning} for additional details regarding the fine-tuning model. The fine-tuned model was then used to convert the source Bangla captions and VQA items into dialect-specific forms. By keeping the image fixed, we created a controlled setting for studying how model performance changes when the same Bengali cultural content is expressed through different dialectal varieties.

To rigorously validate the authenticity of our generated artifacts, we also conducted a comprehensive human evaluation on a representative subset of the data. Five senior Computer Science undergraduates familiar with LLMs, each a native speaker of one of the five different dialects, voluntarily evaluated the dialectal outputs. Concurrently, three of the authors performed back-translation validation on the multilingual variants. We systematically sampled 10 instances per domain for each of the 9 linguistic variants, resulting in 90 samples per variant and 810 manually assessed instances in total. Evaluators assigned scores using a discrete 3-point scale: 0 (Inaccurate), 1 (Acceptable with Minor Revisions), and 2 (Flawless). The results demonstrated exceptional generation quality and authenticity, with 92.6\% of the evaluated artifacts receiving the maximum score of 2. We have included a boolean \texttt{is\_human\_validated} metadata field across the dataset to explicitly track this distinction. A value of \texttt{true} indicates the artifact was manually authenticated by our evaluators, while \texttt{false} denotes it was generated or translated without direct human review. Further details regarding the human evaluation are available in Appendix \ref{app:human-eval}.

\subsection{Dataset Statistics}

\begin{figure}[t]
    \centering
    \includegraphics[width=0.9\linewidth]{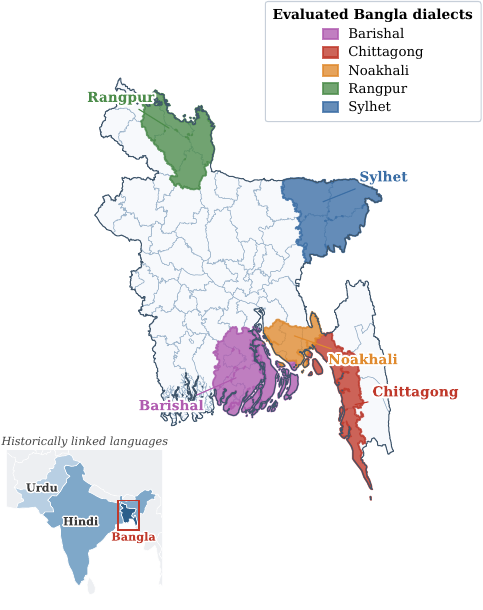}
    \caption{Geographical distribution of evaluated dialects and the historically linked languages.}
    \label{fig:map}
    \vspace{-3mm}
\end{figure}

Table~\ref{tab:dataset-stats} summarizes the overall corpus statistics across nine cultural domains, four languages, and five Bangla dialects. The benchmark comprises a foundational set of 1,152 unique images. To evaluate multidialectal and multilingual capabilities, the base annotations for each image were expanded across nine distinct linguistic variations (Figure \ref{fig:map}). Consequently, the base caption and VQA counts are multiplied by 9, resulting in 10,368 captions and 20,727 VQA pairs, yielding 32,247 artifacts overall. Domains such as \textit{Food}, \textit{History}, \textit{Media \& Movies}, \textit{Nature}, \textit{Personalities}, and \textit{Politics} contribute the largest number of samples, whereas \textit{National Achievements} ($n=75$) and \textit{Sports} ($n=63$) remain comparatively smaller. The smaller scale of these two domains reflects their inherently factual and event-based nature. Their inclusion strictly depends on the occurrence of highly notable, documented milestones or specific sporting events, which are naturally fewer in number than general cultural concepts. Among all domains, \textit{History} and \textit{Politics} contain the highest number of VQA annotations. The average caption length is $\approx$ 90--140 characters, while the average VQA question length ranges from 50--80 characters. This indicates that the dataset goes beyond very short or template-like annotations and supports more meaningful evaluation of culturally grounded multimodal understanding.

\subsection{Overview of Tasks}

\paragraph{Visual Question Answering.}
For each image, we construct a VQA instance designed to test both direct visual understanding and culturally grounded commonsense reasoning. Each question is formulated based on the visible image content, while also requiring the model to interpret culturally specific practices, objects, events, or implicit context when necessary. To maintain consistency and evaluability, each VQA item is paired with four multiple-choice answer options, consisting of one correct ground-truth answer and three carefully crafted distractors. For example, for \texttt{image\_id: culture\_114}, we include the question:
{\kalpurush ছবিতে গরু ব্যবহার করে কী কাজ করা হচ্ছে?} 
\textit{(What task is being performed using the cow in the image?)} 
with options 
{\kalpurush ["ধান মাড়াই","জমি চাষ","শস্য মজুদ","গো-খাদ্য প্রস্তুত"]} 
\textit{([``Paddy threshing", ``Land cultivation", ``Grain reserve", ``Cattle feed preparation"])}, 
where the correct answer is 
{\kalpurush "জমি চাষ"} 
\textit{(``Plowing the field")}. To prevent models from relying on simple process-of-elimination, the distractors are designed to be culturally relevant and visually plausible. Instead of using random or out-of-context phrases, we curate incorrect options that represent related regional activities, similar objects, or plausible alternative scenarios (e.g., ``Threshing rice" is a valid agricultural task but incorrect for the specific image). 



\paragraph{Caption Generation.}
Each image is also paired with a natural language caption describing the visual content in Bengali. The captions are designed to provide concise yet culturally informative descriptions of the entities, actions, and contexts depicted in the image. Rather than merely listing visible objects, the captions often include culturally meaningful details that help situate the image within Bengali social, historical, or everyday life. For instance, for \texttt{image\_id: culture\_114}, the caption is:
{\kalpurush "ছবিতে কৃষক গরুর হাল দিয়ে জমি চাষ করছেন, যা গ্রামীণ কৃষির ঐতিহ্য।"} 
\textit{(The farmer is plowing the field with oxen, a tradition of rural agriculture.)}. This task evaluates whether a model can generate fluent and visually grounded descriptions while preserving the cultural significance.


\begin{figure*}[t]
    \centering
    \includegraphics[width=0.9\linewidth]{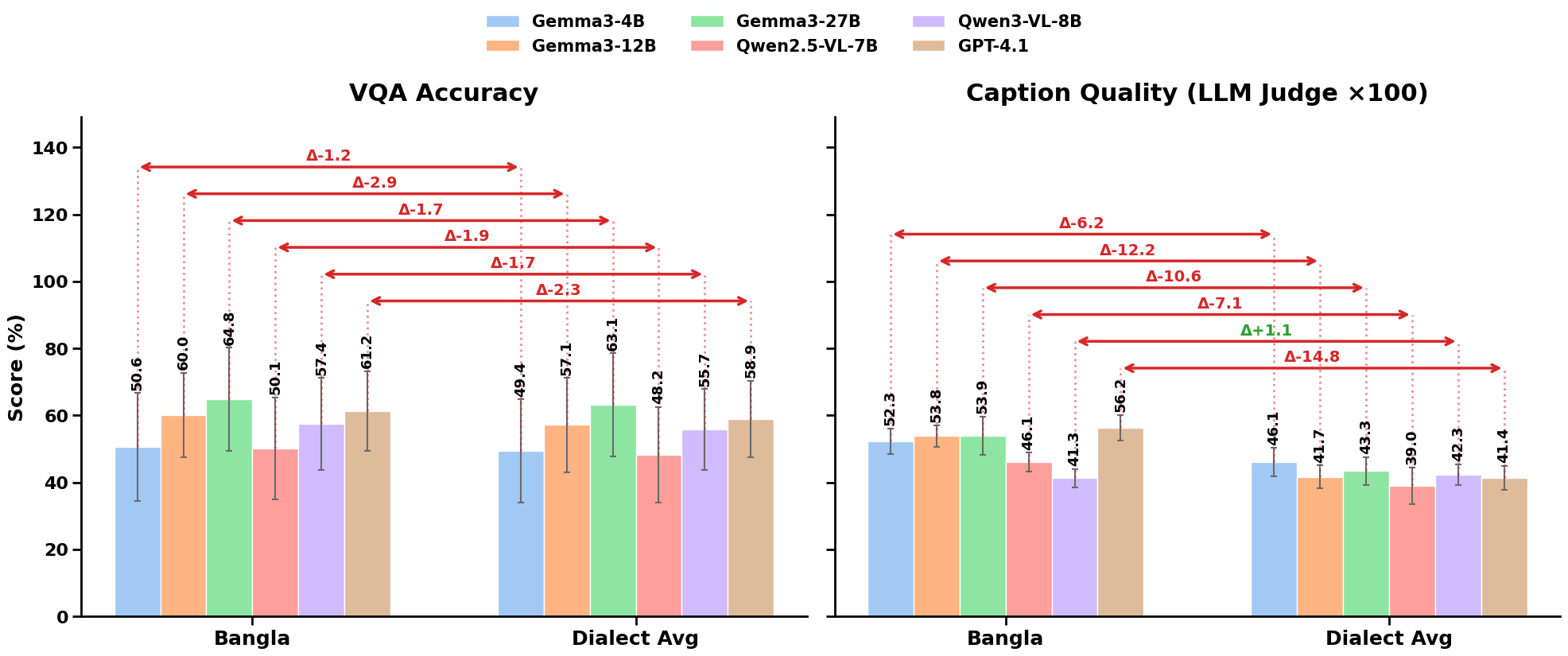}
    \caption{\textbf{Dialect variation across models.} We compare each model's average VQA accuracy (zero-shot, few-shot, and CoT) and caption quality on standard Bangla against its mean performance across the five regional dialects.}
    \label{fig:rq1_dialect_robustness}
\end{figure*}


\section{Experimental Setup}

\subsection{Models}


We evaluated several multilingual and multimodal open- and closed-source models. Our experiments include \texttt{Gemma3-4B}, \texttt{Gemma3-12B}, \texttt{Gemma3-27B}, \texttt{Qwen2.5-VL-7B}, \texttt{Qwen3-VL-8B}, and \texttt{GPT-4.1-mini}. This selection allows us to compare models of different scales, architectural families, and training regimes, and to examine how such differences affect culturally grounded vision--language understanding.

\subsection{Prompting Techniques}
To systematically evaluate the models' culturally grounded vision-language understanding, we employ zero-shot, few-shot, and chain-of-thought (CoT) \citep{wei2022chain} prompting strategies. The zero-shot and few-shot templates strictly instruct the models to generate outputs exclusively in the source language without extraneous explanations. For the few-shot setting, a fixed set of examples was randomly sampled from a held-out split for each specific domain and language, and these identical examples were applied consistently across all evaluated models. For more complex reasoning in the VQA task, CoT prompting is utilized, allowing models to generate intermediate logical steps before yielding the final formatted answer. Across all evaluation settings, the decoding temperature is set to 0.1 to ensure highly deterministic outputs. All prompts are provided in Appendix \ref{app:prompt}.


\subsection{Evaluation Metrics}

We evaluate the benchmark using task-specific metrics tailored to captioning and visual question answering. For \textbf{image captioning}, we report BERTScore-F1 \citep{Zhang*2020BERTScore:} to measure semantic similarity between generated and reference captions, and LLM-as-a-Judge \citep{gu2024survey} scores, primarily utilizing \texttt{Gemini-2.5-Flash} alongside \texttt{GPT-5-mini} for robust cross-validation, to capture overall caption quality beyond surface-level lexical overlap. The models evaluate captions across four dimensions: Relevance, Clarity, Conciseness, and Creativity, to compute a final holistic score ranging from 0 to 1. For \textbf{visual question answering}, we use accuracy (\%), defined as the percentage of questions for which the model selects the correct answer option. 

To assess the reliability of the LLM-as-a-Judge evaluation, we conducted a focused human review with two independent subject-matter experts (SMEs) who are native Bangla speakers and hold degrees in culturally grounded language interpretation. Inter-annotator agreement between the two SMEs was measured over the evaluated samples using Cohen’s $\kappa$, yielding a score of $0.78$, which indicates strong agreement. We adopted the 0--100 human scoring rubric with three components proposed by \citet{e-sobhani-etal-2026-mathmist}: (i) \textit{decision accuracy} (0--50 points): evaluates whether the judge makes the correct quality judgment; (ii) \textit{reasoning alignment} (0--40 points): measures whether the supporting analysis is logically consistent with the decision; and (iii) \textit{explanation clarity} (0--10 points): captures how clear and well-justified the explanation is. Using this rubric, the SMEs reviewed 100 output samples. All the samples evaluated in this phase were exclusively in Standard Bangla, specifically chosen to leverage the SMEs' native proficiency and deep expertise in Bengali history, culture, and linguistic depth. The Gemini-as-a-Judge achieved a Human Consistency Score ranging from 90.89 to 92.01 out of 100 across experiments, indicating that its judgments align closely with expert evaluation. Furthermore, the inclusion of \texttt{GPT-5-mini} successfully mitigated concerns of self-preference bias. Across all nine linguistic variants, the two judges exhibited a strong positive correlation (Avg. Pearson $r=0.706$, Avg. Spearman Rank $\rho=0.701$), confirming that the relative rankings of model caption quality assessment remain highly stable (see Appendix~\ref{app:cross-model-validation} for full details).


\section{Results and Analysis}

In this section, we analyze the performance of multilingual VLMs through three core research questions (RQs) For full experimental results, see the Appendix \ref{app:full_results}.




\begin{table*}[t]
\centering
\small
\resizebox{\textwidth}{!}{%
\begin{tabular}{lcccccccc}
\toprule
\multirow{2}{*}{\textbf{Model}} & \multicolumn{2}{c}{\textbf{Bangla}} & \multicolumn{2}{c}{\textbf{English (vs. BN)}} & \multicolumn{2}{c}{\textbf{Hindi (vs. BN)}} & \multicolumn{2}{c}{\textbf{Urdu (vs. BN)}} \\
\cmidrule(lr){2-3} \cmidrule(lr){4-5} \cmidrule(lr){6-7} \cmidrule(lr){8-9}
 & \textbf{VQA} & \textbf{Cap.} & \textbf{VQA} & \textbf{Cap.} & \textbf{VQA} & \textbf{Cap.} & \textbf{VQA} & \textbf{Cap.} \\
\midrule
\textbf{Gemma3-4B} 
& 50.6 & 52.3 & 51.6 (\textcolor{green!60!black}{+1.0}) & 60.2 (\textcolor{green!60!black}{+7.9}) & 50.5 (\textcolor{red}{-0.1}) & 53.2 (\textcolor{green!60!black}{+0.9}) & 46.2 (\textcolor{red}{-4.4}) & 47.2 (\textcolor{red}{-5.1}) \\
\quad \textit{Std. Dev.} & $\pm$14.77 & $\pm$3.80 & $\pm$11.85 & $\pm$4.24 & $\pm$12.26 & $\pm$3.55 & $\pm$10.60 & $\pm$1.76 \\
\midrule
\textbf{Gemma3-12B} 
& 60.0 & 53.8 & 59.3 (\textcolor{red}{-0.7}) & 58.6 (\textcolor{green!60!black}{+4.8}) & 53.0 (\textcolor{red}{-7.0}) & 54.0 (\textcolor{green!60!black}{+0.2}) & 47.7 (\textcolor{red}{-12.3}) & 53.0 (\textcolor{red}{-0.8}) \\
\quad \textit{Std. Dev.} & $\pm$14.46 & $\pm$3.15 & $\pm$10.65 & $\pm$4.23 & $\pm$10.85 & $\pm$2.98 & $\pm$11.30 & $\pm$2.16 \\
\midrule
\textbf{Gemma3-27B} 
& 64.8 & 53.9 & 66.2 (\textcolor{green!60!black}{+1.4}) & 61.3 (\textcolor{green!60!black}{+7.4}) & 60.7 (\textcolor{red}{-4.1}) & 55.8 (\textcolor{green!60!black}{+1.9}) & 54.6 (\textcolor{red}{-10.2}) & 56.2 (\textcolor{green!60!black}{+2.3}) \\
\quad \textit{Std. Dev.} & $\pm$14.87 & $\pm$5.81 & $\pm$14.11 & $\pm$5.36 & $\pm$13.73 & $\pm$3.75 & $\pm$12.91 & $\pm$3.96 \\
\midrule
\textbf{Qwen2.5-VL-7B} 
& 50.1 & 46.1 & 55.7 (\textcolor{green!60!black}{+5.6}) & 59.5 (\textcolor{green!60!black}{+13.4}) & 49.6 (\textcolor{red}{-0.5}) & 45.5 (\textcolor{red}{-0.6}) & 50.1 (\textcolor{gray}{0.0}) & 41.0 (\textcolor{red}{-5.1}) \\
\quad \textit{Std. Dev.} & $\pm$13.12 & $\pm$2.75 & $\pm$11.42 & $\pm$5.94 & $\pm$11.60 & $\pm$3.89 & $\pm$11.43 & $\pm$4.59 \\
\midrule
\textbf{Qwen3-VL-8B} 
& 57.4 & 41.3 & 58.8 (\textcolor{green!60!black}{+1.4}) & 60.2 (\textcolor{green!60!black}{+18.9}) & 53.3 (\textcolor{red}{-4.1}) & 53.8 (\textcolor{green!60!black}{+12.5}) & 49.8 (\textcolor{red}{-7.6}) & 50.0 (\textcolor{green!60!black}{+8.7}) \\
\quad \textit{Std. Dev.} & $\pm$12.71 & $\pm$2.69 & $\pm$12.42 & $\pm$6.19 & $\pm$13.78 & $\pm$4.72 & $\pm$12.40 & $\pm$6.82 \\
\midrule
\textbf{GPT-4.1-mini} 
& 61.2 & 56.2 & 65.0 (\textcolor{green!60!black}{+3.8}) & 58.4 (\textcolor{green!60!black}{+2.2}) & 57.2 (\textcolor{red}{-4.0}) & 56.2 (\textcolor{gray}{0.0}) & 52.3 (\textcolor{red}{-8.9}) & 55.1 (\textcolor{red}{-1.1}) \\
\quad \textit{Std. Dev.} & $\pm$11.60 & $\pm$3.85 & $\pm$13.37 & $\pm$7.56 & $\pm$12.80 & $\pm$4.07 & $\pm$9.97 & $\pm$4.45 \\
\bottomrule
\end{tabular}
}
\caption{\textbf{Cross-lingual preservation of Bengali cultural meaning across models.} Values in parentheses denote the delta (increase in \textcolor{green!60!black}{green}, degradation in \textcolor{red}{red}) compared to the Standard Bangla baseline. VQA is measured in average Accuracy (\%) over each language, and Caption Quality (Cap.) is measured via LLM-as-a-Judge ($\times 100$).}
\label{tab:rq2_historically_linked_languages}
\end{table*}



\noindent \textbf{RQ1: How robust are multilingual VLMs to dialectal variation when understanding the same Bengali cultural content?} To answer RQ1, we compare standard Bangla with the five Bangla dialects, as shown in Figure~\ref{fig:rq1_dialect_robustness}, while keeping the underlying image fixed so that any gap reflects linguistic variation rather than visual difficulty. Overall, multilingual VLMs are \textbf{not fully robust to dialectal variation}. The drop is modest but consistent for VQA, and much larger for caption generation. For instance, \texttt{GPT-4.1-mini} drops from 61.22\% to 58.93\% in VQA from Bangla to Dialect Avg ($-2.29$ points), but its caption quality falls from 56.24 to 41.41 on LLM-Judge$\times 100$ ($-14.84$ points). A similar pattern holds for \texttt{Gemma3-27B}, which remains the strongest open model on this comparison: its VQA decreases only from 64.82\% to 63.14\% ($-1.68$ points), while caption quality drops from 53.90 to 43.35 ($-10.55$ points). These results show that dialectal robustness is substantially weaker for free-form generation than for answer-constrained reasoning.

The dialect penalty is also model-dependent, but no model is fully dialect-invariant. \texttt{Gemma3-12B} shows a $-2.90$-point VQA drop and a $-12.15$-point caption drop, while \texttt{Qwen2.5-VL-7B} drops by $-1.91$ VQA points and $-7.10$ caption points. Even when the VQA degradation is small, the captioning degradation remains substantial, indicating that models can often still identify the correct answer from options but struggle to generate culturally grounded descriptions when the input is phrased in dialect. One exception is \texttt{Qwen3-VL-8B}, whose caption score is nearly flat between Bangla and Dialect Avg (41.26 vs.\ 42.32), although its overall caption quality remains much lower than the strongest models. 

Taken together, these findings answer RQ1 clearly: \textbf{current multilingual VLMs show only limited robustness to Bangla dialect variation, and evaluation on standard Bangla alone would overestimate their true cultural understanding}. To confirm that these degradations are distinct from random noise, we conducted paired $t$-tests comparing Standard Bangla against the dialectal average, revealing statistically significant performance drops for both VQA accuracy ($p < 0.001$) and generative captioning ($p \ll 0.001$). Furthermore, while smaller domains (e.g., \textit{National Achievements}, \textit{Sports}) naturally exhibit wider 95\% confidence intervals ($\approx \pm 3.5\%$ to $4.5\%$), this expected variance does not alter the broader aggregated trends of dialectal degradation. Full statistical metrics and domain-wise confidence intervals are detailed in Appendix~\ref{app:statistical_significance}.

\noindent \textbf{RQ2. Do historically linked languages preserve Bengali cultural meaning better than standard cross-lingual translation would suggest?} To answer RQ2, we compare the four standard-language settings, \textit{Bangla}, \textit{English}, \textit{Hindi}, and \textit{Urdu}, to test whether historically linked languages preserve Bengali cultural meaning better than a simple translation baseline would suggest. Table~\ref{tab:rq2_historically_linked_languages} presents the comparative results alongside their relative differences (deltas) from the Standard Bangla baseline. The results show a mixed but informative pattern: \textbf{English remains the strongest overall transfer language}, but \textbf{Hindi and, to a lesser extent, Urdu preserve cultural meaning in caption generation to a remarkably high degree despite the cross-lingual shift}. Averaged across models, English achieves the best performance (59.44\% VQA; 59.69 LLM-Judge$\times 100$), followed by Bangla (57.37\%; 50.59), Hindi (54.04\%; 53.08), and Urdu (50.12\%; 50.40). 

Notably, Hindi's average caption quality (53.08) slightly exceeds Bangla's (50.6), even though its VQA is lower (54.04\% vs.\ 57.37\%). This divergence reveals a critical architectural nuance regarding how these models store cultural knowledge: global association versus fine-grained grounding. Because captioning relies heavily on global features and statistical co-occurrence, the models can recognize the broad cultural footprint of Bengali concepts and map them to fluid Hindi descriptions. However, VQA requires precise spatial reasoning and object-attribute binding. The distinct drop in VQA indicates a cross-lingual reasoning bottleneck; while the models possess surface-level cultural familiarity in historically linked languages, they struggle to parse localized visual queries and ground specific sub-components across linguistic boundaries. This pattern is especially visible in representative strong models. For \texttt{GPT-4.1-mini}, caption quality is identical between Bangla and Hindi (56.2), despite a noticeable VQA gap (61.2\% vs.\ 57.2\%). Similarly, for \texttt{Gemma3-27B}, Urdu caption quality (56.2) even exceeds its Bangla score (53.9), while its VQA drops from 64.8\% to 54.6\%. This suggests models are effectively exploiting memorized textual patterns during open-ended generation, whereas the strict, unforgiving nature of exact-match VQA exposes their lack of deep visual grounding in transferred languages. In short, the answer to RQ2 is \textbf{partially yes}: historically linked languages like Hindi and Urdu retain strong ``macro" cultural semantic associations—allowing for robust descriptive generation—but they suffer from a grounding bottleneck that prevents them from matching Bangla or English in structured, ``micro" cultural reasoning.

\begin{table}[t]
\centering
\scriptsize
\renewcommand{\arraystretch}{1.1}
\setlength{\tabcolsep}{4.0pt}
\begin{tabular}{clcccccc}
\toprule
\rowcolor{gray!20} \textbf{R} & \textbf{Domain} & \textbf{ZS} & \textbf{FS} & \textbf{CoT} & \textbf{Cap.} & \textbf{$f(d)$} & \textbf{Type} \\
\midrule
1 & Food       & 76.56 & 69.03 & 77.31 & 47.54 & 75.15  & $\circ$ \\
2 & Cult.      & 63.42 & 59.01 & 65.48 & 49.23 & 85.29  & $\triangle$ \\
3 & Nat. Achv. & 55.35 & 50.86 & 62.33 & 49.78 & 87.89  & $\blacktriangle$ \\
4 & Hist.      & 61.07 & 56.74 & 63.59 & 48.35 & 88.06  & $\triangle$ \\
5 & Pers.      & 58.67 & 57.42 & 62.28 & 43.45 & 94.27  & $\blacktriangle$ \\
6 & Sports     & 51.91 & 47.44 & 57.21 & 48.26 & 94.53  & $\blacktriangle$ \\
7 & Nature     & 52.29 & 53.05 & 52.71 & 50.58 & 96.37  & $\circ$ \\
8 & Pol.       & 45.16 & 44.37 & 48.11 & 44.95 & 106.94 & $\blacktriangle$ \\
9 & M\&M       & 33.63 & 31.93 & 36.27 & 43.12 & 120.61 & $\blacktriangle$ \\
\bottomrule
\end{tabular}
\caption{\textbf{Estimating domain-level bottlenecks.} Domains are ranked from easier to harder using the $f(d)$. Based on how domains need to be answered, we labeled them into three buckets: $\circ$ = visually concrete, $\triangle$ = mixed, $\blacktriangle$ = knowledge-intensive. R = Rank, ZS = Zero-shot VQA, FS = Few-shot VQA, CoT = CoT VQA, Cap. = Caption Quality, $f(d)$ = Difficulty Score.}
\label{tab:rq3_domain_difficulty}
\vspace{-3mm}
\end{table}

\noindent \textbf{RQ3. Is the main bottleneck in Bengali culture understanding visual grounding, linguistic variation, or missing cultural knowledge?} To answer RQ3, we average results by domain across all models and all language/dialect variants, and compare four signals: \textit{zero-shot VQA}, \textit{few-shot VQA}, \textit{CoT VQA}, and \textit{caption quality}. To make the ranking reproducible, we define a heuristic \textbf{difficulty score} function, $f(d)$, where $d$ represents a specific domain data point. This function measures how far the results are from perfect (100\%) by summing the error rate of the best-performing VQA setting and the error rate of caption generation:

\[
\small
f(d) = (100 - \max(\text{ZS}_d, \text{FS}_d, \text{CoT}_d)) + (100 - \text{Cap.}_d)
\]

Intuitively, this formulation measures the absolute gap from optimal performance. A domain is ranked as more difficult if it consistently exhibits lower accuracy despite few-shot or CoT assistance, alongside poor caption generation. Table~\ref{tab:rq3_domain_difficulty} suggests that \textbf{domain-level difficulty is strongly associated with knowledge demands, which may be a more significant factor than pure visual recognition.} Visually concrete domains such as Food are the easiest (difficulty 75.15), while culturally knowledge-intensive domains such as Media \& Movies and Politics are the hardest (120.61 and 106.95). This spread indicates models struggle when interpretation depends on background knowledge, named entities, or culturally specific context. We note $f(d)$ is a heuristic proxy; low performance in Politics could also reflect weak entity recognition or answer-option confusability rather than a clean "knowledge versus grounding" distinction \citep{hossain2026right}. Still, compared to much smaller language/dialect gaps, domain-level knowledge demands appear as the primary bottleneck. Moreover, CoT yields the largest gains in knowledge-heavy domains like National Achievements (+6.97) and Sports (+5.31) compared to visually easier ones like Food (+0.75), though overall performance remains far from perfect. Conversely, few-shot prompting generally degrades VQA accuracy across most domains, suggesting models struggle to effectively leverage in-context cultural examples. Caption quality follows a similar pattern, with Media \& Movies being the weakest (43.12). Taken together, these results support the hypothesis that models \textbf{often lack the specific cultural knowledge and fine-grained entity recognition needed to interpret what they see}.


\section{Error Analysis}
\label{sec:error_analysis}

To gain actionable insights beyond aggregate metrics, we conducted a systematic error analysis on a stratified random sample of 200 model generations, restricted exclusively to Standard Bangla (100 samples) and English (100 samples) to ensure culturally rigorous manual coding. Out of these, 94 instances exhibited notable errors, revealing that the primary driver of model failure is a fundamental lack of grounded local entity recognition (42.6\% of total failures), where models hallucinate unrelated or foreign concepts instead of local artifacts. The second most common mode is surface-level descriptive fallback (30.8\%), where models provide literal visual descriptions but miss the deeper cultural context. Societal and cultural biases, such as gender stereotyping based on traditional attire, account for an additional 14.9\% of failures. Evaluating these errors across languages reveals distinct failure pathways. When prompted in English, models are slightly more prone to entity misrecognition, suggesting that high-resource language generation pathways often overwrite localized visual grounding with Western-centric concepts. Conversely, when prompted in Bangla, models struggle heavily with language constraint violations (11.7\% of total failures). Despite strict source-language instructions, models frequently default back to English terminology during open-ended generation, exposing brittle cross-lingual alignment \citep{dipta2026ganitllm}. A comprehensive quantitative breakdown of these categories, exact language-wise error distributions, and detailed illustrative examples are provided in Appendix \ref{app:examples}.

\section{Conclusion}
In this paper, we introduce \textsc{BanglaVerse}, a multilingual and multidialectal benchmark evaluating VLM performance on Bengali culture. Through a curated dataset of $\sim$32.2K artifacts tested across four languages and five regional dialects, we demonstrate that current VLMs are highly sensitive to linguistic variation. The observed performance drops highlight that the core limitation of these models lies in insufficient cultural knowledge rather than pure visual grounding.

\clearpage
\newpage

\section*{Limitations}
\textsc{BanglaVerse} offers a culturally grounded benchmark that, by design, prioritizes depth of cultural representation across nine carefully curated domains. While the current dataset comprises 1,152 expert-annotated images, this scale was intentionally chosen to ensure high annotation quality and cultural authenticity. Future iterations can naturally expand both in volume and modality coverage as the benchmark evolves.

On the linguistic front, \textsc{BanglaVerse} covers five regional dialects, Barishal, Chittagong, Noakhali, Rangpur, and Sylhet, selected for their broad social and cultural salience. These five varieties should not, however, be read as representing the full linguistic and cultural diversity of Bengali-speaking communities, which extends to further varieties within Bangladesh as well as West Bengal, Tripura, and Assam's Barak Valley. Our results therefore characterize model behaviour on the varieties evaluated here and should not be generalized to Bengali linguistic variation as a whole. Extending coverage to less widely spoken varieties, including those of hill tribes, indigenous communities, and ethnic minorities, is a meaningful direction for future work.

Our annotation pipeline involved rigorous cross-verification by native speakers and cultural experts, which helps minimize subjective bias. As with any culturally rich dataset, certain local nuances may benefit from further community-driven refinement over time. Reassuringly, our systematic human evaluation did not surface notable inconsistencies in dialectal translations, lending confidence to the current annotations.

Finally, our evaluation covers a representative set of recent multilingual vision-language models. Given the rapid pace of model development, we view our results as establishing a reliable baseline for current capabilities, and we anticipate that \textsc{BanglaVerse} will serve as a continuing testbed as new architectures emerge. Related audits of Bengali--English models report cross-lingual polarity inversion and weaker alignment on the formal Bengali register in the text-only setting \citep{lia-roy-dipta-2026-cross}, suggesting the effects we observe are not confined to multimodal models. None of the evaluated systems is Bangla-specific; Bangla-adapted language models exist for text \citep{zehady-etal-2026-banglallama} but not yet for the vision--language setting.

\section*{Ethical Considerations}

The dataset was developed with careful attention to ethical standards. All images were collected from publicly available sources and contain no personally identifiable information. Annotations were manually cross-verified to minimize bias, ensure cultural sensitivity, and avoid harmful or offensive content. We also note a risk inherent to culturally grounded benchmarks. Because each item encodes a single ground-truth reading, treating our annotations as exhaustive accounts of Bengali culture could reinforce narrow or stereotyped representations. Model outputs carry a related risk: our error analysis finds political skew toward prominent ruling-party figures and gender stereotyping from visual attributes, echoing text-only findings that LLMs over-predict government-leaning stances in Bangla news \citep{lia-etal-2025-read}. Such outputs are failure cases, not properties of Bengali culture, and reproducing them without context could propagate the very biases the benchmark exposes. The dataset is released solely for research purposes to advance multimodal understanding in Bangla, and we encourage responsible use that respects cultural contexts and does not promote misuse or discrimination. 

\bibliography{custom}

@inproceedings{rafi2022deep,
  title={A deep learning-based bengali visual question answering system},
  author={Rafi, Mahamudul Hasan and Islam, Shifat and Labib, SM Hasan Imtiaz and Hasan, SM Sajid and Shah, Faisal Muhammad and Ahmed, Sifat},
  booktitle={2022 25th International Conference on Computer and Information Technology (ICCIT)},
  pages={114--119},
  year={2022},
  organization={IEEE}
}

@inproceedings{hasan-etal-2023-visual,
    title = "Visual Question Generation in {B}engali",
    author = "Hasan, Mahmud  and
      Islam, Labiba  and
      Ruma, Jannatul  and
      Mayeesha, Tasmiah  and
      Rahman, Rashedur",
    editor = "Gatt, Albert  and
      Gardent, Claire  and
      Cripwell, Liam  and
      Belz, Anya  and
      Borg, Claudia  and
      Erdem, Aykut  and
      Erdem, Erkut",
    booktitle = "Proceedings of the Workshop on Multimodal, Multilingual Natural Language Generation and Multilingual WebNLG Challenge (MM-NLG 2023)",
    month = sep,
    year = "2023",
    address = "Prague, Czech Republic",
    publisher = "Association for Computational Linguistics",
    url = "https://aclanthology.org/2023.mmnlg-1.2/",
    pages = "10--19"
}

@inproceedings{10.1145/3530190.3534837,
author = {Islam, S M Shahriar and Auntor, Riyad Ahsan and Islam, Minhajul and Anik, Mohammad Yousuf Hossain and Islam, A. B. M. Alim Al and Noor, Jannatun},
title = {Note: Towards Devising an Efficient VQA in the Bengali Language},
year = {2022},
isbn = {9781450393478},
publisher = {Association for Computing Machinery},
address = {New York, NY, USA},
url = {https://doi.org/10.1145/3530190.3534837},
doi = {10.1145/3530190.3534837},
booktitle = {Proceedings of the 5th ACM SIGCAS/SIGCHI Conference on Computing and Sustainable Societies},
pages = {632–637},
numpages = {6},
location = {Seattle, WA, USA},
series = {COMPASS '22}
}

@inproceedings{DBLP:conf/nips/RomeroLWGMPOVBJ24,
  author       = {David Romero and Chenyang Lyu and Haryo Akbarianto Wibowo and Santiago Góngora and Aishik Mandal and Sukannya Purkayastha and Jesús-Germán Ortiz-Barajas and Emilio Villa-Cueva and Jinheon Baek and Soyeong Jeong and Injy Hamed and Zheng Xin Yong and Zheng Wei Lim and Paula Mónica Silva and Jocelyn Dunstan and Mélanie Jouitteau and David Le Meur and Joan Nwatu and Ganzorig Batnasan and Munkh-Erdene Otgonbold and Munkhjargal Gochoo and Guido Ivetta and Luciana Benotti and Laura Alonso Alemany and Hernán Maina and Jiahui Geng and Tiago Timponi Torrent and Frederico Belcavello and Marcelo Viridiano and Jan Christian Blaise Cruz and Dan John Velasco and Oana Ignat and Zara Burzo and Chenxi Whitehouse and Artem Abzaliev and Teresa Clifford and Grainne Caulfield and Teresa Lynn and Christian Salamea Palacios and Vladimir Araujo and Yova Kementchedjhieva and Mihail Mihaylov and Israel Abebe Azime and Henok Biadglign Ademtew and Bontu Fufa Balcha and Naome A. Etori and David Ifeoluwa Adelani and Rada Mihalcea and Atnafu Lambebo Tonja and Maria Camila Buitrago Cabrera and Gisela Vallejo and Holy Lovenia and Ruochen Zhang and Marcos Estecha-Garitagoitia and Mario Rodríguez-Cantelar and Toqeer Ehsan and Rendi Chevi and Muhammad Farid Adilazuarda and Ryandito Diandaru and Samuel Cahyawijaya and Fajri Koto and Tatsuki Kuribayashi and Haiyue Song and Aditya Khandavally and Thanmay Jayakumar and Raj Dabre and Mohamed Fazli Mohamed Imam and Kumaranage Ravindu Yasas Nagasinghe and Alina Dragonetti and Luis Fernando D'Haro and Olivier Niyomugisha and Jay Gala and Pranjal A. Chitale and Fauzan Farooqui and Thamar Solorio and Alham Fikri Aji},
  title        = {CVQA: Culturally-diverse Multilingual Visual Question Answering Benchmark},
  booktitle    = {Advances in Neural Information Processing Systems (NeurIPS 2024) — Datasets and Benchmarks Track},
  year         = {2024},
  url          = {https://papers.nips.cc/paper_files/paper/2024/hash/1568882ba1a50316e87852542523739c-Abstract-Datasets_and_Benchmarks_Track.html},
  publisher    = {Curran Associates, Inc.}
}

@inproceedings{DBLP:journals/corr/abs-2410-14991,
  title={Chitrojera: A regionally relevant visual question answering dataset for bangla},
  author={Barua, Deeparghya Dutta and Sourove, Md Sakib Ul Rahman and Fahim, Md and Haider, Fabiha and Shifat, Fariha Tanjim and Rahman Adib, Md Tasmim and Uddin, Anam Borhan and Ishmam, Md Farhan and Alam, Md Farhad},
  booktitle={Joint European Conference on Machine Learning and Knowledge Discovery in Databases},
  pages={473--491},
  year={2025},
  organization={Springer}
}

@inproceedings{maji2025drishtikon,
    title = "{DRISHTIKON}: A Multimodal Multilingual Benchmark for Testing Language Models' Understanding on {I}ndian Culture",
    author = "Maji, Arijit  and
      Kumar, Raghvendra  and
      Ghosh, Akash  and
      Anushka  and
      Shah, Nemil  and
      Borah, Abhilekh  and
      Shah, Vanshika  and
      Mishra, Nishant  and
      Saha, Sriparna",
    editor = "Christodoulopoulos, Christos  and
      Chakraborty, Tanmoy  and
      Rose, Carolyn  and
      Peng, Violet",
    booktitle = "Proceedings of the 2025 Conference on Empirical Methods in Natural Language Processing",
    month = nov,
    year = "2025",
    address = "Suzhou, China",
    publisher = "Association for Computational Linguistics",
    url = "https://aclanthology.org/2025.emnlp-main.68/",
    doi = "10.18653/v1/2025.emnlp-main.68",
    pages = "1289--1313",
    ISBN = "979-8-89176-332-6"
}

@inproceedings{alwajih-etal-2025-palm,
    title = "Palm: A Culturally Inclusive and Linguistically Diverse Dataset for {A}rabic {LLM}s",
    author = "Alwajih, Fakhraddin  and
      El Mekki, Abdellah  and
      Magdy, Samar Mohamed  and
      Elmadany, AbdelRahim A.  and
      Nacar, Omer  and
      Nagoudi, El Moatez Billah  and
      Abdel-Salam, Reem  and
      Atwany, Hanin  and
      Nafea, Youssef  and
      Yahya, Abdulfattah Mohammed  and
      Alhamouri, Rahaf  and
      Alsayadi, Hamzah A.  and
      Zayed, Hiba  and
      Shatnawi, Sara  and
      Sibaee, Serry  and
      Ech-chammakhy, Yasir  and
      Al-Dhabyani, Walid  and
      Ali, Marwa Mohamed  and
      Jarraya, Imen  and
      El-Shangiti, Ahmed Oumar  and
      Alraeesi, Aisha  and
      AL-Ghrawi, Mohammed Anwar  and
      Al-Batati, Abdulrahman S.  and
      Mohamed, Elgizouli  and
      Elgindi, Noha Taha  and
      Saeed, Muhammed  and
      Atou, Houdaifa  and
      Yahia, Issam Ait  and
      Bouayad, Abdelhak  and
      Machrouh, Mohammed  and
      Makouar, Amal  and
      Alkawi, Dania  and
      Mohamed, Mukhtar  and
      Abdelfadil, Safaa Taher  and
      Ounnoughene, Amine Ziad  and
      Rouabhia, Anfel  and
      Assi, Rwaa  and
      Sorkatti, Ahmed  and
      Tourad, Mohamedou Cheikh  and
      Koubaa, Anis  and
      Berrada, Ismail  and
      Jarrar, Mustafa  and
      Shehata, Shady  and
      Abdul-Mageed, Muhammad",
    editor = "Che, Wanxiang  and
      Nabende, Joyce  and
      Shutova, Ekaterina  and
      Pilehvar, Mohammad Taher",
    booktitle = "Proceedings of the 63rd Annual Meeting of the Association for Computational Linguistics (Volume 1: Long Papers)",
    month = jul,
    year = "2025",
    address = "Vienna, Austria",
    publisher = "Association for Computational Linguistics",
    url = "https://aclanthology.org/2025.acl-long.1579/",
    doi = "10.18653/v1/2025.acl-long.1579",
    pages = "32871--32894",
    ISBN = "979-8-89176-251-0"
}

@article{10.1162/COLI.a.14,
    author = {Pawar, Siddhesh and Park, Junyeong and Jin, Jiho and Arora, Arnav and Myung, Junho and Yadav, Srishti and Haznitrama, Faiz Ghifari and Song, Inhwa and Oh, Alice and Augenstein, Isabelle},
    title = {Survey of Cultural Awareness in Language Models: Text and Beyond},
    journal = {Computational Linguistics},
    volume = {51},
    number = {3},
    pages = {907-1004},
    year = {2025},
    month = {09},
    issn = {0891-2017},
    doi = {10.1162/COLI.a.14},
    url = {https://doi.org/10.1162/COLI.a.14},
    eprint = {https://direct.mit.edu/coli/article-pdf/51/3/907/2523159/coli.a.14.pdf},
}

@inproceedings{
Zhang*2020BERTScore:,
title={BERTScore: Evaluating Text Generation with BERT},
author={Tianyi Zhang and Varsha Kishore* and Felix Wu* and Kilian Q. Weinberger and Yoav Artzi},
booktitle={International Conference on Learning Representations},
year={2020},
url={https://openreview.net/forum?id=SkeHuCVFDr}
}

@inproceedings{fahim2026banglaprotha,
  title={BanglaProtha: Evaluating Vision Language Models in Underrepresented Long-tail Cultural Contexts},
  author={Fahim, Md and Rahman, Md Sakib Ul and Rahman, Akm Moshiur and Ishmam, Md Farhan and Rahman, Md Tasmim and Shifat, Fariha Tanjim and Haider, Fabiha and Alam Bhuiyan, Md Farhad},
  booktitle={Proceedings of the IEEE/CVF Winter Conference on Applications of Computer Vision},
  pages={1159--1169},
  year={2026}
}

@article{mahi2025bangladial,
  title={BanglaDial: A merged and imbalanced text dataset for Bengali regional dialect analysis},
  author={Mahi, Mehraj Hossain and Khan, Anzir Rahman and Mojumdar, Mayen Uddin},
  journal={Data in Brief},
  pages={112200},
  year={2025},
  publisher={Elsevier}
}

@article{gu2024survey,
  title={A survey on llm-as-a-judge},
  author={Gu, Jiawei and Jiang, Xuhui and Shi, Zhichao and Tan, Hexiang and Zhai, Xuehao and Xu, Chengjin and Li, Wei and Shen, Yinghan and Ma, Shengjie and Liu, Honghao and others},
  journal={The Innovation},
  year={2024},
  publisher={Elsevier}
}

@inproceedings{nayak-etal-2024-benchmarking,
    title = "Benchmarking Vision Language Models for Cultural Understanding",
    author = "Nayak, Shravan  and
      Jain, Kanishk  and
      Awal, Rabiul  and
      Reddy, Siva  and
      Steenkiste, Sjoerd Van  and
      Hendricks, Lisa Anne  and
      Stanczak, Karolina  and
      Agrawal, Aishwarya",
    editor = "Al-Onaizan, Yaser  and
      Bansal, Mohit  and
      Chen, Yun-Nung",
    booktitle = "Proceedings of the 2024 Conference on Empirical Methods in Natural Language Processing",
    month = nov,
    year = "2024",
    address = "Miami, Florida, USA",
    publisher = "Association for Computational Linguistics",
    url = "https://aclanthology.org/2024.emnlp-main.329/",
    doi = "10.18653/v1/2024.emnlp-main.329",
    pages = "5769--5790"
}

@inproceedings{adilazuarda-etal-2024-towards,
    title = "Towards Measuring and Modeling ``Culture'' in {LLM}s: A Survey",
    author = "Adilazuarda, Muhammad Farid  and
      Mukherjee, Sagnik  and
      Lavania, Pradhyumna  and
      Singh, Siddhant Shivdutt  and
      Aji, Alham Fikri  and
      O{'}Neill, Jacki  and
      Modi, Ashutosh  and
      Choudhury, Monojit",
    editor = "Al-Onaizan, Yaser  and
      Bansal, Mohit  and
      Chen, Yun-Nung",
    booktitle = "Proceedings of the 2024 Conference on Empirical Methods in Natural Language Processing",
    month = nov,
    year = "2024",
    address = "Miami, Florida, USA",
    publisher = "Association for Computational Linguistics",
    url = "https://aclanthology.org/2024.emnlp-main.882/",
    doi = "10.18653/v1/2024.emnlp-main.882",
    pages = "15763--15784"
}

@article{faraz2025indicvisionbench,
  title={IndicVisionBench: Benchmarking Cultural and Multilingual Understanding in VLMs},
  author={Faraz, Ali and Khan, Shaharukh and Kolla, Raja and Patidar, Akshat and Goswami, Suranjan and Ravi, Abhinav and Khatri, Chandra and Agarwal, Shubham and others},
  journal={arXiv preprint arXiv:2511.04727},
  year={2025}
}

@article{liu2025culturevlm,
  title={Culturevlm: Characterizing and improving cultural understanding of vision-language models for over 100 countries},
  author={Liu, Shudong and Jin, Yiqiao and Li, Cheng and Wong, Derek F and Wen, Qingsong and Sun, Lichao and Chen, Haipeng and Xie, Xing and Wang, Jindong},
  journal={arXiv preprint arXiv:2501.01282},
  year={2025}
}

@article{shahen2019globalization,
  title={Globalization and Bangladesh: An analysis from cultural perspective},
  author={Shahen, Abu and Hossain, Bellal and Hossain, Md Bokul and Jahan, Most Nushrat},
  journal={IOSR Journal of Humanities and Social Science},
  volume={25},
  number={1},
  pages={32--41},
  year={2019}
}

@article{hasan2014standard,
  title={Standard Dialect Ideology in Bangladesh: A Field Study.},
  author={Hasan, Sheikh Mehedi and Rahaman, Adilur},
  journal={Language in India},
  volume={14},
  number={10},
  year={2014}
}

@article{karmaker2019dialectical,
  title={Dialectical and linguistic variations of Bangla sounds: Phonemic analysis},
  author={Karmaker, Protiva Rani},
  journal={Jagannath University Journal of Arts},
  volume={9},
  number={2},
  pages={125--130},
  year={2019}
}

@inproceedings{keleg2025llm,
  title={LLM Alignment for the Arabs: A Homogenous Culture or Diverse Ones?},
  author={Keleg, Amr},
  booktitle={The 3rd Workshop on Cross-Cultural Considerations in NLP (C3NLP 2025)},
  pages={1},
  year={2025}
}

@article{wei2022chain,
  title={Chain-of-thought prompting elicits reasoning in large language models},
  author={Wei, Jason and Wang, Xuezhi and Schuurmans, Dale and Bosma, Maarten and Xia, Fei and Chi, Ed and Le, Quoc V and Zhou, Denny and others},
  journal={Advances in neural information processing systems},
  volume={35},
  pages={24824--24837},
  year={2022}
}

@article{sakib2023comparing,
  title={Comparing the sociology of culture in Bangladesh and India: Similarities and differences in Bangladeshi and Indian cultures},
  author={Sakib, SM Nazmuz},
  journal={Simulacra},
  year={2023}
}

@inproceedings{e-sobhani-etal-2026-mathmist,
    title = "{M}ath{M}ist: A Parallel Multilingual Benchmark Dataset for Mathematical Problem Solving and Reasoning",
    author = "E Sobhani, Mahbub  and
      Sayeedi, Md. Faiyaz Abdullah  and
      Mohiuddin, Tasnim  and
      Islam, Md Mofijul  and
      Shatabda, Swakkhar",
    editor = "Demberg, Vera  and
      Inui, Kentaro  and
      Marquez, Llu{\'i}s",
    booktitle = "Findings of the {A}ssociation for {C}omputational {L}inguistics: {EACL} 2026",
    month = mar,
    year = "2026",
    address = "Rabat, Morocco",
    publisher = "Association for Computational Linguistics",
    url = "https://aclanthology.org/2026.findings-eacl.131/",
    doi = "10.18653/v1/2026.findings-eacl.131",
    pages = "2524--2550",
    ISBN = "979-8-89176-386-9"
}

@inproceedings{hasan-roy-dipta-2025-banglatalk,
    title = "{B}angla{T}alk: Towards Real-Time Speech Assistance for {B}engali Regional Dialects",
    author = "Hasan, Jakir  and
      Roy Dipta, Shubhashis",
    editor = "Alam, Firoj  and
      Kar, Sudipta  and
      Chowdhury, Shammur Absar  and
      Hassan, Naeemul  and
      Hoque, Enamul  and
      Laskar, Md Tahmid Rahman  and
      Mohiuddin, Tasnim  and
      Rony, Md Rashad Al Hasan",
    booktitle = "Proceedings of the Second Workshop on Bangla Language Processing (BLP-2025)",
    month = dec,
    year = "2025",
    address = "Mumbai, India",
    publisher = "Association for Computational Linguistics",
    url = "https://aclanthology.org/2025.banglalp-1.4/",
    doi = "10.18653/v1/2025.banglalp-1.4",
    pages = "44--60",
    ISBN = "979-8-89176-314-2"
}

@inproceedings{lia-etal-2025-read,
    title = "Read Between the Lines: A Benchmark for Uncovering Political Bias in {B}angla News Articles",
    author = "Lia, Nusrat Jahan  and
      Roy Dipta, Shubhashis  and
      Zehady, Abdullah Khan  and
      Islam, Naymul  and
      Chakraborty, Madhusodan  and
      Al Wasif, Abdullah",
    editor = "Alam, Firoj  and
      Kar, Sudipta  and
      Chowdhury, Shammur Absar  and
      Hassan, Naeemul  and
      Hoque, Enamul  and
      Laskar, Md Tahmid Rahman  and
      Mohiuddin, Tasnim  and
      Rony, Md Rashad Al Hasan",
    booktitle = "Proceedings of the Second Workshop on Bangla Language Processing (BLP-2025)",
    month = dec,
    year = "2025",
    address = "Mumbai, India",
    publisher = "Association for Computational Linguistics",
    url = "https://aclanthology.org/2025.banglalp-1.5/",
    doi = "10.18653/v1/2025.banglalp-1.5",
    pages = "61--79",
    ISBN = "979-8-89176-314-2"
}

@inproceedings{hossan-roy-dipta-2025-promptguard,
    title = "{P}rompt{G}uard at {BLP}-2025 Task 1: A Few-Shot Classification Framework Using Majority Voting and Keyword Similarity for {B}engali Hate Speech Detection",
    author = "Hossan, Rakib  and
      Roy Dipta, Shubhashis",
    editor = "Alam, Firoj  and
      Kar, Sudipta  and
      Chowdhury, Shammur Absar  and
      Hassan, Naeemul  and
      Hoque, Enamul  and
      Laskar, Md Tahmid Rahman  and
      Mohiuddin, Tasnim  and
      Rony, Md Rashad Al Hasan",
    booktitle = "Proceedings of the Second Workshop on Bangla Language Processing (BLP-2025)",
    month = dec,
    year = "2025",
    address = "Mumbai, India",
    publisher = "Association for Computational Linguistics",
    url = "https://aclanthology.org/2025.banglalp-1.35/",
    doi = "10.18653/v1/2025.banglalp-1.35",
    pages = "414--420",
    ISBN = "979-8-89176-314-2"
}

@inproceedings{zehady-etal-2026-banglallama,
    title = "{B}angla{L}lama: {LL}a{MA} for {B}angla Language",
    author = "Zehady, Abdullah Khan  and
      Roy Dipta, Shubhashis  and
      Islam, Naymul  and
      Al Mamun, Safi  and
      Karmaker, Santu",
    editor = "Hettiarachchi, Hansi  and
      Ranasinghe, Tharindu  and
      Plum, Alistair  and
      Rayson, Paul  and
      Mitkov, Ruslan  and
      Gaber, Mohamed  and
      Premasiri, Damith  and
      Tan, Fiona Anting  and
      Uyangodage, Lasitha",
    booktitle = "Proceedings of the Second Workshop on Language Models for Low-Resource Languages (LoResLM 2026)",
    month = mar,
    year = "2026",
    address = "Rabat, Morocco",
    publisher = "Association for Computational Linguistics",
    url = "https://aclanthology.org/2026.loreslm-1.7/",
    doi = "10.18653/v1/2026.loreslm-1.7",
    pages = "73--89",
    ISBN = "979-8-89176-377-7"
}

@inproceedings{roy-dipta-vallurupalli-2024-umbclu,
    title = "{UMBCLU} at {S}em{E}val-2024 Task 1: Semantic Textual Relatedness with and without machine translation",
    author = "Roy Dipta, Shubhashis  and
      Vallurupalli, Sai",
    editor = "Ojha, Atul Kr.  and
      Do{\u{g}}ru{\"o}z, A. Seza  and
      Tayyar Madabushi, Harish  and
      Da San Martino, Giovanni  and
      Rosenthal, Sara  and
      Ros{\'a}, Aiala",
    booktitle = "Proceedings of the 18th International Workshop on Semantic Evaluation (SemEval-2024)",
    month = jun,
    year = "2024",
    address = "Mexico City, Mexico",
    publisher = "Association for Computational Linguistics",
    url = "https://aclanthology.org/2024.semeval-1.195/",
    doi = "10.18653/v1/2024.semeval-1.195",
    pages = "1351--1357"
}

@inproceedings{roy-dipta-etal-2026-vc,
    title = "{VC}-{I}nspector: Advancing Reference-free Evaluation of Video Captions with Factual Analysis",
    author = "Roy Dipta, Shubhashis  and
      Wu, Tz-Ying  and
      Tripathi, Subarna",
    editor = "Liakata, Maria  and
      Moreira, Viviane P.  and
      Zhang, Jiajun  and
      Jurgens, David",
    booktitle = "Proceedings of the 64th Annual Meeting of the Association for Computational Linguistics (Volume 1: Long Papers)",
    month = jul,
    year = "2026",
    address = "San Diego, California, United States",
    publisher = "Association for Computational Linguistics",
    url = "https://aclanthology.org/2026.acl-long.1552/",
    doi = "10.18653/v1/2026.acl-long.1552",
    pages = "33657--33672",
    ISBN = "979-8-89176-390-6"
}

@inproceedings{lia-roy-dipta-2026-cross,
    title = "Cross-Lingual Sentiment Misalignment: Auditing Multilingual Language Models for Inversion Risk, Dialectal Representation, and Affective Stability",
    author = "Lia, Nusrat Jahan  and
      Roy Dipta, Shubhashis",
    editor = "Huang, Kaiyu  and
      Mo, Fengran  and
      Chen, Pinzhen  and
      Jiang, Meng",
    booktitle = "Proceedings of the 1st Workshop on Multilinguality in the Era of Large Language Models (MeLLM 2026)",
    month = jul,
    year = "2026",
    address = "San Diego, United States",
    publisher = "Association for Computational Linguistics",
    url = "https://aclanthology.org/2026.mellm-1.12/",
    doi = "10.18653/v1/2026.mellm-1.12",
    pages = "127--139",
    ISBN = "979-8-89176-430-9"
}

@inproceedings{roy-dipta-ferraro-2025-q2e,
    title = "{Q}2{E}: Query-to-Event Decomposition for Zero-Shot Multilingual Text-to-Video Retrieval",
    author = "Roy Dipta, Shubhashis  and
      Ferraro, Francis",
    editor = "Inui, Kentaro  and
      Sakti, Sakriani  and
      Wang, Haofen  and
      Wong, Derek F.  and
      Bhattacharyya, Pushpak  and
      Banerjee, Biplab  and
      Ekbal, Asif  and
      Chakraborty, Tanmoy  and
      Singh, Dhirendra Pratap",
    booktitle = "Proceedings of the 14th International Joint Conference on Natural Language Processing and the 4th Conference of the Asia-Pacific Chapter of the Association for Computational Linguistics",
    month = dec,
    year = "2025",
    address = "Mumbai, India",
    publisher = "The Asian Federation of Natural Language Processing and The Association for Computational Linguistics",
    url = "https://aclanthology.org/2025.ijcnlp-long.121/",
    doi = "10.18653/v1/2025.ijcnlp-long.121",
    pages = "2225--2245",
    ISBN = "979-8-89176-298-5"
}

@inproceedings{hasan2026banglaipa,
    title = "{B}angla{IPA}: Towards Robust Text-to-{IPA} Transcription with Contextual Rewriting in {B}engali",
    author = "Hasan, Jakir  and
      Datta, Shrestha  and
      Islam, Md Saiful  and
      Roy Dipta, Shubhashis  and
      Debnath, Ameya",
    editor = "Hettiarachchi, Hansi  and
      Ranasinghe, Tharindu  and
      Plum, Alistair  and
      Rayson, Paul  and
      Mitkov, Ruslan  and
      Gaber, Mohamed  and
      Premasiri, Damith  and
      Tan, Fiona Anting  and
      Uyangodage, Lasitha",
    booktitle = "Proceedings of the Second Workshop on Language Models for Low-Resource Languages ({L}o{R}es{LM} 2026)",
    month = mar,
    year = "2026",
    address = "Rabat, Morocco",
    publisher = "Association for Computational Linguistics",
    url = "https://aclanthology.org/2026.loreslm-1.12/",
    doi = "10.18653/v1/2026.loreslm-1.12",
    pages = "132--139",
    ISBN = "979-8-89176-377-7"
}

@inproceedings{dipta2026ganitllm,
    title = "{G}anit{LLM}: Difficulty-Aware {B}engali Mathematical Reasoning through Curriculum-{GRPO}",
    author = "Roy Dipta, Shubhashis  and
      Mahbub, Khairul  and
      Najjar, Nadia",
    editor = "Liakata, Maria  and
      Moreira, Viviane P.  and
      Zhang, Jiajun  and
      Jurgens, David",
    booktitle = "Findings of the {A}ssociation for {C}omputational {L}inguistics: {ACL} 2026",
    month = jul,
    year = "2026",
    address = "San Diego, California, United States",
    publisher = "Association for Computational Linguistics",
    url = "https://aclanthology.org/2026.findings-acl.1995/",
    doi = "10.18653/v1/2026.findings-acl.1995",
    pages = "40133--40148",
    ISBN = "979-8-89176-395-1"
}

@article{abdullah2026breaking,
  title={Breaking the Silence: A Dataset and Benchmark for Bangla Text-to-Gloss Translation},
  author={Abdullah, Sharif Mohammad and Paul, Abhijit and Dipta, Shubhashis Roy and Masud, Zarif and Rayana, Shebuti and Kabir, Ahmedul},
  journal={arXiv preprint arXiv:2504.02293},
  year={2026}
}

@article{nazi2026dagger,
  title={DAGGER: Distractor-Aware Graph Generation for Executable Reasoning in Math Problems},
  author={Nazi, Zabir Al and Dipta, Shubhashis Roy and Kar, Sudipta},
  journal={arXiv preprint arXiv:2601.06853},
  year={2026}
}

@article{hossain2026right,
  title={Right Knowledge, Wrong Answer: Characterizing Parametric Temporal Conflict in Open-Weight Language Models},
  author={Hossain, Elias and Saha, Sourav and Ornee, Tasfia Nuzhat and Jennifer, Sanjeda Sara and Biswas, Umesh Chandra and Dipta, Shubhashis Roy and Rana, Rajib and Yousefi, Niloofar},
  journal={arXiv preprint arXiv:2606.20959},
  year={2026}
}

@article{nazi2026omd,
  title={Omni-Modal Dissonance Benchmark: Systematically Breaking Modality Consensus to Probe Robustness and Calibrated Abstention},
  author={Nazi, Zabir Al and Dipta, Shubhashis Roy and Parvez, Md Rizwan},
  journal={arXiv preprint arXiv:2603.27187},
  year={2026}
}

\clearpage
\newpage

\appendix

\section{Additional Information on Data Collection}

\subsection{Data Sources and Licensing}
\label{app:data-sources}

To construct \textsc{BanglaVerse}, we aggregated data from a diverse set of online portals, encyclopedias, and printed literature. We strictly adhered to the copyright and licensing requirements of each respective source to ensure ethical data collection and usage.

Data sourced from open-collaboration platforms, such as Wikipedia, were utilized under the Creative Commons Attribution-ShareAlike (CC BY-SA 4.0) license. For proprietary sources, including news outlets, Banglapedia, and offline books, data was collected under the provisions of \textit{Fair Use} for non-commercial, academic research purposes. Specifically, we only extracted brief, factual snippets and low-resolution image artifacts strictly necessary for multimodal reasoning and cultural benchmark creation. None of the collected texts reproduce the creative or monetary value of the original works. The complete list of sources and their associated licensing terms is summarized in Table~\ref{tab:source_urls}.

\begin{table*}[t]
\centering
\small
\renewcommand{\arraystretch}{1.3}
\begin{tabular}{p{10cm} l}
\toprule
\rowcolor{gray!20} \textbf{Source Website URL / Book Category} & \textbf{License / Usage Rights} \\
\midrule
\rowcolor{blue!10} \multicolumn{2}{l}{\textbf{Online Sources}} \\
\midrule
\url{https://allbanglanewspapersbd.com/} & Copyrighted (Academic Fair Use) \\
\url{https://www.wikipedia.org/} & CC BY-SA 4.0 \\
\url{https://en.banglapedia.org/} & Copyrighted (Educational Fair Use) \\
\url{https://www.goodreads.com/shelf/show/history-bangladesh} & Copyrighted (Academic Fair Use) \\
\url{https://www.rokomari.com/book/category/1263/general-knowledge} & Copyrighted (Academic Fair Use) \\
\midrule
\rowcolor{blue!10} \multicolumn{2}{l}{\textbf{Offline Books (Printed Literature)}} \\
\midrule
\href{https://www.rokomari.com/book/542142/sadharon-gyan-joykoli-gk}{{\fontspec{kalpurush.ttf} সাধারণ জ্ঞান জয়কলি জিকে}} & Copyrighted (Academic Fair Use) \\
\href{https://www.rokomari.com/book/544267/medi-gk-sadharon-gyan}{{\fontspec{kalpurush.ttf} মেডি জিকে সাধারণ জ্ঞান}} & Copyrighted (Academic Fair Use) \\
\href{https://www.rokomari.com/book/464415/sadaron-gyan-chotoder-choloman-bissho}{{\fontspec{kalpurush.ttf} সাধারণ জ্ঞান ছোটদের চলমান বিশ্ব}} & Copyrighted (Academic Fair Use) \\
\href{https://www.rokomari.com/book/203465/sadharon-gan-bangladesh-o-antorjatik-bishoyaboli}{{\fontspec{kalpurush.ttf} সাধারণ জ্ঞান: বাংলাদেশ ও আন্তর্জাতিক বিষয়াবলি}} & Copyrighted (Academic Fair Use) \\
\href{https://www.rokomari.com/book/534406/global-affairs-saltamami-2025}{{\fontspec{kalpurush.ttf} গ্লোবাল অ্যাফেয়ার্স সালতামামি ২০২৫}} & Copyrighted (Academic Fair Use) \\
\href{https://www.rokomari.com/book/442855/gk-foundation-bangladesh-o-international}{{\fontspec{kalpurush.ttf} জিকে ফাউন্ডেশন বাংলাদেশ ও আন্তর্জাতিক বিষয়াবলি}} & Copyrighted (Academic Fair Use) \\
\href{https://www.rokomari.com/book/453231/professor-s-basic-gk-review}{{\fontspec{kalpurush.ttf} প্রফেসর’স বেসিক জিকে রিভিউ}} & Copyrighted (Academic Fair Use) \\
\href{https://www.rokomari.com/book/437409/current-news-sadharon-gaan}{{\fontspec{kalpurush.ttf} কারেন্ট নিউজ সাধারণ জ্ঞান}} & Copyrighted (Academic Fair Use) \\
\bottomrule
\end{tabular}
\caption{List of source websites and specific books utilized for \textsc{BanglaVerse} data collection, along with their respective licensing or fair use designations.}
\label{tab:source_urls}
\end{table*}

\subsection{Data Curation Guidelines and Taxonomy}
\label{app:curation_guidelines}

To distill our initial pool of 2,019 candidate instances into a high-quality, culturally representative benchmark, we followed a strict set of taxonomy-driven guidelines. The overarching objective of the curation process was to ensure that every selected image and its corresponding annotation genuinely reflected the lived reality, heritage, and multifaceted identity of the Bengali people. 

During curation, annotators evaluated each instance against the following domain-specific inclusion criteria (our cultural taxonomy):

\begin{itemize}
    \item \textbf{Culture:} Instances must depict traditional customs, festivals (e.g., Pohela Boishakh, Eid, Durga Puja), indigenous arts and crafts (e.g., Rickshaw painting, Jamdani weaving, Nakshi Kantha), or traditional attire. 
    \item \textbf{Food:} Instances should capture authentic Bengali culinary traditions. This includes staple dishes (e.g., rice and Hilsa fish), traditional desserts and sweets (e.g., Pitha, Rosogolla), street food culture (e.g., Fuchka, Jhalmuri), and regional culinary specialties.
    \item \textbf{History:} Data must represent significant historical milestones and heritage sites. Key focus areas include the 1952 Language Movement, the 1971 Liberation War, ancient archaeological sites (e.g., Mahasthangarh, Somapura Mahavihara), and historical monuments (e.g., Lalbagh Fort, Shaheed Minar).
    \item \textbf{Media \& Movies:} Instances must cover iconic elements of Bengali entertainment and arts. This encompasses legendary cinema, prominent directors and actors, traditional music forms (e.g., Baul music), indigenous musical instruments (e.g., Ektara), and notable television or theatrical productions.
    \item \textbf{National Achievements:} Data in this category is strictly factual and event-based, requiring the depiction of highly notable milestones. Examples include major infrastructural mega-projects (e.g., the Padma Multipurpose Bridge, Karnaphuli Tunnel), technological advancements (e.g., Bangabandhu-1 Satellite), and international recognitions.
    \item \textbf{Nature:} Instances should highlight the unique geographical and ecological landscape of the region. This includes the Sundarbans mangrove forest, Cox's Bazar beach, Sylhet tea gardens, riverine landscapes, and native flora and fauna (e.g., the Royal Bengal Tiger, Water Lily/Shapla).
    \item \textbf{Personalities:} Data must feature historically or culturally significant figures who have shaped Bengali society. This includes national poets and literary giants (e.g., Kazi Nazrul Islam, Rabindranath Tagore, Begum Rokeya), scientists, social reformers, and scholars.
    \item \textbf{Politics:} Instances should capture the political history and democratic evolution of the region, including influential political leaders, historic gatherings (e.g., the historic 7th March Speech), and key political institutions like the Jatiya Sangsad Bhaban (National Parliament House).
    \item \textbf{Sports:} Instances should depict both traditional rural games (e.g., Kabaddi/Ha-du-du, Nouka Baich/Boat Racing, Lathi Khela) and culturally significant moments in modern international sports (e.g., landmark victories in international Cricket or Football).
\end{itemize}

\paragraph{Quality Checks:}
Beyond fitting into the taxonomy, each curated instance had to pass secondary checks for visual clarity, factual accuracy of the text, and freedom from ambiguity. The native curators leveraged their inherent cultural context to discard instances where the visual cues were geographically ambiguous (e.g., generic South Asian content that lacks distinct Bengali characteristics) or where the textual context misrepresented the cultural significance of the image.

\section{Detailed Inter-Annotator Agreement Analysis}
\label{app:inter_annotator_details}

Table~\ref{tab:domain_kappa} provides the full domain-wise breakdown of the Cohen's kappa scores between the paired annotators (Annotators 1 through 4). The analysis demonstrates consistent reliability across all varied cultural domains, which averages out to the reported overall $\kappa = 0.87$.

\begin{table*}[h!]
\centering
\small
\begin{tabular}{@{}llcc@{}}
\toprule
\rowcolor{gray!20} \textbf{Domain}                & \textbf{Assigned Annotators} & \textbf{Cohen's $\kappa$} & \textbf{Sample Size ($n$)} \\ \midrule
\textbf{Culture}               & Annotator 1 \& Annotator 2   & 0.86                      & 114                        \\
\textbf{Food}                  & Annotator 2 \& Annotator 3   & 0.91                      & 150                        \\
\textbf{History}               & Annotator 3 \& Annotator 4   & 0.85                      & 150                        \\
\textbf{Media \& Movies}       & Annotator 4 \& Annotator 1   & 0.84                      & 150                        \\
\textbf{National Achievements} & Annotator 1 \& Annotator 3   & 0.88                      & 75                         \\
\textbf{Nature}                & Annotator 2 \& Annotator 4   & 0.89                      & 150                        \\
\textbf{Personalities}         & Annotator 3 \& Annotator 1   & 0.87                      & 150                        \\
\textbf{Politics}              & Annotator 4 \& Annotator 2   & 0.85                      & 150                        \\
\textbf{Sports}                & Annotator 1 \& Annotator 4   & 0.88                      & 63                         \\ \midrule
\rowcolor{blue!10} \textbf{Overall Average}       & -                            & \textbf{0.87}             & \textbf{Total: 1,152}      \\ \bottomrule
\end{tabular}
\caption{Domain-wise Inter-Annotator Agreement (Cohen's $\kappa$)}
\label{tab:domain_kappa}
\end{table*}

\section{Details of the Fine-Tuning Model}
\label{app:fine-tuning}

To systematically generate multidialectal variants of the \textsc{BanglaVerse} benchmark, we fine-tuned the \texttt{Qwen2.5-3B-Instruct} large language model. The model was specifically adapted to translate standard Bangla textual artifacts into five distinct regional dialects (Barishal, Chittagong, Noakhali, Rangpur, and Sylhet) and to natively handle dialect-based conversational generation.

\subsection{Dataset Preparation}
The fine-tuning dataset was derived from the BanglaDial corpus. To establish a robust evaluation framework, we randomly partitioned the dataset into an 80\% training split and a 20\% held-out test split, as the original corpus does not provide predefined partitions. The data was structured into two primary instruction formats to teach the model both direct translation and context-aware generation. 

\textbf{Task 1: Standard-to-Dialect Translation.} For caption and direct context translation, the model was fed pairs of standard Bengali and their dialectal equivalents using the following structured prompt:
\begin{tcolorbox}[enhanced,
  colback=blue!5!white,
  colframe=blue!75!black,
  fonttitle=\bfseries, 
  breakable]
\begin{verbatim}
Translate the following standard
Bengali to [Dialect] dialect:
Standard Bengali: [Standard Text]
[Dialect] dialect: [Target Dialect 
Text]
\end{verbatim}
\end{tcolorbox}

\textbf{Task 2: Native Dialect Conversation.} To ensure the model could process and answer questions formatted in native dialects (essential for the VQA task), we formulated conversational QA pairs:

\begin{tcolorbox}[enhanced,
  colback=blue!5!white,
  colframe=blue!75!black,
  fonttitle=\bfseries, 
  breakable]
\begin{verbatim}
User: [Question in Dialect]
Assistant: [Answer in Dialect]
\end{verbatim}
\end{tcolorbox}

During tokenization, inputs and target outputs were concatenated, and the causal language modeling objective was applied across the sequence. Tokenized sequences were capped at a maximum length of 512 tokens.

\subsection{Memory Optimization and PEFT Configuration}
Given the substantial parameter count of the base model, we utilized Low-Rank Adaptation (LoRA) alongside several memory-saving techniques to stabilize training on standard GPU infrastructure. 

Prior to applying LoRA, we disabled the KV-cache (\texttt{use\_cache=False}) and explicitly enabled gradient checkpointing to significantly reduce the VRAM footprint. LoRA adapters were subsequently injected into all primary linear projection layers within the transformer blocks to maximize adaptation capability. The target modules included the attention matrices (\texttt{q\_proj}, \texttt{k\_proj}, \texttt{v\_proj}, \texttt{o\_proj}) and the feed-forward network gates (\texttt{gate\_proj}, \texttt{up\_proj}, \texttt{down\_proj}). The LoRA rank ($r$) was set to 16 with a scaling factor ($\alpha$) of 32 and a dropout rate of 0.1.

\subsection{Training Dynamics and Hyperparameters}
Training was executed using mixed precision (FP16). To mitigate memory constraints while maintaining a stable gradient update, we utilized a micro-batch size of 1 per device, compensated by 8 gradient accumulation steps. 

To further optimize training speed, we enabled sequence length grouping (\texttt{group\_by\_length=True}), which minimizes the amount of padding required per batch. The data collator was configured to pad sequences to a multiple of 8, optimizing tensor core utilization on the GPU. The comprehensive hyperparameter configuration is summarized in Table \ref{tab:hyperparams}.

\begin{table}[t]
\centering
\small
\setlength{\tabcolsep}{4pt}
\begin{tabular}{@{}lc@{}}
\toprule
\rowcolor{gray!20} \textbf{Hyperparameter} & \textbf{Value} \\
\midrule
Base Architecture & \texttt{Qwen2.5-3B-Instruct} \\
Task Type & Causal LM \\
Training Epochs & 3 \\
Max Sequence Length & 512 \\
Per-Device Batch Size & 1 \\
Gradient Accumulation Steps & 8 \\
Learning Rate & $5 \times 10^{-5}$ \\
Warmup Steps & 500 \\
Precision & FP16 \\
LoRA Rank ($r$) & 16 \\
LoRA Alpha ($\alpha$) & 32 \\
LoRA Target Modules & All linear proj. \\
LoRA Dropout & 0.1 \\
Sequence Grouping & Enabled \\
\bottomrule
\end{tabular}
\caption{Hyperparameters and configurations used for fine-tuning the dialect generation model.}
\label{tab:hyperparams}
\end{table}


\subsection{Convergence and Inference Generation}
The model converged successfully over the course of 3 epochs, completing exactly 8,600 global training steps. Logging was captured every 50 steps, with checkpoints saved every 1,000 steps. At the final training step, the model achieved a training loss of \textbf{0.4528}, with a gradient norm of 1.8488 and a decayed learning rate of approximately $1.41 \times 10^{-7}$. For the generation of the final dialectal artifacts used in the benchmark, the fine-tuned model was deployed with a temperature of 0.7 and sampling enabled (\texttt{do\_sample=True}) to ensure fluent, naturally varying dialectal structures, generating up to 50 new tokens per artifact.

\begin{table*}[t]
\centering
\small
\begin{tabular}{lcccccc}
\toprule
\rowcolor{gray!20} \textbf{Dialect} & \textbf{N} & \textbf{BLEU} & \textbf{ChrF} & \textbf{WER} $\downarrow$ & \textbf{BS-L3Cube F1} & \textbf{Gemini Em. Sim.} \\
\midrule
\textbf{Barishal} & 248 & 39.85 & 63.91 & 48.15 & 0.832 & 0.978 \\
\textbf{Chittagong} & 248 & 22.14 & 43.12 & 67.85 & 0.715 & 0.964 \\
\textbf{Rangpur} & 247 & 41.25 & 68.42 & 42.18 & 0.871 & 0.986 \\
\textbf{Noakhali} & 247 & 24.12 & 49.88 & 59.10 & 0.738 & 0.965 \\
\textbf{Sylhet} & 248 & 23.45 & 47.53 & 61.76 & 0.779 & 0.971 \\
\midrule
\rowcolor{blue!10} \textbf{Average} & \textbf{--} & \textbf{30.16} & \textbf{54.57} & \textbf{55.81} & \textbf{0.787} & \textbf{0.973} \\
\bottomrule
\end{tabular}
\caption{Intrinsic Evaluation of the Fine-Tuned Dialect Generation Model on the BanglaDial Test Split. WER indicates Word Error Rate (lower is better). \textit{Note: BS-L3Cube F1 uses the L3Cube Bengali sentence-similarity SBERT model. Gemini Em. Sim. score is calculated using the gemini-embedding-001 model.}}
\label{tab:intrinsic-eval}
\end{table*}

\subsection{Intrinsic Evaluation Details}

To rigorously validate the translation quality of our fine-tuned \texttt{Qwen2.5-3B-Instruct} model, we conducted an intrinsic evaluation of the model weights against the held-out test split of the BanglaDial dataset. 

The results, presented in Table~\ref{tab:intrinsic-eval}, demonstrate strong performance across the dialects, particularly in terms of character-level n-gram matching (ChrF) and semantic preservation (Gemini Em. Sim. and BS-L3Cube F1). As expected due to varying degrees of linguistic distance from standard Bangla, dialects like Rangpur and Barishal show higher syntactic overlap, while all linguistic variants maintain exceptional semantic fidelity throughout the translation process.

While socio-politically categorized as dialects, varieties like Chittagong and Sylhet are morpho-syntactically and lexically highly distinct from Standard Bangla, to the point where linguists often classify them as separate languages. The high Word Error Rate (WER) and low BLEU scores directly reflect this profound linguistic divergence (lack of exact surface token overlap) rather than poor generation quality. This is precisely why we paired automated metrics with native human evaluators, who verified that the generated text accurately captured the unique regional phrasings. 

\section{Human Evaluation Details}
\label{app:human-eval}

To ensure the multilingual and multidialectal artifacts in \textsc{BanglaVerse} maintained high semantic fidelity and cultural authenticity, we conducted a targeted human evaluation. This appendix provides a detailed breakdown of the evaluation methodology, the score distribution, and a domain-wise analysis.

\subsection{Evaluation Setup and Guidelines}
We systematically sampled 10 instances (comprising both the generated caption and the VQA pair) from each of the 9 domains across all 9 linguistic variants (Standard Bangla, 3 translated languages, and 5 Bangla dialects). This yielded 90 samples per linguistic variant, resulting in a comprehensive total of 810 manually evaluated instances. Including Standard Bangla in this phase served as a baseline control to ensure the manually curated source text maintained the highest possible quality.

The evaluation was conducted by two distinct groups. Three of the authors evaluated the Standard Bangla, English, Hindi, and Urdu texts using a back-translation and cross-referencing approach to verify semantic and cultural preservation, since the benefit of machine translation into English is known to vary by language for low-resource Asian and African languages \citep{roy-dipta-vallurupalli-2024-umbclu}. Concurrently, five undergraduate students—selected specifically because each is a native speaker of one of the five target dialects (Barishal, Chittagong, Noakhali, Rangpur, and Sylhet)—evaluated the dialectal generations to ensure naturalness and regional authenticity. Evaluators were instructed to rate each instance using a strict 3-point scale, defined in Table~\ref{tab:human-eval-scale}.

\begin{table}[hbt!]
\centering
\small
\setlength{\tabcolsep}{4pt}
\renewcommand{\arraystretch}{1.2}
\begin{tabular}{@{}c>{\raggedright\arraybackslash}p{1.5cm}p{4.5cm}@{}}
\toprule
\rowcolor{gray!20} \textbf{Score} & \textbf{Rating} & \textbf{Description} \\
\midrule
\textbf{0} & Inaccurate & The translation alters the core meaning of the original Bangla, hallucinates information, or fails to capture the target dialect's vocabulary and grammar. \\
\textbf{1} & Acceptable & The core meaning is preserved, but the phrasing is slightly unnatural, or a cultural entity was translated too literally, requiring minor manual correction. \\
\textbf{2} & Flawless & The translation is natural, grammatically correct, culturally accurate, and (for dialects) reads as a native speaker would converse. \\
\bottomrule
\end{tabular}
\caption{The 3-point grading scale used by human annotators to evaluate the generated artifacts. Score 1 denotes output acceptable with minor revisions.}
\label{tab:human-eval-scale}
\end{table}


\subsection{Score Distribution and Domain Analysis}

Each of the 810 sampled instances was independently reviewed. The vast majority of the generated artifacts received perfect scores, validating the effectiveness of our LLM translation pipeline and our fine-tuned \texttt{Qwen2.5-3B-Instruct} dialect model. A small expert-annotated core expanded with model-generated data and then re-validated by native experts has proven effective for other low-resource Bangla tasks as well \citep{abdullah2026breaking}. The absolute breakdown of the 810 evaluated samples is presented in Table~\ref{tab:human-eval-scores}.

\begin{table}[hbt!]
\centering
\small
\begin{tabular}{lcc}
\toprule
\rowcolor{gray!20} \textbf{Score Rating} & \textbf{Count} & \textbf{Percentage} \\
\midrule
2 (Flawless) & 750 & 92.59\% \\
1 (Acceptable) & 45 & 5.56\% \\
0 (Inaccurate) & 15 & 1.85\% \\
\midrule
\rowcolor{blue!10} \textbf{Total Evaluated Instances} & \textbf{810} & \textbf{100\%} \\
\bottomrule
\end{tabular}
\caption{Overall distribution of human evaluation scores across 810 randomly sampled artifacts.}
\label{tab:human-eval-scores}
\end{table}

To provide deeper insight into model performance across different cultural contexts, we also provide a domain-wise breakdown of the evaluation results in Table~\ref{tab:domain-wise-human}. The results indicate consistent high performance across all categories, with domains like \textit{Nature} and \textit{Food} proving particularly robust during cross-lingual and dialectal translation. Slight drops in flawless scores were observed in visually complex domains such as \textit{Media \& Movies}, where translating highly specific pop-culture terminology occasionally required minor phrasing revisions.

\begin{table*}[hbt!]
\centering
\small
\begin{tabular}{lccccr}
\toprule
\rowcolor{gray!20} \textbf{Cultural Domain} & \textbf{Total Samples} & \textbf{Score 2} & \textbf{Score 1} & \textbf{Score 0} & \textbf{\% Flawless} \\
\midrule
Culture & 90 & 83 & 5 & 2 & 92.2\% \\
Food & 90 & 85 & 4 & 1 & 94.4\% \\
History & 90 & 82 & 6 & 2 & 91.1\% \\
Media \& Movies & 90 & 80 & 7 & 3 & 88.9\% \\
National Achievements & 90 & 84 & 5 & 1 & 93.3\% \\
Nature & 90 & 86 & 3 & 1 & 95.6\% \\
Personalities & 90 & 82 & 6 & 2 & 91.1\% \\
Politics & 90 & 83 & 5 & 2 & 92.2\% \\
Sports & 90 & 85 & 4 & 1 & 94.4\% \\
\midrule
\rowcolor{blue!10} \textbf{Total} & \textbf{810} & \textbf{750} & \textbf{45} & \textbf{15} & \textbf{92.6\%} \\
\bottomrule
\end{tabular}
\caption{Domain-wise breakdown of human evaluation scores. Each domain contains 10 samples evaluated across all 9 linguistic variants.}
\label{tab:domain-wise-human}
\end{table*}

\section{Prompting Strategies: Templates and Analysis}
\label{app:prompt}

\subsection{Prompt Templates}

\begin{tcolorbox}[colback=blue!5!white, colframe=blue!75!black, title=Zero-shot Prompt: Caption, breakable]
\small
\ttfamily
You are an assistant who generates short, fluent captions in [SOURCE LANGUAGE] only.

Look carefully at the given image and write exactly one meaningful sentence describing it.

Do not use any [OTHER LANGUAGE] words, do not add extra explanations, labels, or quotes. Your entire output must be only the Bangla caption as plain text.  
\end{tcolorbox}

\begin{tcolorbox}[colback=blue!5!white, colframe=blue!75!black, title=Few-shot Prompt: Caption, breakable]
\small
\ttfamily
You are an assistant who generates short, fluent captions in [SOURCE LANGUAGE] only.  

Here are three example captions:  

Example 1:  
Image: \{example\_image\_1\}  
\{example\_caption\_1\}  

Example 2:  
Image: \{example\_image\_2\}  
\{example\_caption\_2\}  

Example 3:  
Image: \{example\_image\_3\}  
\{example\_caption\_3\}  

Now, generate a caption for the following image:  
Image: \{image\_path\}  

Write exactly one meaningful [SOURCE LANGUAGE] sentence.  
Do not use any [OTHER LANGUAGE] words, do not add extra explanations, labels, or quotes.  
Your entire output must be only the [SOURCE LANGUAGE] caption as plain text.  
\end{tcolorbox}

\begin{tcolorbox}[colback=yellow!5!white, colframe=yellow!75!black, coltitle=black!50!black, title=Zero-shot Prompt: VQA, breakable]
\small
\ttfamily
You are an AI assistant that answers visual multiple-choice questions in [SOURCE LANGUAGE].  

Task:  

1. Look carefully at the given image: \{image\_path\}  

2. Read the question: \{question\}  

3. Review the provided answer choices: \{options\}  

4. Select the single most accurate answer.  

Response Rules:  

- The index must be the programming list index (starting from 0).  

- Respond ONLY with the exact format below.  

- Use [SOURCE LANGUAGE] text for the answer option.  

- Do NOT add explanations, extra words, reasoning steps, or anything outside the specified format.  

- Follow this exact structure:  

Index: <option\_index>, Answer: "<option\_text\_in\_[SOURCE LANGUAGE]>"  
\end{tcolorbox}

\begin{tcolorbox}[colback=yellow!5!white, colframe=yellow!75!black, coltitle=black!50!black, title=Few-shot Prompt: VQA, breakable]
\small
\ttfamily
You are an AI assistant that answers visual multiple-choice questions in [SOURCE LANGUAGE].  

Task:  

1. Look carefully at the given image: \{image\_path\}  

2. Read the question: \{question\}  

3. Review the provided answer choices: \{options\}  

4. Select the single most accurate answer.  

Response Rules:  

- The index must be the programming list index (starting from 0).  

- Respond ONLY with the exact format below.  

- Use [SOURCE LANGUAGE] text for the answer option.  

- Do NOT add explanations, extra words, reasoning steps, or anything outside the specified format.  

- Follow this exact structure:  
Index: <option\_index>, Answer: "<option\_text\_in\_[SOURCE LANGUAGE]>"  

Here are three example QA pairs:  

Example 1:  
Image: \{example\_image\_1\}  
Question: \{example\_question\_1\}  
Options: \{example\_options\_1\}  
Answer: Index: \{example\_index\_1\}, Answer: "\{example\_answer\_1\}"  

Example 2:  
Image: \{example\_image\_2\}  
Question: \{example\_question\_2\}  
Options: \{example\_options\_2\}  
Answer: Index: \{example\_index\_2\}, Answer: "\{example\_answer\_2\}"  

Example 3:  
Image: \{example\_image\_3\}  
Question: \{example\_question\_3\}  
Options: \{example\_options\_3\}  
Answer: Index: \{example\_index\_3\}, Answer: "\{example\_answer\_3\}"  

Now, answer for the given image.  
Image: \{image\_path\}  
Question: \{question\}  
Options: \{options\}  
Answer:  
\end{tcolorbox}

\begin{tcolorbox}[colback=yellow!5!white, colframe=yellow!75!black, coltitle=black!50!black, title=CoT Prompt: VQA, breakable]
\small
\ttfamily
You are an AI assistant that answers visual multiple-choice questions in [SOURCE LANGUAGE].  

Task:  

1. Look carefully at the given image: \{image\_path\}  

2. Read the question: \{question\}  

3. Review the provided answer choices: \{options\}  

4. Select the single most accurate answer.  

Response Rules:  

- The index must be the programming list index (starting from 0).  

- Use [SOURCE LANGUAGE] text for the answer option.  

- In Reasoning\_En, write step-by-step reasoning in English — break down the solution logically:  
  
  Step 1: Describe key visual observations.  
  
  Step 2: Match observations to relevant answer choices.  
  
  Step 3: Eliminate incorrect choices with brief justification.  
  
  Step 4: Conclude why the final choice is correct.  

- Be clear, concise, and factual (avoid overly long explanations).  

- Follow this exact response format:  

Reasoning\_En:  

Step 1: <your\_observations>  

Step 2: <your\_matching\_logic>  

Step 3: <your\_elimination\_of 
\_wrong\_options>  

Step 4: <your\_final\_choice 
\_reasoning>  

Final Answer: Index: <option\_index>, Answer: "<option\_text\_in\_[SOURCE LANGUAGE]>"  
\end{tcolorbox}

\begin{tcolorbox}[colback=gray!5!white, colframe=gray!75!black, title=LLM-as-a-Judge Prompt, breakable]
\small
\ttfamily
You are a highly skilled and impartial caption evaluator. Your task is to carefully compare a generated caption with a reference caption and score it according to the following dimensions:  

1. Relevance (0–1): How well does the generated caption describe the main objects, actions, and context of the reference caption? Reward high semantic overlap and penalize missing or hallucinated details.  

2. Clarity (0–1): Is the caption grammatically correct, well-structured, and easy to read?  

3. Conciseness (0–1): Is the caption free of redundancy, filler words, or unnecessary complexity while still conveying the full meaning?  

4. Creativity (0–1): Does the caption show originality or an engaging phrasing, rather than being overly generic?  

After scoring each dimension, compute an Overall (0–1) score that reflects the holistic quality of the generated caption, giving slightly higher weight to Relevance and Clarity.  

Your response must strictly follow this JSON-like structure:  

Relevance: [float between 0 and 1]  

Clarity: [float between 0 and 1]  

Conciseness: [float between 0 and 1]  

Creativity: [float between 0 and 1]  

Overall: [float between 0 and 1]  

Explanation:  
[Concise explanation: mention key strengths, weaknesses, and reasoning for the scores.]  

Reference Caption: \{reference\_caption\}  

Generated Caption: \{generated\_caption\}  
\end{tcolorbox}

\begin{figure*}[t]
    \centering
    \includegraphics[width=\textwidth]{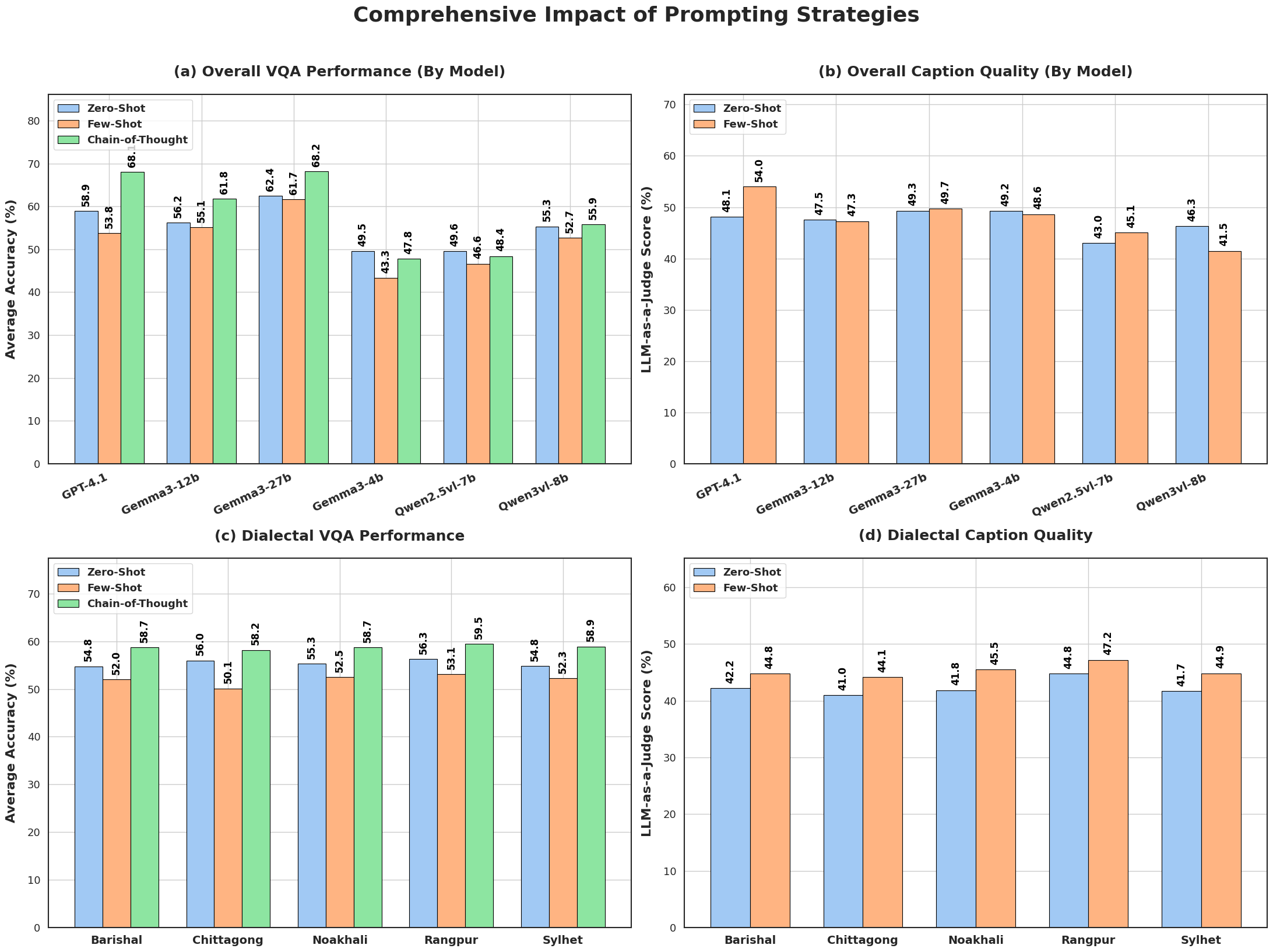}
    \caption{\textbf{Impact of Prompting Strategies on Culturally Grounded Tasks.} A unified comparison of Zero-Shot, Few-Shot, and Chain-of-Thought (CoT) prompting. \textbf{(Top Row):} (a) Overall performance across all 9 linguistic variants and domains by evaluated models. CoT consistently enhances VQA (b) while Few-Shot degrades it, though Few-Shot slightly improves stylistic caption generation. \textbf{(Bottom Row):} (c) Performance specifically isolated across the five regional Bangla dialects. Dialects universally mirror the overall trend, with CoT enhancing discriminative tasks (d) and Few-Shot aiding in generating dialectal captions.}
    \label{fig:prompting_strategies}
\end{figure*}

\subsection{Additional Analysis}
\label{app:prompting-rq}

\textbf{RQ4. How do different prompting strategies impact culturally grounded multimodal tasks?}

\paragraph{Impact of Prompting Strategies Across Models.} We investigate how large vision-language models respond to three distinct prompting techniques: zero-shot, few-shot (in-context learning), and Chain-of-Thought (CoT). To isolate the effect of these strategies, we aggregate performance across all six evaluated models, nine cultural domains, and all nine linguistic variants (four standard languages and five Bangla dialects). The results reveal a striking divergence between reasoning-based prompting and example-based prompting in cultural contexts.

\begin{table*}[t]
\centering
\small
\begin{tabular}{lcccc}
\toprule
\rowcolor{gray!20} \textbf{Language / Dialect} & \textbf{Average Score} & \textbf{Average Score} & \textbf{Pearson} & \textbf{Spearman Rank} \\ 
\rowcolor{gray!20} & \textbf{(Gemini)} & \textbf{(GPT-5)} & \textbf{Correlation ($r$)} & \textbf{Correlation ($\rho$)} \\
\midrule
English & 0.572 & 0.542 & 0.597 & 0.598 \\
Bangla & 0.504 & 0.505 & 0.374 & 0.353 \\
Hindi & 0.513 & 0.514 & 0.820 & 0.829 \\
Urdu & 0.494 & 0.484 & 0.844 & 0.875 \\
Barishal & 0.435 & 0.432 & 0.745 & 0.729 \\
Chittagong & 0.426 & 0.416 & 0.689 & 0.673 \\
Noakhali & 0.437 & 0.418 & 0.752 & 0.743 \\
Rangpur & 0.460 & 0.450 & 0.789 & 0.791 \\
Sylhet & 0.433 & 0.407 & 0.747 & 0.719 \\
\midrule
\rowcolor{blue!10} \textbf{Average} & \textbf{0.475} & \textbf{0.463} & \textbf{0.706} & \textbf{0.701} \\
\bottomrule
\end{tabular}
\caption{Cross-model validation comparing \texttt{Gemini-2.5-Flash} and \texttt{GPT-5-mini} across all nine languages.}
\label{tab:cross-model-validation}
\end{table*}

First, \textbf{explicit reasoning significantly enhances cultural understanding.} Compared to standard zero-shot VQA (55.34\% average accuracy), applying CoT yields a consistent overall improvement, raising average accuracy to 58.36\%. This gain is most pronounced in highly capable models, as illustrated in Figure~\ref{fig:prompting_strategies}a. For instance, \texttt{GPT-4.1-mini} surges from 58.93\% to 68.09\% under CoT, an absolute gain of over 9\%, while \texttt{Gemma3-27B} improves from 62.45\% to 68.21\%. By generating intermediate reasoning steps, models can successfully unpack visual evidence, recognize regional entities, and connect them to specific background knowledge before committing to an answer. This is particularly beneficial in knowledge-heavy domains such as \textit{National Achievements}, which sees a jump from 55.35\% to 62.33\%. This is consistent with prior work showing that explicitly surfacing the latent event knowledge inside LLMs and VLMs improves multilingual multimodal tasks \citep{roy-dipta-ferraro-2025-q2e}, although CoT is not uniformly beneficial: under conflicting modality evidence it improves agreement with human abstention judgments while amplifying overconfidence \citep{nazi2026omd}.

Conversely, we observe a surprising phenomenon regarding in-context learning: \textbf{few-shot prompting consistently degrades VQA performance.} Across all evaluated models, few-shot VQA accuracy (52.20\%) is markedly lower than zero-shot accuracy. For example, \texttt{Gemma3-4B} experiences a substantial 6.20\% absolute drop, and even state-of-the-art models like \texttt{GPT-4.1-mini} suffer a 5.12\% decline. In culturally knowledge-intensive tasks, few-shot examples appear to act as distractors rather than guides. We hypothesize that providing interleaved image-text examples from different cultural domains clutters the limited multimodal context window, inducing entity bias and causing models to hallucinate visual features from the demonstrations into the target image. Text-only Bangla work points in the same direction: a Bangla hate speech classifier performs best with only three in-context examples and degrades as more are added \citep{hossan-roy-dipta-2025-promptguard}, and Bangla mathematical reasoning drops by up to 41 points once semantically coherent but irrelevant context is present \citep{nazi2026dagger}.

However, for generative tasks, the trend is more nuanced. While few-shot prompting harms discriminative VQA, it provides marginal benefits for open-ended captioning (Figure~\ref{fig:prompting_strategies}b), raising the overall LLM-as-a-Judge score from 47.25\% to 47.69\%. Notably, it yields significant gains in specific domains like \textit{Media \& Movies} (improving from 43.12\% to 50.52\%). This suggests that while in-context examples do not successfully inject missing factual knowledge, they effectively teach models the expected stylistic formatting, tone, and length required for culturally descriptive captions.

\paragraph{Impact of Prompting Strategies Across Regional Dialects.} To understand if non-standard linguistic variations react differently to these frameworks, we isolate the performance of the five Bangla dialects (Barishal, Chittagong, Noakhali, Rangpur, and Sylhet). Figure~\ref{fig:prompting_strategies} (c and d) demonstrates that the overall prompting trends hold consistently across every regional dialect, albeit with varying degrees of sensitivity. 

For the VQA task, \textbf{CoT acts as a universal enhancer} for dialectal understanding. Rangpur proved to be the most resilient dialect overall, achieving the highest average accuracy under CoT (59.47\%) compared to its zero-shot baseline (56.29\%). Other dialects, such as Sylhet and Noakhali, experienced similar absolute gains of roughly 3.5\% to 4.1\% when reasoning steps were explicitly generated. This indicates that guiding models to "think" before answering helps them better parse phonological and lexical variations in regional text.

Conversely, \textbf{few-shot prompting universally degraded VQA performance} for every single dialect. While the average zero-shot accuracy across all dialects stood at 55.41\%, providing few-shot examples dragged the average down to 52.00\%. Notably, \textit{Chittagong} exhibited the most severe vulnerability to in-context distractors; its VQA accuracy plummeted from a strong 55.95\% in zero-shot to just 50.11\% in few-shot, a steep absolute drop of nearly 6\%. This suggests that models struggle particularly hard to map cross-image visual entities when the text is written in highly divergent regional scripts. For the generative captioning task, however, the dialects benefited universally from few-shot prompting. While the average zero-shot LLM-as-a-Judge score for dialects was low (42.29\%), providing in-context examples consistently raised the average to 45.29\%. \textit{Barishal} and \textit{Noakhali} saw notable improvements of roughly 2.5\% to 3.6\% in caption quality, reinforcing the hypothesis that few-shot examples successfully teach models the expected generative structure and tone, even when the target output must be written in a specific regional dialect.

\begin{table*}[t]
\centering
\small
\begin{tabular}{llcccc}
\toprule
\rowcolor{gray!20} \textbf{Model} & \textbf{Domain} & \textbf{Accuracy (\%)} & \textbf{Variance} & \textbf{Std. Error} & \textbf{95\% CI} \\
\midrule
\textbf{Gemma3-4B} & Nat. Achievements & 30.77 & 0.001439 & $\pm$ 3.79\% & [23.3\%, 38.2\%] \\
\textbf{Gemma3-4B} & Sports & 43.55 & 0.001967 & $\pm$ 4.43\% & [34.9\%, 52.2\%] \\
\midrule
\textbf{Gemma3-12B} & Nat. Achievements & 49.23 & 0.001689 & $\pm$ 4.11\% & [41.2\%, 57.3\%] \\
\textbf{Gemma3-12B} & Sports & 53.23 & 0.001992 & $\pm$ 4.46\% & [44.5\%, 62.0\%] \\
\midrule
\textbf{Gemma3-27B} & Nat. Achievements & 50.77 & 0.001689 & $\pm$ 4.11\% & [42.7\%, 58.8\%] \\
\textbf{Gemma3-27B} & Sports & 69.35 & 0.001700 & $\pm$ 4.12\% & [61.3\%, 77.4\%] \\
\midrule
\textbf{Qwen2.5-VL-7B} & Nat. Achievements & 69.23 & 0.001439 & $\pm$ 3.79\% & [61.8\%, 76.7\%] \\
\textbf{Qwen2.5-VL-7B} & Sports & 48.39 & 0.001998 & $\pm$ 4.47\% & [39.6\%, 57.2\%] \\
\midrule
\textbf{Qwen3-VL-8B} & Nat. Achievements & 77.78 & 0.001168 & $\pm$ 3.42\% & [71.1\%, 84.5\%] \\
\textbf{Qwen3-VL-8B} & Sports & 49.18 & 0.001999 & $\pm$ 4.47\% & [40.4\%, 57.9\%] \\
\midrule
\textbf{GPT-4.1-mini} & Nat. Achievements & 58.11 & 0.001645 & $\pm$ 4.06\% & [50.2\%, 66.1\%] \\
\textbf{GPT-4.1-mini} & Sports & 53.23 & 0.001992 & $\pm$ 4.46\% & [44.5\%, 62.0\%] \\
\bottomrule
\end{tabular}
\caption{Variance, Standard Error, and 95\% Confidence Intervals for VQA Accuracy in the smallest domains (\textit{National Achievements} and \textit{Sports}) across all evaluated models in Standard Bangla.}
\label{tab:confidence_intervals}
\end{table*}

\section{Cross-Model Validation of LLM-as-a-Judge}
\label{app:cross-model-validation}

Because \texttt{Gemini-2.5-Flash} was utilized for translation and evaluation in parts of our pipeline, we conducted a cross-model validation step to empirically rule out the risk of circularity and self-preference bias. We ran an independent judge, \texttt{GPT-5-mini}, across the caption generation outputs for all nine linguistic variants (English, Standard Bangla, Hindi, Urdu, and the five regional Bangla dialects).

Table~\ref{tab:cross-model-validation} presents the direct comparison of the outputs from ``LLM as a judge (\texttt{Gemini-2.5-Flash})'' against ``LLM as a judge (\texttt{GPT-5-mini})'' for both zero-shot and few-shot caption generation tasks. 

\paragraph{Analysis:} The validation demonstrates a consistent alignment between \texttt{Gemini-2.5-Flash} and \texttt{GPT-5-mini} scoring. For the translated languages (English, Hindi, Urdu) and the regional dialects, both Pearson ($r$) and Spearman ($\rho$) correlation coefficients are remarkably strong, generally ranging from 0.60 to 0.87. This high correlation confirms that the relative ranking of model caption quality remains robust regardless of the LLM judge used. 

While the rank correlation in standard Bangla is comparatively lower ($r=0.374$), the absolute mean scores between the two judges are virtually identical (0.504 vs. 0.505). This indicates that even where the exact rank ordering shifts slightly, both models assess the baseline generation quality with the exact same level of severity. Overall, this comprehensive evaluation confirms that using \texttt{Gemini-2.5-Flash} as the primary judge provides a stable, unbiased, and representative reflection of caption quality across these diverse cultural and linguistic variants. Both judges are nonetheless proprietary, which limits full reproducibility; open, fact-grounded caption evaluators have been proposed for exactly this reason, though none is yet available for Bangla \citep{roy-dipta-etal-2026-vc}.

\section{Statistical Significance and Variance Analysis}
\label{app:statistical_significance}

To ensure that point estimates effectively distinguish true model degradation from noise, particularly in smaller domains, we computed paired significance tests, explicit standard deviation estimates, and 95\% confidence intervals across our experimental conditions.

\begin{table*}[t]
\centering
\small
\setlength{\tabcolsep}{4pt}
\renewcommand{\arraystretch}{1.2}
\begin{tabular}{@{}p{3cm}p{4.0cm}
  >{\centering\arraybackslash}p{1.1cm}
  >{\centering\arraybackslash}p{1.1cm}
  >{\centering\arraybackslash}p{1.1cm}
  >{\centering\arraybackslash}p{1.2cm}
  p{2.7cm}@{}}
\toprule
\textbf{Error Category} & \textbf{Description} &
\textbf{\makecell{Bangla\\\textit{\small(n=100)}}} &
\textbf{\makecell{English\\\textit{\small(n=100)}}} &
\textbf{\makecell{Total\\\textit{\small(n=200)}}} &
\textbf{\makecell{\% of\\Failures}} &
\textbf{Most Affected Domains} \\
\midrule
\textbf{Entity \& Artifact Misrecognition} & Hallucinating unrelated or foreign entities instead of local items. & 18 & 22 & 40 & 42.6\% & Food, Media \& Movies \\
\textbf{Surface-Level Descriptive Fallback} & Providing literal visual descriptions devoid of deeper cultural context. & 14 & 15 & 29 & 30.8\% & Personalities, History \\
\textbf{Societal \& Cultural Bias} & Exhibiting political skew or visual stereotyping (e.g., gender). & 8 & 6 & 14 & 14.9\% & Politics, Personalities \\
\textbf{Language Constraint Violation} & Generating English text despite strict source-language prompts. & 10 & 1 & 11 & 11.7\% & Sports, Nat. Achievements \\
\bottomrule
\end{tabular}
\caption{Systematic breakdown of 94 notable errors identified within a 200-sample manual evaluation split evenly between Standard Bangla and English. Overall error rate across the sample is 47.0\% (20.0\% entity misrecognition, 14.5\% descriptive fallback, 7.0\% societal bias, 5.5\% constraint violation).}
\label{tab:error_quantification}
\end{table*}

\subsection{Statistical Significance of Dialectal Degradation}
To determine whether the performance degradation from Standard Bangla to regional dialects is statistically distinct from noise, we conducted paired $t$-tests across 54 conditions, obtained from the full crossing of 6 evaluated models with 9 cultural domains. For each condition, we compared the Standard Bangla performance directly against the mean performance of the five generated dialectal variants. Pairing observations in this way is appropriate here because both measurements within a condition come from the same model evaluated on the same domain and the same underlying images, so the test isolates the effect of linguistic variety while controlling for the substantial variation in difficulty across models and domains.

For VQA accuracy, the mean difference across the 54 paired conditions is a 1.95\% drop from Standard Bangla to the dialect average. The test yields a statistic of $t = 3.960$ and a $p$-value of $p = 2.249 \times 10^{-4}$, indicating that the accuracy degradation attributable to dialectal variation is statistically significant ($p < 0.001$). While the absolute magnitude of this drop is modest, its consistency across models and domains is what drives significance: the degradation is small but systematic rather than sporadic.

For caption generation, scored by LLM-as-a-Judge, the mean difference is a 0.083 drop on a 0--1 scale. The test yields a statistic of $t = 10.347$ and a $p$-value of $p = 2.516 \times 10^{-14}$, making the degradation in generative captioning under dialectal shifts highly statistically significant ($p \ll 0.001$). The considerably larger test statistic relative to VQA reflects both a larger effect size and lower variance across conditions, quantitatively confirming the central observation of RQ1 that free-form generation is markedly more fragile under dialectal variation than answer-constrained reasoning.




\subsection{Variance and 95\% Confidence Intervals for Small-Sample Domains}

We acknowledge that domains such as \textit{National Achievements} ($n=75$) and \textit{Sports} ($n=63$) require confidence intervals to properly contextualize the point estimates due to their smaller sample sizes. We calculated the variance of the sample proportions $\frac{p(1-p)}{n}$, the Standard Error (SE), and the 95\% Confidence Intervals using the formulation $p \pm 1.96 \cdot SE$. Table~\ref{tab:confidence_intervals} provides these metrics for VQA Accuracy in Standard Bangla across all evaluated models for these specific domains. While the margins of error (ranging from approximately $\pm 3.42\%$ to $4.47\%$) introduce necessary nuance for cross-model comparisons within these specific sub-domains, they remain tightly bound and do not invalidate the broader aggregated findings of the paper.


\section{Examples}
\label{app:examples}

\subsection{Systematic Error Quantification}
\label{sec:appendix_h2_quant}

To systematically quantify the failure modes summarized in Section \ref{sec:error_analysis}, we manually coded the 94 failed outputs from our stratified random sample of 200 generations (yielding 50 errors in Standard Bangla and 44 in English). Table \ref{tab:error_quantification} presents the complete language-wise breakdown of these errors. The data demonstrates that while entity misrecognition is universally the most common bottleneck (comprising 20.0\% of all evaluated samples), Bangla-prompted generations are uniquely susceptible to language constraint violations, frequently defaulting to English outputs despite explicit instructions to generate in Bangla.

\subsection{Failed Examples}
\label{app:failed_examples}

In this section, we present a selection of failed examples that highlight common error modes encountered by the evaluated models. These failures illustrate various limitations in current multilingual VLMs, ranging from cultural misidentification and surface-level descriptive fallbacks to language constraint violations and implicit biases. Analyzing these specific instances provides deeper insight into the gaps between standard visual recognition and genuine, culturally grounded understanding.










\begin{tcolorbox}[enhanced,
  colback=red!5!white,
  colframe=red!75!black,
  title=Personality\_026,
  fonttitle=\bfseries,
  breakable]

\small

\begin{center}
\includegraphics[width=0.35\textwidth]{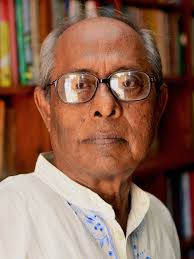}
\end{center}

\textbf{Ground Truth:}  
{\kalpurush "হাসান আজিজুল হক — লেখক, ছোটগল্প ও উপন্যাসের মাধ্যমে মানুষের জটিল মানসিকতা ফুটিয়ে তুলেছেন।"}  
\textit{(en: Hasan Azizul Haque — Writer who has portrayed the complex human psyche through short stories and novels.) } 

\vspace{0.5em}

\textbf{\texttt{GPT-4.1-mini} Generated Caption:}  
{\kalpurush "চশমা ও সাদা পোশাক পরিহিত একজন প্রবীণ ব্যক্তি বইয়ের তাকের সামনে চিন্তাশীল ভঙ্গিতে তাকিয়ে আছেন।"}  
\textit{(en: An elderly man wearing glasses and white clothes looks thoughtfully in front of a bookshelf.)}  

\vspace{0.5em}

\textbf{Explanation:}  
While the generated caption is factually correct in describing visual attributes, it fails to capture the person’s identity and cultural significance. This shows a common model limitation: defaulting to surface-level descriptions when unable to recognize notable personalities.

\end{tcolorbox}

\begin{tcolorbox}[enhanced,
  colback=red!5!white,
  colframe=red!75!black,
  title=Food\_010,
  fonttitle=\bfseries,
  breakable]

\small

\begin{center}
\includegraphics[width=0.45\textwidth]{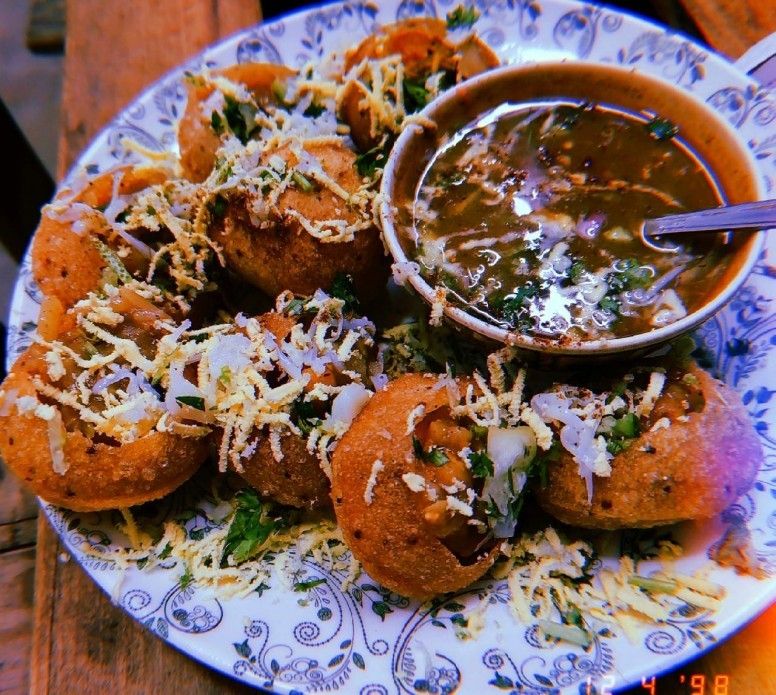}
\end{center}

\textbf{Ground Truth:}  
{\kalpurush "ফুচকা, ঝাল ও টক-মিষ্টি স্বাদের এক অসাধারণ সমন্বয়।"}
\textit{(en: A wonderful combination of fuchka, salty, and sweet-sour flavours.)}  

\vspace{0.5em}

\textbf{\texttt{Gemma3-27B} Generated Caption:}  
{\kalpurush "পানির মধ্যে ভেজানো কিছু মশলা, সবুজ পাতা, এবং নারকেল ছাড়াও একটি সুস্বাদু পানীয় এবং ভাজা খাবার সমন্বিত একটি সুস্বাদু ভারতীয় রাস্তার খাবার।"}  
\textit{(en: A delicious Indian street food consisting of some spices, green leaves, and coconut soaked in water, along with a delicious drink and fried food.) } 

\vspace{0.5em}

\textbf{Explanation:}  
The generated caption is overly descriptive and shifts the focus toward generic Indian street food rather than identifying the dish as “fuchka.” While factually plausible, it misrepresents the key cultural context.

\end{tcolorbox}

\begin{tcolorbox}[enhanced,
  colback=red!5!white,
  colframe=red!75!black,
  title=Politics\_006,
  fonttitle=\bfseries, breakable]

\small

\begin{center}
\includegraphics[width=0.45\textwidth]{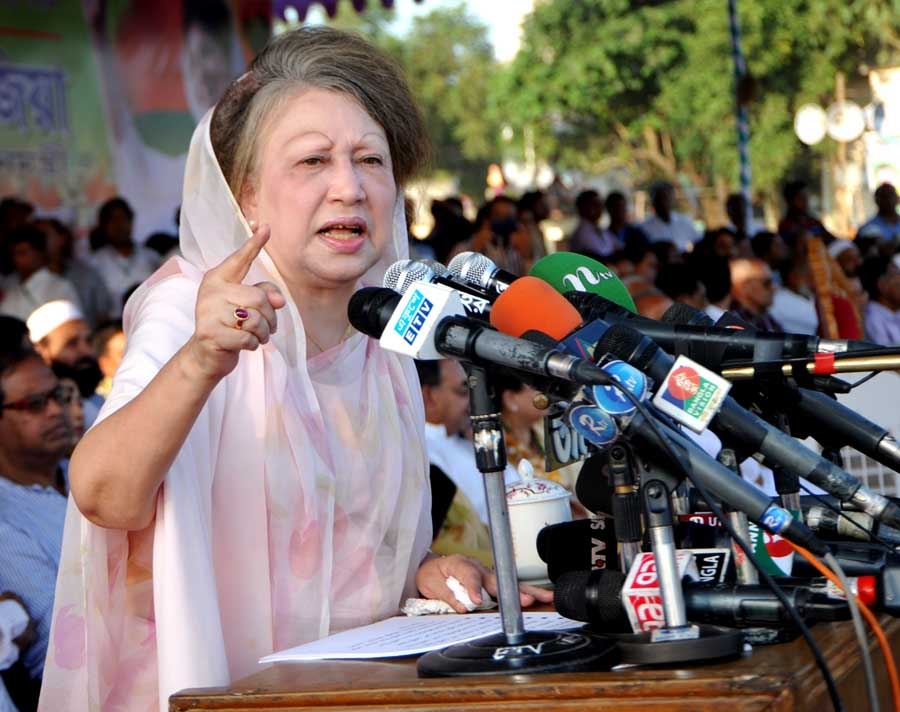}
\end{center}

\textbf{Question:}  
{\kalpurush "ছবির নেত্রী কোন দলের প্রধান?"}  
\textit{(en: Which party is the leader of the picture?)}

\textbf{Ground Truth Answer:}
{\kalpurush "বিএনপি"} \textit{(en: BNP)}

\textbf{\texttt{Gemma3-4B} Predicted Answer:}  
{\kalpurush "শেখ হাসিনা"} \textit{(en: Sheikh Hasina)}

\vspace{0.5em}

\textbf{Explanation:}  
The correct answer is BNP, but the model generated Sheikh Hasina, reflecting a strong political bias toward the Awami League. This not only demonstrates over-association with ruling party figures but also shows factual inconsistency, as Sheikh Hasina is a person and not the name of a political party.

\end{tcolorbox}

\begin{tcolorbox}[enhanced,
  colback=red!5!white,
  colframe=red!75!black,
  title=Food\_008,
  fonttitle=\bfseries, breakable]

\small

\begin{center}
\includegraphics[width=0.45\textwidth]{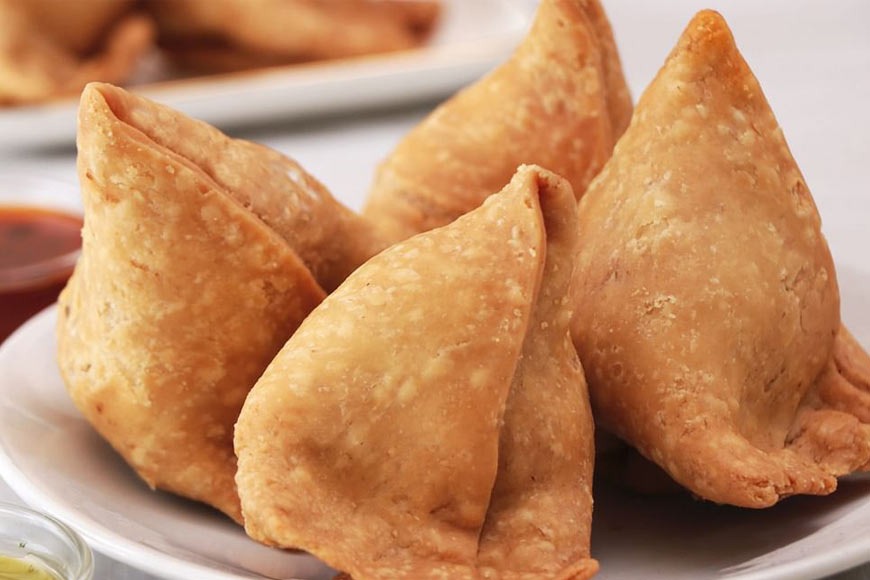}
\end{center}

\textbf{Ground Truth Caption:}  
{\kalpurush "সুস্বাদু শিঙারা, তেলমশলার সোনালি রঙে ভাজা।"}
\textit{(en: Delicious shingara, fried to a golden color in oil and spices.)}  

\textbf{\texttt{Qwen2.5-VL-7B} Generated Caption:}  
{\kalpurush "সামুদ্রিক খাবার দৃশ্য।"}  
\textit{(en: Seafood scene.)}  

\vspace{0.5em}

\textbf{Explanation:}  
The model failed to recognize the Bangladeshi snack \emph{shingara} and instead hallucinated a completely unrelated “seafood scene.” This illustrates cultural and recognition gaps, where local foods are often misinterpreted as generic or foreign items.

\end{tcolorbox}

\begin{tcolorbox}[enhanced,
  colback=red!5!white,
  colframe=red!75!black,
  title=Personality\_004,
  fonttitle=\bfseries, breakable]

\small

\begin{center}
\includegraphics[width=0.45\textwidth]{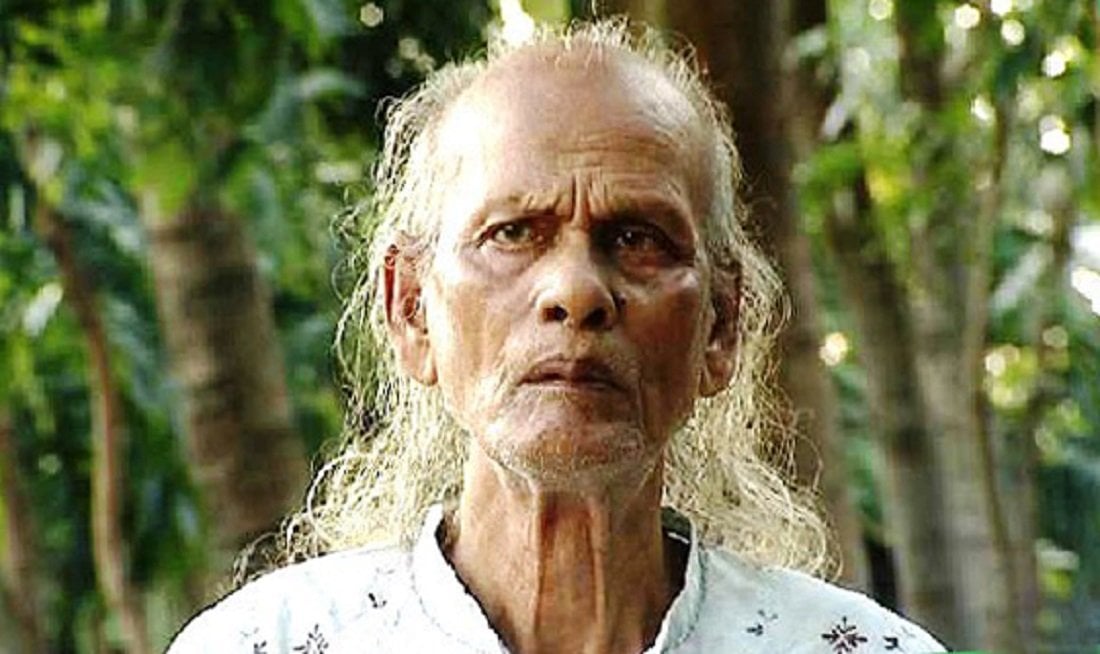}
\end{center}

\textbf{Ground Truth Caption:}  
{\kalpurush "উস্তাদ শাহ আবদুল করিম হচ্ছেন বাংলাদেশের কিংবদন্তি সঙ্গীতশিল্পী, সুরকার, গীতিকার ও সঙ্গীত শিক্ষক।"}
\textit{(en: Ustad Shah Abdul Karim is a legendary musician, composer, lyricist, and music teacher of Bangladesh.)}

\textbf{\texttt{Qwen2.5-VL-7B} Zero-shot Caption:}  
{\kalpurush "আমার চিত্রে একজন প্রাচীন মহিলা দেখা যাচ্ছেন।"} 
\textit{(en: An ancient woman is seen in my picture.)}  

\textbf{\texttt{Qwen2.5-VL-7B} Few-shot Caption:}  
{\kalpurush "বৃদ্ধ মহিলার প্রাণের স্বাদ ও সাহস।"}
\textit{(en: The old woman's zest for life and courage.)}  

\vspace{0.5em}

\textbf{Explanation:}  
The model repeatedly misclassified Shah Abdul Karim, a legendary male musician of Bangladesh, as a woman, reflecting a gender bias where long hair is incorrectly equated with female identity. This highlights both cultural unfamiliarity and gender stereotyping in caption generation.

\end{tcolorbox}

\begin{tcolorbox}[enhanced,
  colback=red!5!white,
  colframe=red!75!black,
  title=Sports\_010,
  fonttitle=\bfseries, breakable]

\small

\begin{center}
\includegraphics[width=0.45\textwidth]{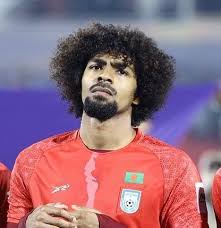}
\end{center}

\textbf{Question:} {\kalpurush "ছবির খেলোয়াড় হামজা চৌধুরি কোন পজিশনে খেলেন?"}
(en: What position does the player in the picture, Hamza Chowdhury, play?)

\textbf{Ground Truth:} {\kalpurush "মিডফিল্ডার"}
\textit{(en: Midfielder)}

\textbf{\texttt{Gemma3-4B} Predicted:} "batsman"

\medskip
\textbf{Explanation:} The model violated the language constraint by answering in English and produced a factually incorrect label. “Batsman” is a cricket role, whereas the correct answer is the football position of midfielder, indicating cross-sport confusion and weak factual grounding.
\end{tcolorbox}

\onecolumn
\newpage
\clearpage

\subsection{Sample Data Example}
\begin{tcolorbox}[enhanced,
  colback=blue!5,
  colframe=blue!75,
  title=Culture\_114,
  fonttitle=\bfseries]

\small

\begin{center}
\includegraphics[width=0.45\textwidth]{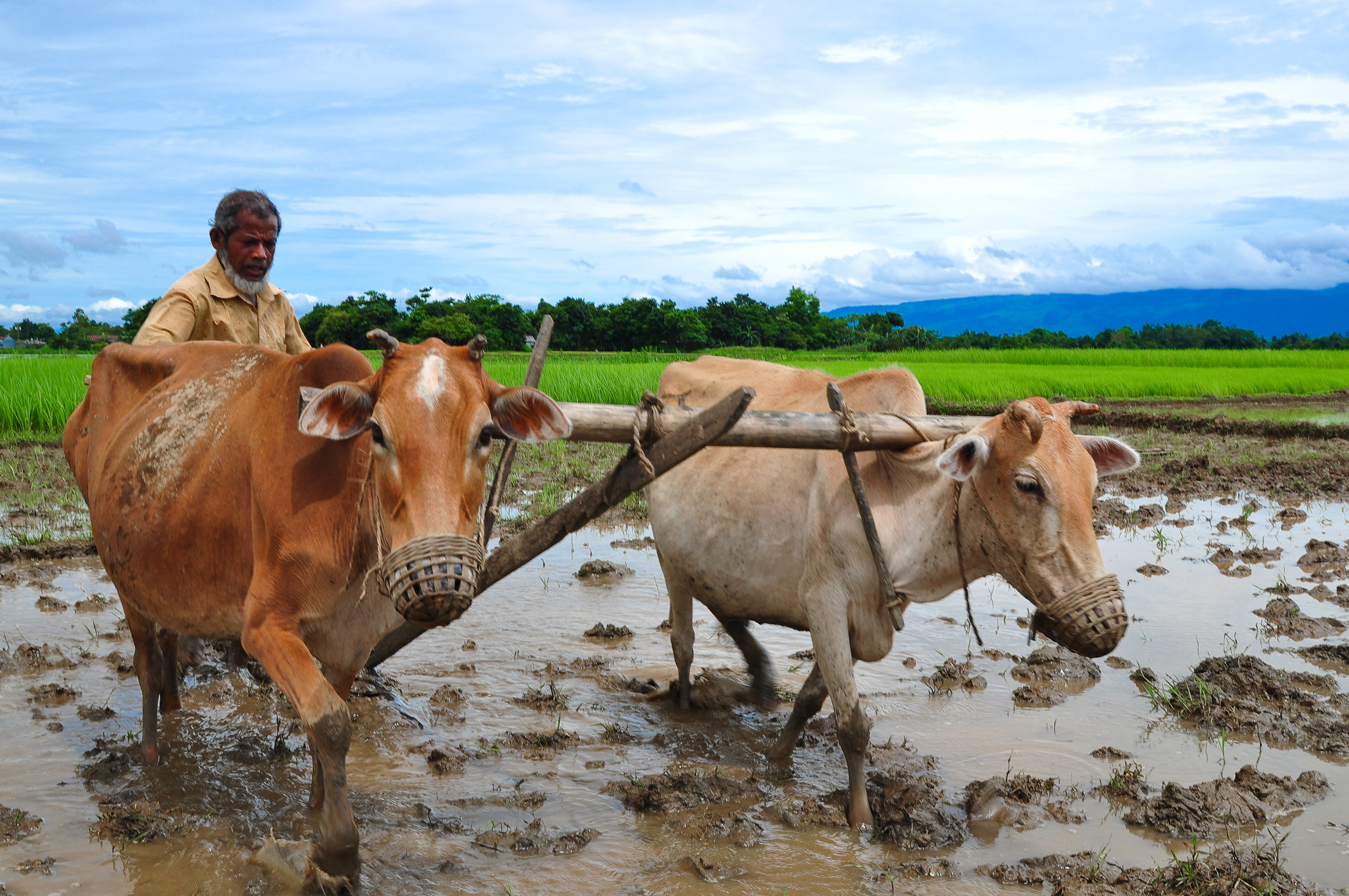}
\end{center}

\begin{center}
\includegraphics[width=\textwidth]{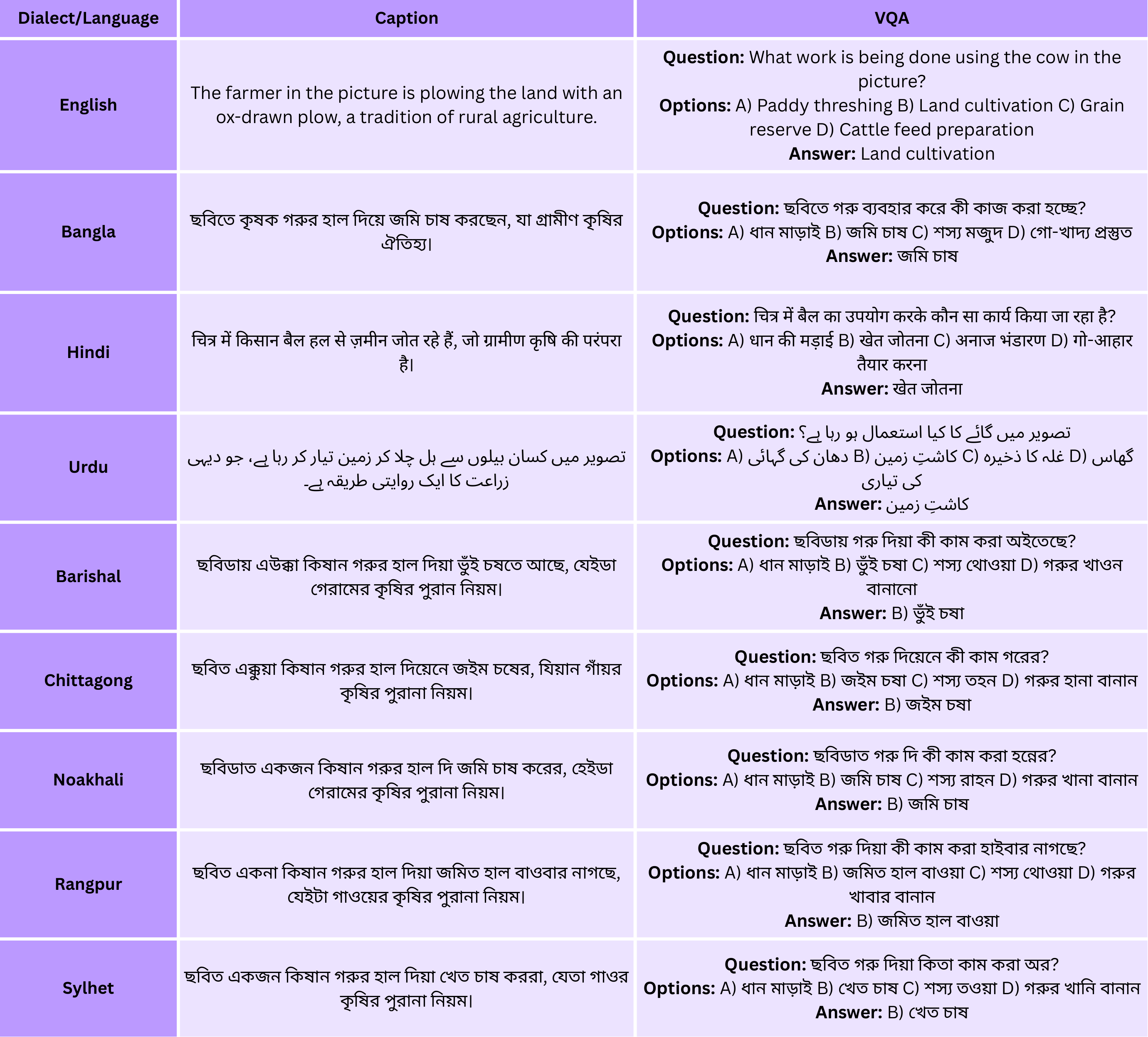}
\end{center}

\end{tcolorbox}

\newpage
\clearpage

\section{Full Experimental Results}\label{app:full_results}

\begin{table*}[!h]
\centering
\small
\scriptsize
\begin{tabular*}{\textwidth}{@{\extracolsep{\fill}} l c | c c c c | c c c c | c }
\toprule
 &  & \multicolumn{4}{c|}{\textbf{Zero-Shot}} & \multicolumn{4}{c|}{\textbf{Few-Shot}} & \textbf{CoT} \\
\textbf{Models} & \textbf{Domain} & \multicolumn{3}{c}{\textbf{Caption}} & \textbf{VQA} & \multicolumn{3}{c}{\textbf{Caption}} & \textbf{VQA} & \textbf{VQA} \\
\cmidrule{3-5}\cmidrule{7-9}
 &  & \textbf{B-F1} & \textbf{Gemini} & \textbf{GPT} & \textbf{Accuracy (\%)} & \textbf{B-F1} & \textbf{Gemini} & \textbf{GPT} & \textbf{Accuracy (\%)} & \textbf{Accuracy (\%)} \\
\midrule
\multirow{9}{*}{\texttt{Gemma3-4B}} & Cult. & 0.69 & 0.53 & 0.56 & 0.63 & 0.69 & 0.52 & 0.48 & 0.59 & 0.52 \\
 & Food & 0.70 & 0.52 & 0.41 & 0.74 & 0.70 & 0.53 & 0.50 & 0.63 & 0.78 \\
 & Hist. & 0.69 & 0.54 & 0.63 & 0.47 & 0.68 & 0.47 & 0.36 & 0.37 & 0.44 \\
 & M\&M & 0.66 & 0.50 & 0.50 & 0.30 & 0.84 & 0.52 & 0.57 & 0.32 & 0.38 \\
 & Nat. Achv. & 0.70 & 0.56 & 0.64 & 0.31 & 0.67 & 0.50 & 0.40 & 0.29 & 0.36 \\
 & Nature & 0.70 & 0.58 & 0.55 & 0.66 & 0.69 & 0.54 & 0.47 & 0.74 & 0.59 \\
 & Pers. & 0.65 & 0.45 & 0.51 & 0.61 & 0.69 & 0.49 & 0.51 & 0.62 & 0.49 \\
 & Pol. & 0.67 & 0.50 & 0.57 & 0.39 & 0.69 & 0.48 & 0.42 & 0.40 & 0.39 \\
 & Sports & 0.68 & 0.52 & 0.48 & 0.44 & 0.68 & 0.51 & 0.49 & 0.26 & 0.48 \\
\midrule
\multirow{9}{*}{\texttt{Gemma3-12B}} & Cult. & 0.70 & 0.55 & 0.58 & 0.68 & 0.69 & 0.52 & 0.50 & 0.70 & 0.72 \\
 & Food & 0.71 & 0.53 & 0.62 & 0.83 & 0.71 & 0.52 & 0.49 & 0.82 & 0.89 \\
 & Hist. & 0.69 & 0.54 & 0.59 & 0.62 & 0.68 & 0.48 & 0.45 & 0.63 & 0.75 \\
 & M\&M & 0.66 & 0.51 & 0.52 & 0.40 & 0.78 & 0.52 & 0.55 & 0.32 & 0.33 \\
 & Nat. Achv. & 0.69 & 0.56 & 0.69 & 0.49 & 0.66 & 0.49 & 0.44 & 0.50 & 0.59 \\
 & Nature & 0.70 & 0.58 & 0.59 & 0.63 & 0.69 & 0.56 & 0.51 & 0.59 & 0.66 \\
 & Pers. & 0.64 & 0.49 & 0.50 & 0.67 & 0.68 & 0.48 & 0.44 & 0.70 & 0.74 \\
 & Pol. & 0.66 & 0.51 & 0.57 & 0.55 & 0.68 & 0.48 & 0.48 & 0.56 & 0.61 \\
 & Sports & 0.68 & 0.57 & 0.66 & 0.53 & 0.67 & 0.48 & 0.47 & 0.47 & 0.46 \\
\midrule
\multirow{9}{*}{\texttt{Gemma3-27B}} & Cult. & 0.67 & 0.44 & 0.44 & 0.72 & 0.70 & 0.53 & 0.54 & 0.73 & 0.73 \\
 & Food & 0.71 & 0.58 & 0.62 & 0.87 & 0.72 & 0.55 & 0.53 & 0.89 & 0.89 \\
 & Hist. & 0.70 & 0.56 & 0.66 & 0.69 & 0.69 & 0.51 & 0.46 & 0.69 & 0.74 \\
 & M\&M & 0.67 & 0.47 & 0.51 & 0.36 & 0.82 & 0.53 & 0.53 & 0.32 & 0.42 \\
 & Nat. Achv. & 0.71 & 0.61 & 0.66 & 0.51 & 0.67 & 0.50 & 0.47 & 0.51 & 0.78 \\
 & Nature & 0.70 & 0.57 & 0.62 & 0.66 & 0.70 & 0.59 & 0.51 & 0.72 & 0.78 \\
 & Pers. & 0.65 & 0.51 & 0.54 & 0.78 & 0.70 & 0.52 & 0.46 & 0.83 & 0.76 \\
 & Pol. & 0.67 & 0.52 & 0.66 & 0.56 & 0.69 & 0.49 & 0.43 & 0.50 & 0.62 \\
 & Sports & 0.70 & 0.58 & 0.66 & 0.69 & 0.70 & 0.52 & 0.45 & 0.60 & 0.74 \\
\midrule
\multirow{9}{*}{\texttt{Qwen2.5-VL-7B}} & Cult. & 0.69 & 0.46 & 0.57 & 0.55 & 0.66 & 0.48 & 0.61 & 0.47 & 0.59 \\
 & Food & 0.70 & 0.42 & 0.52 & 0.68 & 0.64 & 0.55 & 0.65 & 0.54 & 0.68 \\
 & Hist. & 0.70 & 0.51 & 0.60 & 0.60 & 0.67 & 0.57 & 0.56 & 0.52 & 0.52 \\
 & M\&M & 0.70 & 0.44 & 0.50 & 0.22 & 0.76 & 0.50 & 0.64 & 0.27 & 0.17 \\
 & Nat. Achv. & 0.69 & 0.46 & 0.67 & 0.69 & 0.69 & 0.51 & 0.71 & 0.71 & 0.54 \\
 & Nature & 0.70 & 0.49 & 0.63 & 0.42 & 0.55 & 0.38 & 0.66 & 0.64 & 0.39 \\
 & Pers. & 0.65 & 0.44 & 0.52 & 0.47 & 0.69 & 0.48 & 0.51 & 0.46 & 0.56 \\
 & Pol. & 0.66 & 0.46 & 0.61 & 0.39 & 0.69 & 0.48 & 0.51 & 0.32 & 0.41 \\
 & Sports & 0.68 & 0.47 & 0.65 & 0.48 & 0.72 & 0.52 & 0.57 & 0.48 & 0.40 \\
\midrule
\multirow{9}{*}{\texttt{Qwen3-VL-8B}} & Cult. & 0.67 & 0.44 & 0.40 & 0.68 & 0.67 & 0.42 & 0.39 & 0.68 & 0.62 \\
 & Food & 0.68 & 0.42 & 0.39 & 0.70 & 0.69 & 0.47 & 0.40 & 0.66 & 0.66 \\
 & Hist. & 0.67 & 0.44 & 0.43 & 0.69 & 0.67 & 0.39 & 0.34 & 0.69 & 0.78 \\
 & M\&M & 0.67 & 0.36 & 0.34 & 0.40 & 0.67 & 0.37 & 0.37 & 0.28 & 0.38 \\
 & Nat. Achv. & 0.66 & 0.40 & 0.48 & 0.78 & 0.65 & 0.37 & 0.37 & 0.63 & 0.70 \\
 & Nature & 0.68 & 0.43 & 0.47 & 0.46 & 0.64 & 0.54 & 0.45 & 0.47 & 0.52 \\
 & Pers. & 0.63 & 0.41 & 0.37 & 0.51 & 0.65 & 0.45 & 0.39 & 0.52 & 0.57 \\
 & Pol. & 0.64 & 0.40 & 0.50 & 0.45 & 0.68 & 0.43 & 0.36 & 0.48 & 0.45 \\
 & Sports & 0.66 & 0.41 & 0.45 & 0.49 & 0.64 & 0.34 & 0.36 & 0.60 & 0.45 \\
\midrule
\multirow{9}{*}{\texttt{GPT-4.1-mini}} & Cult. & 0.71 & 0.57 & 0.36 & 0.65 & 0.71 & 0.57 & 0.43 & 0.56 & 0.77 \\
 & Food & 0.72 & 0.56 & 0.36 & 0.88 & 0.73 & 0.54 & 0.37 & 0.81 & 0.94 \\
 & Hist. & 0.71 & 0.58 & 0.56 & 0.66 & 0.72 & 0.57 & 0.48 & 0.60 & 0.72 \\
 & M\&M & 0.70 & 0.54 & 0.45 & 0.49 & 0.89 & 0.61 & 0.37 & 0.42 & 0.46 \\
 & Nat. Achv. & 0.71 & 0.62 & 0.49 & 0.58 & 0.71 & 0.60 & 0.53 & 0.57 & 0.78 \\
 & Nature & 0.71 & 0.60 & 0.47 & 0.57 & 0.72 & 0.61 & 0.49 & 0.56 & 0.65 \\
 & Pers. & 0.66 & 0.51 & 0.36 & 0.65 & 0.71 & 0.53 & 0.41 & 0.60 & 0.75 \\
 & Pol. & 0.66 & 0.50 & 0.49 & 0.50 & 0.69 & 0.52 & 0.42 & 0.50 & 0.61 \\
 & Sports & 0.69 & 0.58 & 0.52 & 0.53 & 0.71 & 0.55 & 0.59 & 0.48 & 0.69 \\
\bottomrule
\end{tabular*}
\caption{Full results by model and domain for Bangla Language.}
\label{tab:Results-Bangla}
\end{table*}

\begin{table*}[t]
\centering
\small
\scriptsize
\begin{tabular*}{\textwidth}{@{\extracolsep{\fill}} l c | c c c c | c c c c | c }
\toprule
 &  & \multicolumn{4}{c|}{\textbf{Zero-Shot}} & \multicolumn{4}{c|}{\textbf{Few-Shot}} & \textbf{CoT} \\
\textbf{Models} & \textbf{Domain} & \multicolumn{3}{c}{\textbf{Caption}} & \textbf{VQA} & \multicolumn{3}{c}{\textbf{Caption}} & \textbf{VQA} & \textbf{VQA} \\
\cmidrule{3-5}\cmidrule{7-9}
 &  & \textbf{B-F1} & \textbf{Gemini} & \textbf{GPT} & \textbf{Accuracy (\%)} & \textbf{B-F1} & \textbf{Gemini} & \textbf{GPT} & \textbf{Accuracy (\%)} & \textbf{Accuracy (\%)} \\
\midrule
\multirow{9}{*}{\texttt{Gemma3-4B}} & Cult. & 0.87 & 0.65 & 0.57 & 0.72 & 0.86 & 0.56 & 0.54 & 0.58 & 0.50 \\
 & Food & 0.85 & 0.61 & 0.54 & 0.75 & 0.86 & 0.54 & 0.47 & 0.70 & 0.68 \\
 & Hist. & 0.86 & 0.61 & 0.42 & 0.43 & 0.84 & 0.44 & 0.43 & 0.40 & 0.44 \\
 & M\&M & 0.85 & 0.55 & 0.58 & 0.26 & 0.92 & 0.55 & 0.54 & 0.34 & 0.42 \\
 & Nat. Achv. & 0.87 & 0.63 & 0.44 & 0.41 & 0.87 & 0.54 & 0.42 & 0.34 & 0.33 \\
 & Nature & 0.86 & 0.66 & 0.63 & 0.55 & 0.86 & 0.59 & 0.56 & 0.55 & 0.39 \\
 & Pers. & 0.83 & 0.55 & 0.47 & 0.57 & 0.84 & 0.51 & 0.46 & 0.52 & 0.56 \\
 & Pol. & 0.84 & 0.56 & 0.47 & 0.41 & 0.83 & 0.49 & 0.40 & 0.43 & 0.44 \\
 & Sports & 0.86 & 0.60 & 0.52 & 0.54 & 0.84 & 0.43 & 0.44 & 0.42 & 0.55 \\
\midrule
\multirow{9}{*}{\texttt{Gemma3-12B}} & Cult. & 0.87 & 0.62 & 0.68 & 0.69 & 0.86 & 0.55 & 0.59 & 0.72 & 0.71 \\
 & Food & 0.85 & 0.59 & 0.43 & 0.73 & 0.86 & 0.54 & 0.44 & 0.77 & 0.73 \\
 & Hist. & 0.85 & 0.60 & 0.44 & 0.62 & 0.85 & 0.51 & 0.42 & 0.60 & 0.74 \\
 & M\&M & 0.85 & 0.53 & 0.49 & 0.44 & 0.90 & 0.59 & 0.53 & 0.30 & 0.40 \\
 & Nat. Achv. & 0.87 & 0.63 & 0.50 & 0.57 & 0.86 & 0.48 & 0.43 & 0.50 & 0.60 \\
 & Nature & 0.86 & 0.64 & 0.46 & 0.53 & 0.85 & 0.56 & 0.44 & 0.58 & 0.64 \\
 & Pers. & 0.83 & 0.53 & 0.50 & 0.62 & 0.85 & 0.50 & 0.46 & 0.71 & 0.65 \\
 & Pol. & 0.83 & 0.54 & 0.42 & 0.52 & 0.85 & 0.50 & 0.44 & 0.52 & 0.63 \\
 & Sports & 0.85 & 0.58 & 0.48 & 0.61 & 0.85 & 0.50 & 0.45 & 0.56 & 0.62 \\
\midrule
\multirow{9}{*}{\texttt{Gemma3-27B}} & Cult. & 0.87 & 0.66 & 0.47 & 0.77 & 0.86 & 0.57 & 0.50 & 0.75 & 0.80 \\
 & Food & 0.85 & 0.62 & 0.56 & 0.88 & 0.86 & 0.54 & 0.45 & 0.84 & 0.82 \\
 & Hist. & 0.85 & 0.65 & 0.51 & 0.62 & 0.87 & 0.57 & 0.42 & 0.62 & 0.73 \\
 & M\&M & 0.86 & 0.53 & 0.41 & 0.41 & 0.92 & 0.54 & 0.48 & 0.36 & 0.45 \\
 & Nat. Achv. & 0.87 & 0.62 & 0.59 & 0.65 & 0.86 & 0.55 & 0.43 & 0.50 & 0.86 \\
 & Nature & 0.86 & 0.67 & 0.54 & 0.60 & 0.85 & 0.55 & 0.52 & 0.61 & 0.68 \\
 & Pers. & 0.83 & 0.55 & 0.46 & 0.81 & 0.85 & 0.50 & 0.40 & 0.79 & 0.77 \\
 & Pol. & 0.83 & 0.55 & 0.48 & 0.51 & 0.84 & 0.51 & 0.41 & 0.47 & 0.49 \\
 & Sports & 0.87 & 0.66 & 0.60 & 0.70 & 0.85 & 0.50 & 0.43 & 0.70 & 0.61 \\
\midrule
\multirow{9}{*}{\texttt{Qwen2.5-VL-7B}} & Cult. & 0.87 & 0.69 & 0.68 & 0.62 & 0.84 & 0.59 & 0.57 & 0.65 & 0.58 \\
 & Food & 0.85 & 0.60 & 0.69 & 0.77 & 0.81 & 0.62 & 0.65 & 0.72 & 0.75 \\
 & Hist. & 0.86 & 0.58 & 0.55 & 0.62 & 0.83 & 0.52 & 0.49 & 0.54 & 0.60 \\
 & M\&M & 0.86 & 0.53 & 0.49 & 0.32 & 0.90 & 0.54 & 0.42 & 0.30 & 0.37 \\
 & Nat. Achv. & 0.88 & 0.64 & 0.70 & 0.60 & 0.87 & 0.62 & 0.62 & 0.51 & 0.61 \\
 & Nature & 0.87 & 0.68 & 0.67 & 0.49 & 0.77 & 0.56 & 0.52 & 0.40 & 0.48 \\
 & Pers. & 0.83 & 0.54 & 0.56 & 0.51 & 0.84 & 0.51 & 0.43 & 0.52 & 0.60 \\
 & Pol. & 0.83 & 0.55 & 0.58 & 0.49 & 0.84 & 0.54 & 0.65 & 0.42 & 0.56 \\
 & Sports & 0.85 & 0.56 & 0.63 & 0.59 & 0.86 & 0.60 & 0.70 & 0.49 & 0.59 \\
\midrule
\multirow{9}{*}{\texttt{Qwen3-VL-8B}} & Cult. & 0.87 & 0.63 & 0.69 & 0.71 & 0.86 & 0.50 & 0.45 & 0.66 & 0.64 \\
 & Food & 0.85 & 0.62 & 0.70 & 0.78 & 0.83 & 0.58 & 0.56 & 0.73 & 0.71 \\
 & Hist. & 0.83 & 0.62 & 0.60 & 0.64 & 0.86 & 0.49 & 0.51 & 0.61 & 0.63 \\
 & M\&M & 0.87 & 0.56 & 0.52 & 0.37 & 0.91 & 0.48 & 0.47 & 0.31 & 0.45 \\
 & Nat. Achv. & 0.87 & 0.73 & 0.72 & 0.65 & 0.88 & 0.60 & 0.56 & 0.56 & 0.69 \\
 & Nature & 0.85 & 0.61 & 0.72 & 0.46 & 0.81 & 0.43 & 0.51 & 0.38 & 0.47 \\
 & Pers. & 0.83 & 0.54 & 0.53 & 0.64 & 0.85 & 0.49 & 0.72 & 0.64 & 0.63 \\
 & Pol. & 0.82 & 0.52 & 0.64 & 0.45 & 0.85 & 0.48 & 0.43 & 0.45 & 0.42 \\
 & Sports & 0.85 & 0.59 & 0.68 & 0.60 & 0.86 & 0.58 & 0.71 & 0.55 & 0.65 \\
\midrule
\multirow{9}{*}{\texttt{GPT-4.1-mini}} & Cult. & 0.87 & 0.64 & 0.71 & 0.72 & 0.87 & 0.66 & 0.75 & 0.68 & 0.76 \\
 & Food & 0.86 & 0.64 & 0.66 & 0.87 & 0.87 & 0.67 & 0.63 & 0.78 & 0.92 \\
 & Hist. & 0.84 & 0.61 & 0.50 & 0.64 & 0.87 & 0.54 & 0.53 & 0.66 & 0.70 \\
 & M\&M & 0.85 & 0.43 & 0.49 & 0.51 & 0.95 & 0.66 & 0.63 & 0.36 & 0.51 \\
 & Nat. Achv. & 0.88 & 0.66 & 0.76 & 0.68 & 0.88 & 0.69 & 0.63 & 0.55 & 0.77 \\
 & Nature & 0.86 & 0.66 & 0.71 & 0.65 & 0.86 & 0.63 & 0.66 & 0.58 & 0.70 \\
 & Pers. & 0.83 & 0.53 & 0.55 & 0.83 & 0.86 & 0.58 & 0.58 & 0.76 & 0.82 \\
 & Pol. & 0.83 & 0.54 & 0.60 & 0.40 & 0.85 & 0.50 & 0.42 & 0.43 & 0.58 \\
 & Sports & 0.86 & 0.55 & 0.61 & 0.56 & 0.87 & 0.61 & 0.63 & 0.46 & 0.75 \\
\bottomrule
\end{tabular*}
\caption{Full results by model and domain for English Language.}
\label{tab:Results-English}
\end{table*}

\begin{table*}[t]
\centering
\small
\scriptsize
\begin{tabular*}{\textwidth}{@{\extracolsep{\fill}} l c | c c c c | c c c c | c }
\toprule
 &  & \multicolumn{4}{c|}{\textbf{Zero-Shot}} & \multicolumn{4}{c|}{\textbf{Few-Shot}} & \textbf{CoT} \\
\textbf{Models} & \textbf{Domain} & \multicolumn{3}{c}{\textbf{Caption}} & \textbf{VQA} & \multicolumn{3}{c}{\textbf{Caption}} & \textbf{VQA} & \textbf{VQA} \\
\cmidrule{3-5}\cmidrule{7-9}
 &  & \textbf{B-F1} & \textbf{Gemini} & \textbf{GPT} & \textbf{Accuracy (\%)} & \textbf{B-F1} & \textbf{Gemini} & \textbf{GPT} & \textbf{Accuracy (\%)} & \textbf{Accuracy (\%)} \\
\midrule
\multirow{9}{*}{\texttt{Gemma3-4B}} & Cult. & 0.71 & 0.57 & 0.47 & 0.72 & 0.69 & 0.52 & 0.51 & 0.53 & 0.58 \\
 & Food & 0.71 & 0.53 & 0.57 & 0.70 & 0.71 & 0.52 & 0.48 & 0.57 & 0.65 \\
 & Hist. & 0.70 & 0.55 & 0.57 & 0.52 & 0.69 & 0.47 & 0.44 & 0.32 & 0.52 \\
 & M\&M & 0.67 & 0.49 & 0.49 & 0.26 & 0.84 & 0.52 & 0.50 & 0.25 & 0.32 \\
 & Nat. Achv. & 0.70 & 0.56 & 0.55 & 0.29 & 0.69 & 0.53 & 0.53 & 0.30 & 0.38 \\
 & Nature & 0.70 & 0.58 & 0.65 & 0.66 & 0.71 & 0.56 & 0.54 & 0.61 & 0.34 \\
 & Pers. & 0.65 & 0.48 & 0.45 & 0.57 & 0.68 & 0.47 & 0.40 & 0.45 & 0.45 \\
 & Pol. & 0.67 & 0.50 & 0.55 & 0.36 & 0.69 & 0.45 & 0.41 & 0.41 & 0.40 \\
 & Sports & 0.68 & 0.54 & 0.54 & 0.47 & 0.69 & 0.46 & 0.51 & 0.37 & 0.44 \\
\midrule
\multirow{9}{*}{\texttt{Gemma3-12B}} & Cult. & 0.71 & 0.57 & 0.60 & 0.61 & 0.70 & 0.51 & 0.53 & 0.62 & 0.68 \\
 & Food & 0.70 & 0.53 & 0.58 & 0.73 & 0.71 & 0.52 & 0.49 & 0.67 & 0.77 \\
 & Hist. & 0.70 & 0.54 & 0.49 & 0.62 & 0.69 & 0.48 & 0.45 & 0.59 & 0.71 \\
 & M\&M & 0.70 & 0.53 & 0.50 & 0.40 & 0.81 & 0.51 & 0.54 & 0.34 & 0.35 \\
 & Nat. Achv. & 0.71 & 0.58 & 0.63 & 0.52 & 0.69 & 0.50 & 0.48 & 0.37 & 0.68 \\
 & Nature & 0.70 & 0.57 & 0.61 & 0.46 & 0.71 & 0.56 & 0.50 & 0.57 & 0.56 \\
 & Pers. & 0.65 & 0.49 & 0.49 & 0.54 & 0.69 & 0.50 & 0.40 & 0.57 & 0.62 \\
 & Pol. & 0.67 & 0.51 & 0.58 & 0.43 & 0.69 & 0.45 & 0.37 & 0.49 & 0.47 \\
 & Sports & 0.68 & 0.54 & 0.63 & 0.47 & 0.69 & 0.49 & 0.43 & 0.45 & 0.52 \\
\midrule
\multirow{9}{*}{\texttt{Gemma3-27B}} & Cult. & 0.72 & 0.58 & 0.57 & 0.71 & 0.71 & 0.53 & 0.50 & 0.73 & 0.73 \\
 & Food & 0.71 & 0.56 & 0.62 & 0.84 & 0.71 & 0.54 & 0.49 & 0.81 & 0.87 \\
 & Hist. & 0.70 & 0.54 & 0.57 & 0.67 & 0.70 & 0.50 & 0.42 & 0.63 & 0.72 \\
 & M\&M & 0.69 & 0.52 & 0.47 & 0.38 & 0.86 & 0.53 & 0.54 & 0.33 & 0.42 \\
 & Nat. Achv. & 0.73 & 0.61 & 0.68 & 0.54 & 0.70 & 0.53 & 0.49 & 0.55 & 0.82 \\
 & Nature & 0.71 & 0.59 & 0.62 & 0.61 & 0.70 & 0.57 & 0.47 & 0.57 & 0.62 \\
 & Pers. & 0.66 & 0.50 & 0.53 & 0.66 & 0.70 & 0.50 & 0.43 & 0.68 & 0.68 \\
 & Pol. & 0.68 & 0.52 & 0.66 & 0.49 & 0.69 & 0.49 & 0.43 & 0.49 & 0.45 \\
 & Sports & 0.71 & 0.58 & 0.68 & 0.56 & 0.69 & 0.51 & 0.51 & 0.53 & 0.53 \\
\midrule
\multirow{9}{*}{\texttt{Qwen2.5-VL-7B}} & Cult. & 0.70 & 0.47 & 0.47 & 0.57 & 0.64 & 0.43 & 0.35 & 0.56 & 0.60 \\
 & Food & 0.71 & 0.41 & 0.35 & 0.64 & 0.65 & 0.36 & 0.31 & 0.67 & 0.63 \\
 & Hist. & 0.70 & 0.50 & 0.51 & 0.58 & 0.68 & 0.38 & 0.42 & 0.48 & 0.55 \\
 & M\&M & 0.70 & 0.39 & 0.46 & 0.29 & 0.80 & 0.43 & 0.38 & 0.22 & 0.36 \\
 & Nat. Achv. & 0.70 & 0.48 & 0.51 & 0.75 & 0.69 & 0.47 & 0.43 & 0.58 & 0.47 \\
 & Nature & 0.71 & 0.49 & 0.44 & 0.41 & 0.65 & 0.48 & 0.36 & 0.50 & 0.45 \\
 & Pers. & 0.65 & 0.43 & 0.30 & 0.33 & 0.69 & 0.48 & 0.41 & 0.43 & 0.45 \\
 & Pol. & 0.67 & 0.46 & 0.42 & 0.40 & 0.69 & 0.47 & 0.45 & 0.34 & 0.41 \\
 & Sports & 0.69 & 0.48 & 0.51 & 0.48 & 0.72 & 0.54 & 0.64 & 0.42 & 0.46 \\
\midrule
\multirow{9}{*}{\texttt{Qwen3-VL-8B}} & Cult. & 0.71 & 0.58 & 0.61 & 0.60 & 0.68 & 0.40 & 0.46 & 0.52 & 0.64 \\
 & Food & 0.71 & 0.53 & 0.47 & 0.72 & 0.67 & 0.34 & 0.31 & 0.65 & 0.75 \\
 & Hist. & 0.71 & 0.50 & 0.55 & 0.66 & 0.68 & 0.43 & 0.41 & 0.71 & 0.70 \\
 & M\&M & 0.67 & 0.49 & 0.42 & 0.37 & 0.78 & 0.41 & 0.35 & 0.29 & 0.37 \\
 & Nat. Achv. & 0.71 & 0.61 & 0.63 & 0.66 & 0.69 & 0.47 & 0.46 & 0.70 & 0.76 \\
 & Nature & 0.70 & 0.60 & 0.57 & 0.36 & 0.60 & 0.38 & 0.31 & 0.40 & 0.47 \\
 & Pers. & 0.65 & 0.49 & 0.45 & 0.48 & 0.69 & 0.50 & 0.46 & 0.48 & 0.49 \\
 & Pol. & 0.66 & 0.51 & 0.60 & 0.43 & 0.69 & 0.47 & 0.47 & 0.45 & 0.43 \\
 & Sports & 0.69 & 0.55 & 0.64 & 0.53 & 0.72 & 0.50 & 0.48 & 0.50 & 0.47 \\
\midrule
\multirow{9}{*}{\texttt{GPT-4.1-mini}} & Cult. & 0.72 & 0.59 & 0.60 & 0.64 & 0.72 & 0.61 & 0.62 & 0.61 & 0.71 \\
 & Food & 0.72 & 0.57 & 0.64 & 0.85 & 0.73 & 0.59 & 0.62 & 0.70 & 0.83 \\
 & Hist. & 0.71 & 0.55 & 0.67 & 0.70 & 0.72 & 0.55 & 0.59 & 0.59 & 0.68 \\
 & M\&M & 0.69 & 0.52 & 0.49 & 0.38 & 0.87 & 0.58 & 0.60 & 0.37 & 0.42 \\
 & Nat. Achv. & 0.72 & 0.61 & 0.67 & 0.59 & 0.72 & 0.56 & 0.70 & 0.64 & 0.73 \\
 & Nature & 0.71 & 0.62 & 0.62 & 0.49 & 0.71 & 0.59 & 0.56 & 0.38 & 0.53 \\
 & Pers. & 0.65 & 0.50 & 0.50 & 0.60 & 0.71 & 0.53 & 0.57 & 0.50 & 0.72 \\
 & Pol. & 0.68 & 0.53 & 0.62 & 0.46 & 0.69 & 0.52 & 0.58 & 0.53 & 0.49 \\
 & Sports & 0.69 & 0.56 & 0.66 & 0.44 & 0.74 & 0.54 & 0.67 & 0.44 & 0.53 \\
\bottomrule
\end{tabular*}
\caption{Full results by model and domain for Hindi Language.}
\label{tab:Results-Hindi}
\end{table*}

\begin{table*}[t]
\centering
\small
\scriptsize
\begin{tabular*}{\textwidth}{@{\extracolsep{\fill}} l c | c c c c | c c c c | c }
\toprule
 &  & \multicolumn{4}{c|}{\textbf{Zero-Shot}} & \multicolumn{4}{c|}{\textbf{Few-Shot}} & \textbf{CoT} \\
\textbf{Models} & \textbf{Domain} & \multicolumn{3}{c}{\textbf{Caption}} & \textbf{VQA} & \multicolumn{3}{c}{\textbf{Caption}} & \textbf{VQA} & \textbf{VQA} \\
\cmidrule{3-5}\cmidrule{7-9}
 &  & \textbf{B-F1} & \textbf{Gemini} & \textbf{GPT} & \textbf{Accuracy (\%)} & \textbf{B-F1} & \textbf{Gemini} & \textbf{GPT} & \textbf{Accuracy (\%)} & \textbf{Accuracy (\%)} \\
\midrule
\multirow{9}{*}{\texttt{Gemma3-4B}} & Cult. & 0.70 & 0.48 & 0.41 & 0.57 & 0.69 & 0.48 & 0.50 & 0.34 & 0.48 \\
 & Food & 0.70 & 0.46 & 0.32 & 0.63 & 0.70 & 0.47 & 0.50 & 0.52 & 0.56 \\
 & Hist. & 0.69 & 0.49 & 0.37 & 0.51 & 0.70 & 0.44 & 0.43 & 0.37 & 0.45 \\
 & M\&M & 0.69 & 0.45 & 0.42 & 0.19 & 0.85 & 0.52 & 0.52 & 0.24 & 0.28 \\
 & Nat. Achv. & 0.69 & 0.49 & 0.43 & 0.31 & 0.69 & 0.46 & 0.42 & 0.26 & 0.38 \\
 & Nature & 0.70 & 0.49 & 0.53 & 0.67 & 0.69 & 0.51 & 0.50 & 0.57 & 0.41 \\
 & Pers. & 0.65 & 0.45 & 0.37 & 0.52 & 0.68 & 0.48 & 0.43 & 0.43 & 0.47 \\
 & Pol. & 0.67 & 0.47 & 0.50 & 0.37 & 0.70 & 0.47 & 0.41 & 0.37 & 0.41 \\
 & Sports & 0.68 & 0.47 & 0.51 & 0.39 & 0.70 & 0.48 & 0.47 & 0.34 & 0.52 \\
\midrule
\multirow{9}{*}{\texttt{Gemma3-12B}} & Cult. & 0.71 & 0.54 & 0.58 & 0.56 & 0.69 & 0.50 & 0.48 & 0.64 & 0.60 \\
 & Food & 0.69 & 0.51 & 0.59 & 0.67 & 0.70 & 0.51 & 0.45 & 0.65 & 0.68 \\
 & Hist. & 0.70 & 0.53 & 0.53 & 0.59 & 0.70 & 0.48 & 0.40 & 0.59 & 0.63 \\
 & M\&M & 0.72 & 0.52 & 0.48 & 0.26 & 0.83 & 0.53 & 0.52 & 0.28 & 0.32 \\
 & Nat. Achv. & 0.70 & 0.55 & 0.63 & 0.55 & 0.70 & 0.50 & 0.48 & 0.43 & 0.64 \\
 & Nature & 0.70 & 0.56 & 0.53 & 0.36 & 0.70 & 0.56 & 0.48 & 0.47 & 0.46 \\
 & Pers. & 0.66 & 0.49 & 0.46 & 0.47 & 0.69 & 0.50 & 0.48 & 0.61 & 0.55 \\
 & Pol. & 0.68 & 0.53 & 0.55 & 0.41 & 0.70 & 0.47 & 0.42 & 0.50 & 0.50 \\
 & Sports & 0.68 & 0.54 & 0.59 & 0.42 & 0.70 & 0.50 & 0.45 & 0.52 & 0.50 \\
\midrule
\multirow{9}{*}{\texttt{Gemma3-27B}} & Cult. & 0.71 & 0.58 & 0.61 & 0.61 & 0.70 & 0.52 & 0.56 & 0.68 & 0.68 \\
 & Food & 0.71 & 0.56 & 0.65 & 0.72 & 0.71 & 0.53 & 0.51 & 0.77 & 0.78 \\
 & Hist. & 0.71 & 0.58 & 0.58 & 0.65 & 0.70 & 0.49 & 0.42 & 0.64 & 0.70 \\
 & M\&M & 0.73 & 0.51 & 0.52 & 0.32 & 0.87 & 0.52 & 0.55 & 0.32 & 0.35 \\
 & Nat. Achv. & 0.71 & 0.61 & 0.68 & 0.58 & 0.70 & 0.54 & 0.42 & 0.57 & 0.83 \\
 & Nature & 0.70 & 0.59 & 0.58 & 0.41 & 0.70 & 0.56 & 0.53 & 0.57 & 0.51 \\
 & Pers. & 0.66 & 0.49 & 0.54 & 0.56 & 0.69 & 0.51 & 0.41 & 0.61 & 0.59 \\
 & Pol. & 0.68 & 0.54 & 0.61 & 0.45 & 0.71 & 0.49 & 0.46 & 0.46 & 0.48 \\
 & Sports & 0.71 & 0.60 & 0.64 & 0.61 & 0.73 & 0.49 & 0.51 & 0.60 & 0.61 \\
\midrule
\multirow{9}{*}{\texttt{Qwen2.5-VL-7B}} & Cult. & 0.69 & 0.43 & 0.42 & 0.58 & 0.65 & 0.42 & 0.30 & 0.54 & 0.55 \\
 & Food & 0.69 & 0.34 & 0.41 & 0.66 & 0.64 & 0.32 & 0.36 & 0.56 & 0.60 \\
 & Hist. & 0.70 & 0.46 & 0.45 & 0.58 & 0.69 & 0.41 & 0.34 & 0.54 & 0.54 \\
 & M\&M & 0.66 & 0.35 & 0.33 & 0.24 & 0.80 & 0.41 & 0.38 & 0.29 & 0.35 \\
 & Nat. Achv. & 0.68 & 0.40 & 0.36 & 0.67 & 0.71 & 0.46 & 0.31 & 0.65 & 0.55 \\
 & Nature & 0.69 & 0.43 & 0.43 & 0.49 & 0.56 & 0.32 & 0.35 & 0.24 & 0.46 \\
 & Pers. & 0.65 & 0.38 & 0.31 & 0.43 & 0.69 & 0.45 & 0.39 & 0.51 & 0.44 \\
 & Pol. & 0.68 & 0.45 & 0.47 & 0.42 & 0.70 & 0.47 & 0.37 & 0.35 & 0.36 \\
 & Sports & 0.67 & 0.46 & 0.42 & 0.44 & 0.72 & 0.44 & 0.48 & 0.29 & 0.50 \\
\midrule
\multirow{9}{*}{\texttt{Qwen3-VL-8B}} & Cult. & 0.70 & 0.53 & 0.50 & 0.53 & 0.67 & 0.40 & 0.42 & 0.59 & 0.57 \\
 & Food & 0.69 & 0.49 & 0.39 & 0.66 & 0.66 & 0.37 & 0.31 & 0.60 & 0.64 \\
 & Hist. & 0.71 & 0.54 & 0.56 & 0.66 & 0.67 & 0.38 & 0.37 & 0.60 & 0.68 \\
 & M\&M & 0.68 & 0.37 & 0.37 & 0.25 & 0.82 & 0.47 & 0.37 & 0.39 & 0.26 \\
 & Nat. Achv. & 0.71 & 0.57 & 0.58 & 0.69 & 0.69 & 0.50 & 0.50 & 0.50 & 0.70 \\
 & Nature & 0.69 & 0.52 & 0.53 & 0.37 & 0.59 & 0.35 & 0.33 & 0.46 & 0.38 \\
 & Pers. & 0.64 & 0.43 & 0.33 & 0.42 & 0.69 & 0.51 & 0.46 & 0.45 & 0.44 \\
 & Pol. & 0.67 & 0.47 & 0.48 & 0.39 & 0.69 & 0.47 & 0.39 & 0.41 & 0.44 \\
 & Sports & 0.69 & 0.58 & 0.57 & 0.51 & 0.72 & 0.47 & 0.47 & 0.46 & 0.49 \\
\midrule
\multirow{9}{*}{\texttt{GPT-4.1-mini}} & Cult. & 0.72 & 0.57 & 0.62 & 0.64 & 0.72 & 0.57 & 0.61 & 0.60 & 0.69 \\
 & Food & 0.71 & 0.54 & 0.56 & 0.69 & 0.72 & 0.56 & 0.64 & 0.61 & 0.78 \\
 & Hist. & 0.72 & 0.55 & 0.56 & 0.60 & 0.71 & 0.56 & 0.62 & 0.56 & 0.63 \\
 & M\&M & 0.70 & 0.50 & 0.49 & 0.43 & 0.87 & 0.55 & 0.57 & 0.35 & 0.43 \\
 & Nat. Achv. & 0.71 & 0.60 & 0.68 & 0.58 & 0.71 & 0.58 & 0.60 & 0.56 & 0.73 \\
 & Nature & 0.71 & 0.62 & 0.62 & 0.41 & 0.71 & 0.59 & 0.61 & 0.35 & 0.56 \\
 & Pers. & 0.66 & 0.49 & 0.50 & 0.50 & 0.69 & 0.50 & 0.52 & 0.45 & 0.73 \\
 & Pol. & 0.68 & 0.52 & 0.60 & 0.39 & 0.70 & 0.51 & 0.54 & 0.46 & 0.56 \\
 & Sports & 0.70 & 0.57 & 0.66 & 0.47 & 0.74 & 0.55 & 0.58 & 0.45 & 0.56 \\
\bottomrule
\end{tabular*}
\caption{Full results by model and domain for Urdu Language.}
\label{tab:Results-Urdu}
\end{table*}

\begin{table*}[t]
\centering
\small
\scriptsize
\begin{tabular*}{\textwidth}{@{\extracolsep{\fill}} l c | c c c c | c c c c | c }
\toprule
 &  & \multicolumn{4}{c|}{\textbf{Zero-Shot}} & \multicolumn{4}{c|}{\textbf{Few-Shot}} & \textbf{CoT} \\
\textbf{Models} & \textbf{Domain} & \multicolumn{3}{c}{\textbf{Caption}} & \textbf{VQA} & \multicolumn{3}{c}{\textbf{Caption}} & \textbf{VQA} & \textbf{VQA} \\
\cmidrule{3-5}\cmidrule{7-9}
 &  & \textbf{B-F1} & \textbf{Gemini} & \textbf{GPT} & \textbf{Accuracy (\%)} & \textbf{B-F1} & \textbf{Gemini} & \textbf{GPT} & \textbf{Accuracy (\%)} & \textbf{Accuracy (\%)} \\
\midrule
\multirow{9}{*}{\texttt{Gemma3-4B}} & Cult. & 0.68 & 0.49 & 0.46 & 0.61 & 0.67 & 0.48 & 0.52 & 0.38 & 0.48 \\
 & Food & 0.69 & 0.45 & 0.51 & 0.74 & 0.69 & 0.53 & 0.48 & 0.53 & 0.75 \\
 & Hist. & 0.65 & 0.47 & 0.43 & 0.48 & 0.66 & 0.43 & 0.37 & 0.33 & 0.48 \\
 & M\&M & 0.64 & 0.43 & 0.38 & 0.29 & 0.75 & 0.47 & 0.48 & 0.29 & 0.29 \\
 & Nat. Achv. & 0.65 & 0.45 & 0.44 & 0.26 & 0.65 & 0.49 & 0.42 & 0.25 & 0.36 \\
 & Nature & 0.68 & 0.52 & 0.61 & 0.67 & 0.68 & 0.56 & 0.52 & 0.60 & 0.45 \\
 & Pers. & 0.63 & 0.44 & 0.43 & 0.56 & 0.64 & 0.45 & 0.39 & 0.46 & 0.52 \\
 & Pol. & 0.63 & 0.45 & 0.48 & 0.41 & 0.66 & 0.45 & 0.42 & 0.37 & 0.40 \\
 & Sports & 0.64 & 0.48 & 0.52 & 0.42 & 0.64 & 0.47 & 0.44 & 0.44 & 0.63 \\
\midrule
\multirow{9}{*}{\texttt{Gemma3-12B}} & Cult. & 0.69 & 0.47 & 0.45 & 0.64 & 0.68 & 0.46 & 0.39 & 0.63 & 0.70 \\
 & Food & 0.70 & 0.44 & 0.43 & 0.86 & 0.70 & 0.49 & 0.45 & 0.76 & 0.95 \\
 & Hist. & 0.67 & 0.42 & 0.42 & 0.63 & 0.67 & 0.41 & 0.41 & 0.55 & 0.75 \\
 & M\&M & 0.66 & 0.39 & 0.39 & 0.32 & 0.80 & 0.49 & 0.51 & 0.37 & 0.35 \\
 & Nat. Achv. & 0.66 & 0.44 & 0.43 & 0.49 & 0.67 & 0.46 & 0.48 & 0.50 & 0.42 \\
 & Nature & 0.69 & 0.46 & 0.46 & 0.46 & 0.67 & 0.47 & 0.39 & 0.51 & 0.52 \\
 & Pers. & 0.64 & 0.42 & 0.40 & 0.67 & 0.66 & 0.41 & 0.36 & 0.65 & 0.67 \\
 & Pol. & 0.64 & 0.42 & 0.40 & 0.51 & 0.67 & 0.42 & 0.42 & 0.55 & 0.50 \\
 & Sports & 0.65 & 0.42 & 0.42 & 0.48 & 0.66 & 0.42 & 0.35 & 0.52 & 0.47 \\
\midrule
\multirow{9}{*}{\texttt{Gemma3-27B}} & Cult. & 0.69 & 0.47 & 0.44 & 0.70 & 0.69 & 0.50 & 0.48 & 0.72 & 0.69 \\
 & Food & 0.70 & 0.51 & 0.46 & 0.88 & 0.72 & 0.54 & 0.42 & 0.87 & 0.90 \\
 & Hist. & 0.66 & 0.41 & 0.40 & 0.67 & 0.70 & 0.47 & 0.42 & 0.64 & 0.71 \\
 & M\&M & 0.64 & 0.42 & 0.38 & 0.36 & 0.78 & 0.48 & 0.48 & 0.31 & 0.39 \\
 & Nat. Achv. & 0.66 & 0.44 & 0.49 & 0.51 & 0.68 & 0.48 & 0.43 & 0.53 & 0.75 \\
 & Nature & 0.68 & 0.45 & 0.54 & 0.61 & 0.69 & 0.51 & 0.53 & 0.63 & 0.75 \\
 & Pers. & 0.63 & 0.42 & 0.46 & 0.82 & 0.67 & 0.45 & 0.40 & 0.80 & 0.75 \\
 & Pol. & 0.63 & 0.39 & 0.38 & 0.53 & 0.68 & 0.45 & 0.41 & 0.51 & 0.58 \\
 & Sports & 0.66 & 0.44 & 0.40 & 0.65 & 0.66 & 0.44 & 0.43 & 0.65 & 0.80 \\
\midrule
\multirow{9}{*}{\texttt{Qwen2.5-VL-7B}} & Cult. & 0.68 & 0.38 & 0.40 & 0.48 & 0.66 & 0.34 & 0.32 & 0.45 & 0.51 \\
 & Food & 0.69 & 0.34 & 0.37 & 0.61 & 0.64 & 0.33 & 0.35 & 0.51 & 0.67 \\
 & Hist. & 0.69 & 0.44 & 0.40 & 0.57 & 0.65 & 0.34 & 0.37 & 0.44 & 0.61 \\
 & M\&M & 0.68 & 0.34 & 0.39 & 0.21 & 0.78 & 0.41 & 0.44 & 0.26 & 0.31 \\
 & Nat. Achv. & 0.67 & 0.30 & 0.36 & 0.66 & 0.69 & 0.44 & 0.43 & 0.68 & 0.55 \\
 & Nature & 0.68 & 0.36 & 0.41 & 0.40 & 0.55 & 0.31 & 0.31 & 0.66 & 0.35 \\
 & Pers. & 0.64 & 0.36 & 0.41 & 0.36 & 0.68 & 0.45 & 0.43 & 0.42 & 0.52 \\
 & Pol. & 0.65 & 0.38 & 0.43 & 0.38 & 0.69 & 0.45 & 0.47 & 0.32 & 0.51 \\
 & Sports & 0.68 & 0.35 & 0.38 & 0.45 & 0.70 & 0.44 & 0.42 & 0.39 & 0.53 \\
\midrule
\multirow{9}{*}{\texttt{Qwen3-VL-8B}} & Cult. & 0.67 & 0.43 & 0.38 & 0.59 & 0.66 & 0.34 & 0.34 & 0.59 & 0.70 \\
 & Food & 0.68 & 0.40 & 0.42 & 0.72 & 0.69 & 0.42 & 0.46 & 0.73 & 0.74 \\
 & Hist. & 0.67 & 0.46 & 0.36 & 0.71 & 0.67 & 0.38 & 0.38 & 0.71 & 0.65 \\
 & M\&M & 0.67 & 0.35 & 0.38 & 0.36 & 0.69 & 0.37 & 0.41 & 0.37 & 0.28 \\
 & Nat. Achv. & 0.66 & 0.41 & 0.41 & 0.67 & 0.65 & 0.35 & 0.37 & 0.59 & 0.57 \\
 & Nature & 0.68 & 0.43 & 0.38 & 0.53 & 0.65 & 0.34 & 0.35 & 0.31 & 0.48 \\
 & Pers. & 0.63 & 0.42 & 0.43 & 0.52 & 0.65 & 0.43 & 0.45 & 0.54 & 0.52 \\
 & Pol. & 0.64 & 0.45 & 0.51 & 0.45 & 0.66 & 0.42 & 0.45 & 0.47 & 0.46 \\
 & Sports & 0.65 & 0.47 & 0.47 & 0.48 & 0.64 & 0.38 & 0.38 & 0.53 & 0.43 \\
\midrule
\multirow{9}{*}{\texttt{GPT-4.1-mini}} & Cult. & 0.70 & 0.46 & 0.44 & 0.64 & 0.70 & 0.50 & 0.53 & 0.59 & 0.75 \\
 & Food & 0.72 & 0.45 & 0.47 & 0.84 & 0.73 & 0.54 & 0.51 & 0.73 & 0.92 \\
 & Hist. & 0.69 & 0.44 & 0.48 & 0.56 & 0.70 & 0.50 & 0.54 & 0.54 & 0.68 \\
 & M\&M & 0.68 & 0.37 & 0.43 & 0.45 & 0.90 & 0.56 & 0.57 & 0.41 & 0.45 \\
 & Nat. Achv. & 0.69 & 0.42 & 0.46 & 0.61 & 0.69 & 0.52 & 0.54 & 0.51 & 0.85 \\
 & Nature & 0.69 & 0.41 & 0.39 & 0.48 & 0.71 & 0.51 & 0.46 & 0.51 & 0.62 \\
 & Pers. & 0.65 & 0.36 & 0.42 & 0.59 & 0.69 & 0.52 & 0.52 & 0.53 & 0.73 \\
 & Pol. & 0.65 & 0.37 & 0.37 & 0.49 & 0.69 & 0.45 & 0.44 & 0.43 & 0.56 \\
 & Sports & 0.66 & 0.43 & 0.45 & 0.52 & 0.71 & 0.45 & 0.48 & 0.53 & 0.71 \\
\bottomrule
\end{tabular*}
\caption{Full results by model and domain for Barishal Dialect.}
\label{tab:Results-Barishal}
\end{table*}

\begin{table*}[t]
\centering
\small
\scriptsize
\begin{tabular*}{\textwidth}{@{\extracolsep{\fill}} l c | c c c c | c c c c | c }
\toprule
 &  & \multicolumn{4}{c|}{\textbf{Zero-Shot}} & \multicolumn{4}{c|}{\textbf{Few-Shot}} & \textbf{CoT} \\
\textbf{Models} & \textbf{Domain} & \multicolumn{3}{c}{\textbf{Caption}} & \textbf{VQA} & \multicolumn{3}{c}{\textbf{Caption}} & \textbf{VQA} & \textbf{VQA} \\
\cmidrule{3-5}\cmidrule{7-9}
 &  & \textbf{B-F1} & \textbf{Gemini} & \textbf{GPT} & \textbf{Accuracy (\%)} & \textbf{B-F1} & \textbf{Gemini} & \textbf{GPT} & \textbf{Accuracy (\%)} & \textbf{Accuracy (\%)} \\
\midrule
\multirow{9}{*}{\texttt{Gemma3-4B}} & Cult. & 0.67 & 0.45 & 0.43 & 0.65 & 0.68 & 0.45 & 0.38 & 0.42 & 0.62 \\
 & Food & 0.69 & 0.48 & 0.47 & 0.75 & 0.69 & 0.51 & 0.49 & 0.62 & 0.77 \\
 & Hist. & 0.65 & 0.40 & 0.44 & 0.47 & 0.68 & 0.45 & 0.39 & 0.33 & 0.41 \\
 & M\&M & 0.65 & 0.42 & 0.35 & 0.25 & 0.78 & 0.51 & 0.51 & 0.25 & 0.30 \\
 & Nat. Achv. & 0.65 & 0.44 & 0.45 & 0.28 & 0.67 & 0.43 & 0.34 & 0.27 & 0.31 \\
 & Nature & 0.69 & 0.51 & 0.45 & 0.63 & 0.69 & 0.53 & 0.48 & 0.57 & 0.49 \\
 & Pers. & 0.63 & 0.40 & 0.35 & 0.52 & 0.66 & 0.42 & 0.36 & 0.44 & 0.51 \\
 & Pol. & 0.63 & 0.41 & 0.50 & 0.38 & 0.67 & 0.41 & 0.33 & 0.35 & 0.53 \\
 & Sports & 0.63 & 0.45 & 0.40 & 0.45 & 0.65 & 0.45 & 0.46 & 0.27 & 0.61 \\
\midrule
\multirow{9}{*}{\texttt{Gemma3-12B}} & Cult. & 0.68 & 0.45 & 0.47 & 0.65 & 0.68 & 0.45 & 0.41 & 0.61 & 0.69 \\
 & Food & 0.70 & 0.48 & 0.56 & 0.85 & 0.71 & 0.51 & 0.48 & 0.71 & 0.83 \\
 & Hist. & 0.67 & 0.40 & 0.42 & 0.62 & 0.68 & 0.40 & 0.33 & 0.59 & 0.74 \\
 & M\&M & 0.66 & 0.42 & 0.39 & 0.32 & 0.83 & 0.51 & 0.52 & 0.37 & 0.28 \\
 & Nat. Achv. & 0.66 & 0.47 & 0.56 & 0.54 & 0.67 & 0.45 & 0.38 & 0.49 & 0.62 \\
 & Nature & 0.68 & 0.43 & 0.44 & 0.51 & 0.68 & 0.46 & 0.46 & 0.45 & 0.59 \\
 & Pers. & 0.64 & 0.41 & 0.30 & 0.69 & 0.66 & 0.41 & 0.39 & 0.64 & 0.83 \\
 & Pol. & 0.64 & 0.41 & 0.40 & 0.47 & 0.68 & 0.40 & 0.42 & 0.49 & 0.54 \\
 & Sports & 0.65 & 0.45 & 0.51 & 0.52 & 0.66 & 0.45 & 0.34 & 0.42 & 0.50 \\
\midrule
\multirow{9}{*}{\texttt{Gemma3-27B}} & Cult. & 0.68 & 0.40 & 0.40 & 0.68 & 0.69 & 0.47 & 0.48 & 0.73 & 0.77 \\
 & Food & 0.70 & 0.44 & 0.52 & 0.90 & 0.71 & 0.50 & 0.49 & 0.88 & 0.85 \\
 & Hist. & 0.66 & 0.38 & 0.49 & 0.67 & 0.69 & 0.46 & 0.40 & 0.67 & 0.77 \\
 & M\&M & 0.66 & 0.35 & 0.37 & 0.34 & 0.79 & 0.50 & 0.54 & 0.30 & 0.34 \\
 & Nat. Achv. & 0.66 & 0.45 & 0.55 & 0.52 & 0.68 & 0.43 & 0.36 & 0.46 & 0.69 \\
 & Nature & 0.69 & 0.43 & 0.53 & 0.59 & 0.70 & 0.50 & 0.62 & 0.57 & 0.70 \\
 & Pers. & 0.63 & 0.35 & 0.34 & 0.78 & 0.67 & 0.46 & 0.40 & 0.82 & 0.80 \\
 & Pol. & 0.64 & 0.36 & 0.44 & 0.55 & 0.68 & 0.46 & 0.44 & 0.48 & 0.48 \\
 & Sports & 0.66 & 0.38 & 0.49 & 0.66 & 0.67 & 0.41 & 0.39 & 0.66 & 0.71 \\
\midrule
\multirow{9}{*}{\texttt{Qwen2.5-VL-7B}} & Cult. & 0.67 & 0.36 & 0.34 & 0.55 & 0.63 & 0.34 & 0.36 & 0.49 & 0.60 \\
 & Food & 0.69 & 0.31 & 0.36 & 0.68 & 0.65 & 0.29 & 0.29 & 0.49 & 0.66 \\
 & Hist. & 0.69 & 0.42 & 0.38 & 0.62 & 0.67 & 0.37 & 0.35 & 0.57 & 0.50 \\
 & M\&M & 0.67 & 0.35 & 0.31 & 0.18 & 0.79 & 0.45 & 0.39 & 0.19 & 0.18 \\
 & Nat. Achv. & 0.67 & 0.34 & 0.37 & 0.66 & 0.70 & 0.39 & 0.39 & 0.60 & 0.43 \\
 & Nature & 0.68 & 0.37 & 0.36 & 0.51 & 0.54 & 0.28 & 0.31 & 0.38 & 0.37 \\
 & Pers. & 0.63 & 0.34 & 0.33 & 0.36 & 0.67 & 0.41 & 0.38 & 0.44 & 0.54 \\
 & Pol. & 0.65 & 0.39 & 0.32 & 0.36 & 0.69 & 0.44 & 0.40 & 0.31 & 0.29 \\
 & Sports & 0.67 & 0.37 & 0.33 & 0.52 & 0.69 & 0.45 & 0.40 & 0.43 & 0.58 \\
\midrule
\multirow{9}{*}{\texttt{Qwen3-VL-8B}} & Cult. & 0.67 & 0.45 & 0.46 & 0.60 & 0.67 & 0.42 & 0.37 & 0.57 & 0.67 \\
 & Food & 0.68 & 0.40 & 0.35 & 0.69 & 0.69 & 0.43 & 0.39 & 0.67 & 0.74 \\
 & Hist. & 0.67 & 0.44 & 0.34 & 0.68 & 0.67 & 0.41 & 0.35 & 0.68 & 0.69 \\
 & M\&M & 0.66 & 0.36 & 0.41 & 0.29 & 0.68 & 0.34 & 0.34 & 0.25 & 0.28 \\
 & Nat. Achv. & 0.65 & 0.41 & 0.32 & 0.59 & 0.66 & 0.42 & 0.36 & 0.53 & 0.59 \\
 & Nature & 0.67 & 0.44 & 0.45 & 0.57 & 0.64 & 0.33 & 0.33 & 0.48 & 0.45 \\
 & Pers. & 0.63 & 0.43 & 0.46 & 0.57 & 0.64 & 0.41 & 0.35 & 0.46 & 0.50 \\
 & Pol. & 0.64 & 0.44 & 0.49 & 0.48 & 0.66 & 0.41 & 0.34 & 0.41 & 0.38 \\
 & Sports & 0.64 & 0.42 & 0.46 & 0.54 & 0.66 & 0.39 & 0.33 & 0.44 & 0.56 \\
\midrule
\multirow{9}{*}{\texttt{GPT-4.1-mini}} & Cult. & 0.69 & 0.41 & 0.50 & 0.70 & 0.70 & 0.45 & 0.38 & 0.58 & 0.73 \\
 & Food & 0.72 & 0.46 & 0.50 & 0.84 & 0.74 & 0.52 & 0.49 & 0.75 & 0.91 \\
 & Hist. & 0.69 & 0.41 & 0.45 & 0.64 & 0.70 & 0.46 & 0.43 & 0.64 & 0.66 \\
 & M\&M & 0.68 & 0.42 & 0.38 & 0.48 & 0.89 & 0.61 & 0.63 & 0.42 & 0.43 \\
 & Nat. Achv. & 0.69 & 0.46 & 0.43 & 0.59 & 0.70 & 0.54 & 0.50 & 0.57 & 0.80 \\
 & Nature & 0.70 & 0.39 & 0.37 & 0.58 & 0.71 & 0.46 & 0.53 & 0.44 & 0.57 \\
 & Pers. & 0.65 & 0.34 & 0.35 & 0.67 & 0.69 & 0.52 & 0.45 & 0.54 & 0.78 \\
 & Pol. & 0.66 & 0.36 & 0.38 & 0.50 & 0.68 & 0.45 & 0.46 & 0.46 & 0.59 \\
 & Sports & 0.67 & 0.41 & 0.38 & 0.50 & 0.70 & 0.48 & 0.47 & 0.44 & 0.66 \\
\bottomrule
\end{tabular*}
\caption{Full results by model and domain for Chittagong Dialect.}
\label{tab:Results-Chittagong}
\end{table*}

\begin{table*}[t]
\centering
\small
\scriptsize
\begin{tabular*}{\textwidth}{@{\extracolsep{\fill}} l c | c c c c | c c c c | c }
\toprule
 &  & \multicolumn{4}{c|}{\textbf{Zero-Shot}} & \multicolumn{4}{c|}{\textbf{Few-Shot}} & \textbf{CoT} \\
\textbf{Models} & \textbf{Domain} & \multicolumn{3}{c}{\textbf{Caption}} & \textbf{VQA} & \multicolumn{3}{c}{\textbf{Caption}} & \textbf{VQA} & \textbf{VQA} \\
\cmidrule{3-5}\cmidrule{7-9}
 &  & \textbf{B-F1} & \textbf{Gemini} & \textbf{GPT} & \textbf{Accuracy (\%)} & \textbf{B-F1} & \textbf{Gemini} & \textbf{GPT} & \textbf{Accuracy (\%)} & \textbf{Accuracy (\%)} \\
\midrule
\multirow{9}{*}{\texttt{Gemma3-4B}} & Cult. & 0.69 & 0.48 & 0.41 & 0.65 & 0.68 & 0.49 & 0.55 & 0.49 & 0.51 \\
 & Food & 0.69 & 0.45 & 0.39 & 0.74 & 0.69 & 0.50 & 0.49 & 0.61 & 0.72 \\
 & Hist. & 0.66 & 0.40 & 0.38 & 0.46 & 0.68 & 0.42 & 0.34 & 0.37 & 0.51 \\
 & M\&M & 0.65 & 0.41 & 0.39 & 0.23 & 0.79 & 0.48 & 0.47 & 0.28 & 0.31 \\
 & Nat. Achv. & 0.66 & 0.49 & 0.41 & 0.31 & 0.68 & 0.45 & 0.44 & 0.20 & 0.43 \\
 & Nature & 0.70 & 0.56 & 0.56 & 0.63 & 0.69 & 0.56 & 0.53 & 0.69 & 0.49 \\
 & Pers. & 0.64 & 0.45 & 0.38 & 0.51 & 0.68 & 0.47 & 0.41 & 0.50 & 0.53 \\
 & Pol. & 0.65 & 0.45 & 0.37 & 0.40 & 0.68 & 0.43 & 0.37 & 0.41 & 0.46 \\
 & Sports & 0.64 & 0.42 & 0.44 & 0.50 & 0.66 & 0.46 & 0.44 & 0.45 & 0.45 \\
\midrule
\multirow{9}{*}{\texttt{Gemma3-12B}} & Cult. & 0.69 & 0.42 & 0.37 & 0.67 & 0.68 & 0.44 & 0.42 & 0.65 & 0.70 \\
 & Food & 0.71 & 0.44 & 0.44 & 0.86 & 0.71 & 0.49 & 0.47 & 0.77 & 0.88 \\
 & Hist. & 0.67 & 0.37 & 0.34 & 0.63 & 0.68 & 0.42 & 0.41 & 0.60 & 0.62 \\
 & M\&M & 0.67 & 0.37 & 0.33 & 0.35 & 0.75 & 0.54 & 0.55 & 0.34 & 0.36 \\
 & Nat. Achv. & 0.66 & 0.42 & 0.41 & 0.55 & 0.68 & 0.45 & 0.45 & 0.44 & 0.44 \\
 & Nature & 0.69 & 0.43 & 0.37 & 0.48 & 0.68 & 0.45 & 0.48 & 0.46 & 0.60 \\
 & Pers. & 0.64 & 0.41 & 0.35 & 0.68 & 0.65 & 0.41 & 0.31 & 0.67 & 0.69 \\
 & Pol. & 0.65 & 0.40 & 0.34 & 0.54 & 0.67 & 0.41 & 0.43 & 0.56 & 0.60 \\
 & Sports & 0.65 & 0.40 & 0.32 & 0.53 & 0.65 & 0.42 & 0.45 & 0.52 & 0.82 \\
\midrule
\multirow{9}{*}{\texttt{Gemma3-27B}} & Cult. & 0.70 & 0.47 & 0.47 & 0.73 & 0.69 & 0.48 & 0.46 & 0.69 & 0.79 \\
 & Food & 0.71 & 0.47 & 0.55 & 0.90 & 0.72 & 0.52 & 0.50 & 0.91 & 0.90 \\
 & Hist. & 0.68 & 0.44 & 0.49 & 0.66 & 0.69 & 0.45 & 0.40 & 0.69 & 0.70 \\
 & M\&M & 0.68 & 0.41 & 0.40 & 0.35 & 0.80 & 0.51 & 0.49 & 0.32 & 0.46 \\
 & Nat. Achv. & 0.67 & 0.47 & 0.55 & 0.48 & 0.67 & 0.43 & 0.38 & 0.53 & 0.85 \\
 & Nature & 0.69 & 0.47 & 0.46 & 0.62 & 0.70 & 0.51 & 0.49 & 0.68 & 0.67 \\
 & Pers. & 0.65 & 0.39 & 0.36 & 0.80 & 0.68 & 0.46 & 0.38 & 0.85 & 0.83 \\
 & Pol. & 0.65 & 0.39 & 0.46 & 0.51 & 0.69 & 0.46 & 0.41 & 0.47 & 0.50 \\
 & Sports & 0.67 & 0.43 & 0.53 & 0.65 & 0.66 & 0.42 & 0.38 & 0.58 & 0.67 \\
\midrule
\multirow{9}{*}{\texttt{Qwen2.5-VL-7B}} & Cult. & 0.68 & 0.37 & 0.38 & 0.51 & 0.69 & 0.36 & 0.40 & 0.49 & 0.53 \\
 & Food & 0.69 & 0.33 & 0.33 & 0.66 & 0.71 & 0.41 & 0.44 & 0.54 & 0.68 \\
 & Hist. & 0.69 & 0.42 & 0.47 & 0.60 & 0.70 & 0.44 & 0.40 & 0.55 & 0.60 \\
 & M\&M & 0.68 & 0.35 & 0.35 & 0.21 & 0.83 & 0.53 & 0.56 & 0.29 & 0.23 \\
 & Nat. Achv. & 0.67 & 0.41 & 0.40 & 0.69 & 0.70 & 0.42 & 0.46 & 0.48 & 0.56 \\
 & Nature & 0.68 & 0.41 & 0.40 & 0.37 & 0.71 & 0.44 & 0.34 & 0.58 & 0.34 \\
 & Pers. & 0.64 & 0.36 & 0.35 & 0.46 & 0.67 & 0.41 & 0.38 & 0.44 & 0.43 \\
 & Pol. & 0.65 & 0.39 & 0.35 & 0.39 & 0.68 & 0.44 & 0.44 & 0.36 & 0.30 \\
 & Sports & 0.67 & 0.40 & 0.33 & 0.44 & 0.72 & 0.49 & 0.57 & 0.42 & 0.41 \\
\midrule
\multirow{9}{*}{\texttt{Qwen3-VL-8B}} & Cult. & 0.67 & 0.45 & 0.41 & 0.65 & 0.67 & 0.35 & 0.38 & 0.62 & 0.64 \\
 & Food & 0.69 & 0.40 & 0.34 & 0.73 & 0.68 & 0.40 & 0.37 & 0.69 & 0.74 \\
 & Hist. & 0.68 & 0.43 & 0.46 & 0.69 & 0.66 & 0.35 & 0.34 & 0.63 & 0.72 \\
 & M\&M & 0.67 & 0.36 & 0.38 & 0.38 & 0.69 & 0.37 & 0.35 & 0.30 & 0.35 \\
 & Nat. Achv. & 0.67 & 0.46 & 0.45 & 0.63 & 0.66 & 0.42 & 0.31 & 0.57 & 0.81 \\
 & Nature & 0.68 & 0.43 & 0.47 & 0.38 & 0.64 & 0.30 & 0.31 & 0.35 & 0.43 \\
 & Pers. & 0.64 & 0.42 & 0.32 & 0.50 & 0.65 & 0.42 & 0.30 & 0.44 & 0.46 \\
 & Pol. & 0.64 & 0.42 & 0.38 & 0.43 & 0.66 & 0.39 & 0.34 & 0.44 & 0.34 \\
 & Sports & 0.66 & 0.47 & 0.40 & 0.50 & 0.66 & 0.38 & 0.34 & 0.47 & 0.58 \\
\midrule
\multirow{9}{*}{\texttt{GPT-4.1-mini}} & Cult. & 0.70 & 0.44 & 0.45 & 0.65 & 0.71 & 0.52 & 0.48 & 0.58 & 0.72 \\
 & Food & 0.71 & 0.41 & 0.40 & 0.85 & 0.73 & 0.51 & 0.51 & 0.73 & 0.88 \\
 & Hist. & 0.69 & 0.44 & 0.43 & 0.61 & 0.71 & 0.50 & 0.46 & 0.57 & 0.68 \\
 & M\&M & 0.68 & 0.36 & 0.41 & 0.48 & 0.90 & 0.64 & 0.56 & 0.38 & 0.49 \\
 & Nat. Achv. & 0.69 & 0.43 & 0.45 & 0.62 & 0.71 & 0.53 & 0.52 & 0.62 & 0.82 \\
 & Nature & 0.70 & 0.39 & 0.41 & 0.49 & 0.71 & 0.50 & 0.47 & 0.51 & 0.59 \\
 & Pers. & 0.65 & 0.37 & 0.39 & 0.57 & 0.70 & 0.51 & 0.43 & 0.57 & 0.74 \\
 & Pol. & 0.67 & 0.38 & 0.37 & 0.42 & 0.69 & 0.49 & 0.47 & 0.51 & 0.57 \\
 & Sports & 0.68 & 0.42 & 0.44 & 0.53 & 0.69 & 0.48 & 0.45 & 0.50 & 0.61 \\
\bottomrule
\end{tabular*}
\caption{Full results by model and domain for Noakhali Dialect.}
\label{tab:Results-Noakhali}
\end{table*}

\begin{table*}[t]
\centering
\small
\scriptsize
\begin{tabular*}{\textwidth}{@{\extracolsep{\fill}} l c | c c c c | c c c c | c }
\toprule
 &  & \multicolumn{4}{c|}{\textbf{Zero-Shot}} & \multicolumn{4}{c|}{\textbf{Few-Shot}} & \textbf{CoT} \\
\textbf{Models} & \textbf{Domain} & \multicolumn{3}{c}{\textbf{Caption}} & \textbf{VQA} & \multicolumn{3}{c}{\textbf{Caption}} & \textbf{VQA} & \textbf{VQA} \\
\cmidrule{3-5}\cmidrule{7-9}
 &  & \textbf{B-F1} & \textbf{Gemini} & \textbf{GPT} & \textbf{Accuracy (\%)} & \textbf{B-F1} & \textbf{Gemini} & \textbf{GPT} & \textbf{Accuracy (\%)} & \textbf{Accuracy (\%)} \\
\midrule
\multirow{9}{*}{\texttt{Gemma3-4B}} & Cult. & 0.68 & 0.49 & 0.55 & 0.63 & 0.68 & 0.49 & 0.50 & 0.50 & 0.59 \\
 & Food & 0.70 & 0.50 & 0.52 & 0.73 & 0.70 & 0.52 & 0.56 & 0.56 & 0.72 \\
 & Hist. & 0.67 & 0.48 & 0.38 & 0.44 & 0.68 & 0.44 & 0.40 & 0.42 & 0.40 \\
 & M\&M & 0.66 & 0.44 & 0.42 & 0.27 & 0.80 & 0.52 & 0.51 & 0.29 & 0.30 \\
 & Nat. Achv. & 0.67 & 0.52 & 0.57 & 0.35 & 0.66 & 0.44 & 0.39 & 0.27 & 0.45 \\
 & Nature & 0.69 & 0.55 & 0.56 & 0.67 & 0.69 & 0.57 & 0.54 & 0.69 & 0.41 \\
 & Pers. & 0.63 & 0.42 & 0.43 & 0.59 & 0.67 & 0.46 & 0.41 & 0.55 & 0.53 \\
 & Pol. & 0.64 & 0.41 & 0.45 & 0.42 & 0.69 & 0.45 & 0.43 & 0.38 & 0.41 \\
 & Sports & 0.65 & 0.48 & 0.41 & 0.48 & 0.66 & 0.47 & 0.46 & 0.37 & 0.46 \\
\midrule
\multirow{9}{*}{\texttt{Gemma3-12B}} & Cult. & 0.69 & 0.45 & 0.48 & 0.62 & 0.68 & 0.44 & 0.43 & 0.62 & 0.82 \\
 & Food & 0.70 & 0.47 & 0.45 & 0.85 & 0.71 & 0.51 & 0.47 & 0.78 & 0.82 \\
 & Hist. & 0.67 & 0.43 & 0.35 & 0.64 & 0.68 & 0.44 & 0.37 & 0.60 & 0.76 \\
 & M\&M & 0.65 & 0.41 & 0.40 & 0.37 & 0.82 & 0.51 & 0.52 & 0.32 & 0.44 \\
 & Nat. Achv. & 0.65 & 0.45 & 0.52 & 0.52 & 0.68 & 0.46 & 0.41 & 0.52 & 0.71 \\
 & Nature & 0.68 & 0.43 & 0.42 & 0.44 & 0.69 & 0.46 & 0.50 & 0.54 & 0.58 \\
 & Pers. & 0.63 & 0.42 & 0.37 & 0.66 & 0.65 & 0.40 & 0.33 & 0.71 & 0.76 \\
 & Pol. & 0.65 & 0.41 & 0.38 & 0.55 & 0.67 & 0.40 & 0.34 & 0.54 & 0.62 \\
 & Sports & 0.64 & 0.44 & 0.42 & 0.50 & 0.66 & 0.44 & 0.39 & 0.52 & 0.54 \\
\midrule
\multirow{9}{*}{\texttt{Gemma3-27B}} & Cult. & 0.70 & 0.52 & 0.59 & 0.69 & 0.68 & 0.49 & 0.45 & 0.72 & 0.64 \\
 & Food & 0.71 & 0.50 & 0.59 & 0.88 & 0.72 & 0.52 & 0.48 & 0.90 & 0.87 \\
 & Hist. & 0.68 & 0.47 & 0.53 & 0.69 & 0.69 & 0.47 & 0.39 & 0.66 & 0.75 \\
 & M\&M & 0.67 & 0.43 & 0.45 & 0.33 & 0.84 & 0.52 & 0.55 & 0.34 & 0.43 \\
 & Nat. Achv. & 0.68 & 0.48 & 0.52 & 0.52 & 0.69 & 0.43 & 0.40 & 0.55 & 0.73 \\
 & Nature & 0.69 & 0.49 & 0.54 & 0.59 & 0.70 & 0.54 & 0.54 & 0.60 & 0.67 \\
 & Pers. & 0.64 & 0.42 & 0.49 & 0.78 & 0.69 & 0.48 & 0.38 & 0.78 & 0.79 \\
 & Pol. & 0.65 & 0.43 & 0.45 & 0.57 & 0.68 & 0.46 & 0.44 & 0.51 & 0.51 \\
 & Sports & 0.67 & 0.48 & 0.53 & 0.66 & 0.68 & 0.45 & 0.44 & 0.53 & 0.84 \\
\midrule
\multirow{9}{*}{\texttt{Qwen2.5-VL-7B}} & Cult. & 0.69 & 0.49 & 0.52 & 0.55 & 0.69 & 0.47 & 0.43 & 0.40 & 0.59 \\
 & Food & 0.69 & 0.43 & 0.34 & 0.68 & 0.72 & 0.46 & 0.34 & 0.57 & 0.68 \\
 & Hist. & 0.70 & 0.51 & 0.52 & 0.60 & 0.71 & 0.52 & 0.54 & 0.51 & 0.50 \\
 & M\&M & 0.69 & 0.44 & 0.31 & 0.22 & 0.81 & 0.57 & 0.55 & 0.26 & 0.20 \\
 & Nat. Achv. & 0.68 & 0.49 & 0.46 & 0.69 & 0.69 & 0.50 & 0.51 & 0.71 & 0.51 \\
 & Nature & 0.70 & 0.51 & 0.55 & 0.42 & 0.71 & 0.50 & 0.46 & 0.70 & 0.40 \\
 & Pers. & 0.65 & 0.45 & 0.41 & 0.47 & 0.68 & 0.47 & 0.41 & 0.39 & 0.53 \\
 & Pol. & 0.66 & 0.47 & 0.49 & 0.39 & 0.67 & 0.49 & 0.55 & 0.36 & 0.33 \\
 & Sports & 0.69 & 0.49 & 0.59 & 0.48 & 0.72 & 0.54 & 0.63 & 0.44 & 0.44 \\
\midrule
\multirow{9}{*}{\texttt{Qwen3-VL-8B}} & Cult. & 0.66 & 0.43 & 0.49 & 0.57 & 0.66 & 0.35 & 0.38 & 0.65 & 0.66 \\
 & Food & 0.68 & 0.41 & 0.32 & 0.75 & 0.68 & 0.41 & 0.41 & 0.70 & 0.73 \\
 & Hist. & 0.67 & 0.44 & 0.40 & 0.70 & 0.67 & 0.38 & 0.31 & 0.72 & 0.65 \\
 & M\&M & 0.66 & 0.34 & 0.34 & 0.32 & 0.67 & 0.38 & 0.32 & 0.27 & 0.38 \\
 & Nat. Achv. & 0.65 & 0.41 & 0.45 & 0.66 & 0.64 & 0.42 & 0.32 & 0.55 & 0.65 \\
 & Nature & 0.67 & 0.42 & 0.47 & 0.57 & 0.65 & 0.34 & 0.39 & 0.50 & 0.49 \\
 & Pers. & 0.62 & 0.41 & 0.34 & 0.57 & 0.64 & 0.44 & 0.40 & 0.53 & 0.54 \\
 & Pol. & 0.64 & 0.42 & 0.42 & 0.45 & 0.67 & 0.45 & 0.37 & 0.47 & 0.44 \\
 & Sports & 0.64 & 0.44 & 0.35 & 0.48 & 0.65 & 0.36 & 0.37 & 0.42 & 0.61 \\
\midrule
\multirow{9}{*}{\texttt{GPT-4.1-mini}} & Cult. & 0.69 & 0.42 & 0.41 & 0.66 & 0.70 & 0.50 & 0.51 & 0.58 & 0.78 \\
 & Food & 0.72 & 0.44 & 0.49 & 0.86 & 0.73 & 0.54 & 0.58 & 0.68 & 0.89 \\
 & Hist. & 0.69 & 0.45 & 0.39 & 0.63 & 0.69 & 0.48 & 0.43 & 0.62 & 0.72 \\
 & M\&M & 0.69 & 0.39 & 0.35 & 0.53 & 0.91 & 0.63 & 0.63 & 0.39 & 0.43 \\
 & Nat. Achv. & 0.69 & 0.45 & 0.45 & 0.57 & 0.70 & 0.52 & 0.54 & 0.58 & 0.74 \\
 & Nature & 0.69 & 0.41 & 0.36 & 0.55 & 0.71 & 0.53 & 0.49 & 0.49 & 0.69 \\
 & Pers. & 0.64 & 0.39 & 0.32 & 0.59 & 0.70 & 0.51 & 0.54 & 0.46 & 0.72 \\
 & Pol. & 0.66 & 0.35 & 0.35 & 0.46 & 0.69 & 0.46 & 0.49 & 0.46 & 0.51 \\
 & Sports & 0.67 & 0.43 & 0.48 & 0.53 & 0.71 & 0.53 & 0.53 & 0.50 & 0.74 \\
\bottomrule
\end{tabular*}
\caption{Full results by model and domain for Rangpur Dialect.}
\label{tab:Results-Rangpur}
\end{table*}

\begin{table*}[t]
\centering
\small
\scriptsize
\begin{tabular*}{\textwidth}{@{\extracolsep{\fill}} l c | c c c c | c c c c | c }
\toprule
 &  & \multicolumn{4}{c|}{\textbf{Zero-Shot}} & \multicolumn{4}{c|}{\textbf{Few-Shot}} & \textbf{CoT} \\
\textbf{Models} & \textbf{Domain} & \multicolumn{3}{c}{\textbf{Caption}} & \textbf{VQA} & \multicolumn{3}{c}{\textbf{Caption}} & \textbf{VQA} & \textbf{VQA} \\
\cmidrule{3-5}\cmidrule{7-9}
 &  & \textbf{B-F1} & \textbf{Gemini} & \textbf{GPT} & \textbf{Accuracy (\%)} & \textbf{B-F1} & \textbf{Gemini} & \textbf{GPT} & \textbf{Accuracy (\%)} & \textbf{Accuracy (\%)} \\
\midrule
\multirow{9}{*}{\texttt{Gemma3-4B}} & Cult. & 0.67 & 0.47 & 0.45 & 0.60 & 0.68 & 0.45 & 0.40 & 0.45 & 0.58 \\
 & Food & 0.70 & 0.46 & 0.50 & 0.72 & 0.70 & 0.50 & 0.48 & 0.62 & 0.64 \\
 & Hist. & 0.66 & 0.42 & 0.43 & 0.44 & 0.68 & 0.43 & 0.33 & 0.40 & 0.47 \\
 & M\&M & 0.65 & 0.42 & 0.35 & 0.27 & 0.80 & 0.50 & 0.50 & 0.35 & 0.36 \\
 & Nat. Achv. & 0.65 & 0.47 & 0.43 & 0.31 & 0.68 & 0.48 & 0.40 & 0.29 & 0.45 \\
 & Nature & 0.69 & 0.58 & 0.58 & 0.59 & 0.69 & 0.56 & 0.54 & 0.63 & 0.52 \\
 & Pers. & 0.63 & 0.45 & 0.40 & 0.58 & 0.67 & 0.45 & 0.38 & 0.54 & 0.51 \\
 & Pol. & 0.64 & 0.45 & 0.49 & 0.40 & 0.68 & 0.43 & 0.33 & 0.42 & 0.40 \\
 & Sports & 0.64 & 0.47 & 0.43 & 0.47 & 0.67 & 0.47 & 0.53 & 0.42 & 0.44 \\
\midrule
\multirow{9}{*}{\texttt{Gemma3-12B}} & Cult. & 0.68 & 0.40 & 0.45 & 0.62 & 0.68 & 0.43 & 0.37 & 0.65 & 0.78 \\
 & Food & 0.70 & 0.40 & 0.38 & 0.82 & 0.71 & 0.48 & 0.37 & 0.67 & 0.91 \\
 & Hist. & 0.66 & 0.37 & 0.34 & 0.64 & 0.69 & 0.39 & 0.35 & 0.64 & 0.64 \\
 & M\&M & 0.66 & 0.39 & 0.33 & 0.35 & 0.80 & 0.49 & 0.48 & 0.33 & 0.45 \\
 & Nat. Achv. & 0.65 & 0.38 & 0.32 & 0.46 & 0.67 & 0.43 & 0.31 & 0.38 & 0.58 \\
 & Nature & 0.68 & 0.41 & 0.34 & 0.47 & 0.68 & 0.45 & 0.38 & 0.54 & 0.42 \\
 & Pers. & 0.63 & 0.31 & 0.31 & 0.68 & 0.65 & 0.34 & 0.35 & 0.65 & 0.81 \\
 & Pol. & 0.63 & 0.33 & 0.31 & 0.54 & 0.67 & 0.39 & 0.30 & 0.52 & 0.59 \\
 & Sports & 0.65 & 0.39 & 0.31 & 0.45 & 0.66 & 0.40 & 0.37 & 0.44 & 0.50 \\
\midrule
\multirow{9}{*}{\texttt{Gemma3-27B}} & Cult. & 0.69 & 0.48 & 0.41 & 0.70 & 0.69 & 0.48 & 0.42 & 0.68 & 0.78 \\
 & Food & 0.71 & 0.45 & 0.36 & 0.84 & 0.72 & 0.53 & 0.47 & 0.82 & 0.86 \\
 & Hist. & 0.69 & 0.44 & 0.41 & 0.66 & 0.69 & 0.46 & 0.46 & 0.63 & 0.75 \\
 & M\&M & 0.67 & 0.42 & 0.37 & 0.34 & 0.84 & 0.51 & 0.55 & 0.33 & 0.45 \\
 & Nat. Achv. & 0.68 & 0.43 & 0.51 & 0.51 & 0.68 & 0.45 & 0.44 & 0.49 & 0.69 \\
 & Nature & 0.69 & 0.46 & 0.44 & 0.59 & 0.70 & 0.50 & 0.43 & 0.60 & 0.63 \\
 & Pers. & 0.65 & 0.36 & 0.33 & 0.79 & 0.69 & 0.48 & 0.37 & 0.82 & 0.78 \\
 & Pol. & 0.66 & 0.38 & 0.40 & 0.51 & 0.68 & 0.46 & 0.39 & 0.47 & 0.55 \\
 & Sports & 0.68 & 0.42 & 0.44 & 0.63 & 0.67 & 0.41 & 0.40 & 0.56 & 0.68 \\
\midrule
\multirow{9}{*}{\texttt{Qwen2.5-VL-7B}} & Cult. & 0.67 & 0.35 & 0.38 & 0.52 & 0.69 & 0.32 & 0.37 & 0.42 & 0.53 \\
 & Food & 0.69 & 0.32 & 0.31 & 0.55 & 0.71 & 0.35 & 0.38 & 0.45 & 0.59 \\
 & Hist. & 0.68 & 0.42 & 0.38 & 0.58 & 0.70 & 0.43 & 0.39 & 0.55 & 0.52 \\
 & M\&M & 0.67 & 0.35 & 0.35 & 0.21 & 0.83 & 0.53 & 0.52 & 0.22 & 0.22 \\
 & Nat. Achv. & 0.67 & 0.36 & 0.39 & 0.66 & 0.70 & 0.40 & 0.42 & 0.71 & 0.52 \\
 & Nature & 0.69 & 0.39 & 0.36 & 0.43 & 0.70 & 0.44 & 0.36 & 0.60 & 0.38 \\
 & Pers. & 0.64 & 0.36 & 0.37 & 0.49 & 0.67 & 0.41 & 0.43 & 0.48 & 0.51 \\
 & Pol. & 0.66 & 0.38 & 0.30 & 0.39 & 0.69 & 0.44 & 0.39 & 0.34 & 0.36 \\
 & Sports & 0.67 & 0.41 & 0.37 & 0.45 & 0.69 & 0.44 & 0.37 & 0.37 & 0.48 \\
\midrule
\multirow{9}{*}{\texttt{Qwen3-VL-8B}} & Cult. & 0.66 & 0.45 & 0.35 & 0.66 & 0.65 & 0.35 & 0.41 & 0.59 & 0.68 \\
 & Food & 0.68 & 0.44 & 0.32 & 0.71 & 0.69 & 0.45 & 0.35 & 0.68 & 0.73 \\
 & Hist. & 0.67 & 0.44 & 0.40 & 0.72 & 0.66 & 0.35 & 0.31 & 0.67 & 0.72 \\
 & M\&M & 0.66 & 0.34 & 0.35 & 0.42 & 0.67 & 0.37 & 0.36 & 0.44 & 0.37 \\
 & Nat. Achv. & 0.67 & 0.45 & 0.45 & 0.66 & 0.65 & 0.34 & 0.39 & 0.69 & 0.72 \\
 & Nature & 0.68 & 0.47 & 0.43 & 0.48 & 0.64 & 0.32 & 0.36 & 0.42 & 0.44 \\
 & Pers. & 0.63 & 0.41 & 0.34 & 0.49 & 0.64 & 0.45 & 0.35 & 0.56 & 0.56 \\
 & Pol. & 0.64 & 0.44 & 0.36 & 0.42 & 0.66 & 0.42 & 0.33 & 0.41 & 0.42 \\
 & Sports & 0.66 & 0.44 & 0.40 & 0.53 & 0.64 & 0.39 & 0.33 & 0.51 & 0.59 \\
\midrule
\multirow{9}{*}{\texttt{GPT-4.1-mini}} & Cult. & 0.69 & 0.43 & 0.35 & 0.63 & 0.70 & 0.51 & 0.56 & 0.59 & 0.71 \\
 & Food & 0.72 & 0.46 & 0.45 & 0.84 & 0.72 & 0.50 & 0.46 & 0.66 & 0.86 \\
 & Hist. & 0.69 & 0.46 & 0.45 & 0.58 & 0.70 & 0.49 & 0.54 & 0.58 & 0.68 \\
 & M\&M & 0.69 & 0.40 & 0.38 & 0.47 & 0.90 & 0.59 & 0.65 & 0.43 & 0.50 \\
 & Nat. Achv. & 0.69 & 0.48 & 0.47 & 0.64 & 0.70 & 0.55 & 0.62 & 0.62 & 0.76 \\
 & Nature & 0.70 & 0.45 & 0.52 & 0.53 & 0.72 & 0.54 & 0.53 & 0.47 & 0.67 \\
 & Pers. & 0.65 & 0.36 & 0.39 & 0.63 & 0.70 & 0.51 & 0.46 & 0.51 & 0.71 \\
 & Pol. & 0.66 & 0.40 & 0.40 & 0.48 & 0.68 & 0.47 & 0.41 & 0.44 & 0.63 \\
 & Sports & 0.68 & 0.46 & 0.55 & 0.47 & 0.70 & 0.50 & 0.48 & 0.47 & 0.73 \\
\bottomrule
\end{tabular*}
\caption{Full results by model and domain for Sylhet Dialect.}
\label{tab:Results-Sylhet}
\end{table*}

\end{document}